\documentclass{article}

\title{ChronoFlow: Hierarchical Flow Matching for Irregular Time Series Generation}

\author{%
Changhun Kim$^{1,2*}$ \quad Sunguk Jang$^{2,3*}$ \quad Jeongjun Lee$^{3}$ \quad Juhwan Choi$^{2}$ \\ \textbf{Sangchul Hahn}$^{2}$ \quad \textbf{Grigorios Chrysos}$^1$ \quad \textbf{Eunho Yang}$^{2, 3\dagger}$ \quad \textbf{Juho Lee}$^{3\dagger}$ \\
$^1$University of Wisconsin--Madison \quad $^2$AITRICS \quad $^3$KAIST \\
\texttt{\{changhun.kim, chrysos\}@wisc.edu} \\
\texttt{\{jhchoi, s.hahn\}@aitrics.com} \\
\texttt{\{tommy9773, lee.jeongjun, eunhoy, juholee\}@kaist.ac.kr} \\
}

\usepackage{iclr2027_conference}

\usepackage{times}
\usepackage[utf8]{inputenc} 
\usepackage[T1]{fontenc}    
\usepackage{hyperref}       
\usepackage{url}            
\usepackage{booktabs}       
\usepackage{amsfonts}       
\usepackage{bbm}            
\usepackage{nicefrac}       
\usepackage{xcolor}         
\usepackage{colortbl}

\usepackage[labelfont=bf]{caption}
\usepackage{graphicx}
\usepackage{amsmath}
\usepackage{amssymb}
\usepackage{mathtools}
\usepackage{amsthm}
\usepackage[capitalize,noabbrev]{cleveref}
\usepackage{enumitem}
\usepackage{algorithm}
\usepackage{algpseudocode}
\usepackage{subcaption}      
\usepackage{multicol}
\usepackage{multirow}
\usepackage{makecell}
\usepackage{pifont}
\usepackage{wrapfig}
\usepackage{longtable}

\usepackage{amsmath,amsfonts,bm}

\def\eqref#1{equation~\ref{#1}}

\def\1{\bm{1}}

\DeclareMathAlphabet{\mathsfit}{\encodingdefault}{\sfdefault}{m}{sl}
\SetMathAlphabet{\mathsfit}{bold}{\encodingdefault}{\sfdefault}{bx}{n}

\theoremstyle{plain}
\newtheorem{theorem}{Theorem}
\newtheorem{proposition}[theorem]{Proposition}

\theoremstyle{definition}

\theoremstyle{remark}

\algrenewcommand\algorithmicrequire{\textbf{Input:}}
\algrenewcommand\algorithmicensure{\textbf{Output:}}
\algnewcommand\Input{\item[\algorithmicinput]}
\algnewcommand\Output{\it em[\algorithmicoutput]}

\newcommand{\method}{ChronoFlow}

\iclrfinalcopy 
\begin{document}
\maketitle
\def\thefootnote{*}\footnotetext{Equal contribution. $^\dagger$Equal advising.}

\begin{abstract}
Recent advances in generative modeling have substantially improved time series generation, yet most existing methods either assume a regular temporal grid or focus on feature dynamics under a given sampling structure.
This makes them ill-suited for generating irregular time series in their native form, where a model must capture not only feature values, but also how many observations occur, when they occur, and which features are observed together.
To address this heterogeneous generation problem, we propose \method{}, a unified hierarchical flow matching framework organized by statistical granularity.
Following a coarse-to-fine hierarchy, \method{} first generates observation counts and feature-wise frequencies, then jointly generates observation times and feature co-observation patterns, and finally generates values conditioned on the realized pattern.
This turns a complex joint generation problem into structurally aligned subproblems while preserving their dependencies.
To evaluate complete irregular time series generation, we introduce complementary metrics spanning sample realism, sampling structure, value fidelity, and temporal and cross-feature dependencies, and validate them through controlled corruptions.
Across five benchmarks, ChronoFlow achieves the lowest mean in most dataset--metric comparisons against existing full-generation baselines, while factorization studies support the proposed hierarchy.
Our code is available at \url{https://anonymous.4open.science/r/\method}.
\end{abstract}
\section{Introduction}

Recent advances in generative modeling, such as diffusion models~\citep{ho2020denoising,song2020score} and flow matching~\citep{lipman2022flow}, have achieved strong performance across diverse data modalities~\citep{rombach2022high,gupta2024photorealistic,chi2025diffusion,li2022diffusion}.
These advances have also been extended to time series generation~\citep{yuan2024diffusion,hu2024flowts}, often on regularly sampled data.
In real-world applications, however, measurements are collected irregularly, with features observed asynchronously and at uneven intervals, as in electronic health records and physiological sensing~\citep{rubanova2019latent,zhang2024trajectory}.
Such sampling patterns exhibit their own structure in observation frequency, timing, and feature co-observation.
Generating irregular time series, therefore, requires modeling not only feature values, but also the sample-specific pattern governing how much, when, and what is observed.

A common way to apply regular-grid time series generators~\citep{yoon2019time,yuan2024diffusion,hu2024flowts} to irregular data is to bin observations onto a fixed temporal grid.
This can merge distinct observation times and constrain generation to the chosen resolution.
Methods for irregular data can instead represent values in continuous time~\citep{chen2018neural,rubanova2019latent,shukla2021multitime,zhang2024trajectory}.
Some also model or predict observation times~\citep{rubanova2019latent,zhang2024trajectory}, while event generators model feature identities~\citep{draxler2025transformers}.
However, none of these explicitly represents which features are observed together at each time.
This raises a central research question:
\begin{center}
\textit{What is the right generative formulation for irregular time series in their native form?}
\end{center}

To answer this question, we formulate irregular samples through observation occasions and their associated feature panels, and propose \emph{\method{}} to generate them by \emph{statistical granularity}.
We build on established flow matching and Transformer components~\citep{lipman2022flow,vaswani2017attention}, with contributions centered on this representation, hierarchical generation, and evaluation.
\method{} first generates global observation statistics, including the number of observation occasions and feature-wise observation frequencies.
Conditioned on this coarse structure, it jointly generates observation times and feature co-observation patterns, treating when and what is observed as coupled components of the same occasion-level sampling structure.
Finally, at the measurement level, it generates feature values conditioned on the realized sampling pattern, since a value is meaningful only once its feature and observation time are specified.
Flow matching models continuous times and values and continuous representations of discrete statistics and panels, while hierarchical conditioning preserves dependencies across stages.

Beyond modeling irregular time series itself, full generation poses a distinct evaluation challenge.
Interpolation and extrapolation evaluate missing or future values given partial observations~\citep{rubanova2019latent,shukla2021multitime,zhang2024trajectory}.
These tasks do not directly diagnose the observation frequency, timing, and feature co-observation of \emph{complete} samples generated without observed context.
We therefore evaluate generation across complementary aspects and resolutions, from global sample-level realism to sampling structure, marginal values, local temporal dynamics, and cross-feature dependencies.
Across five benchmarks, \method{} achieves the lowest mean among generators in most dataset--metric comparisons, while factorization studies on eICU support the proposed hierarchy over joint and reverse-order alternatives.
Controlled corruptions probe the sensitivity of these diagnostics to selected distortions.

To summarize, our contributions are as follows:
\begin{itemize}
    \item We formulate complete irregular samples through observation occasions and feature panels, and introduce \method{} to generate global statistics, sampling patterns, and values through hierarchical conditioning.
    
    \item We introduce an evaluation protocol spanning sample realism, sampling structure, value fidelity, temporal dynamics, and cross-feature dependencies, and assess metric sensitivity through controlled corruptions.

    \item Experiments on five benchmarks show that ChronoFlow achieves state-of-the-art generation fidelity across most datasets and metrics, while eICU factorization studies and oracle interventions examine the hierarchy and its sensitivity to upstream errors.
\end{itemize}
\section{Related Work}

\paragraph{Time series and irregular-trajectory generation.}
TimeGAN, Diffusion-TS, and FlowTS generate on shared temporal grids~\citep{yoon2019time,yuan2024diffusion,hu2024flowts}.
ImagenI2R learns from irregular samples but generates regular sequences~\citep{fadlon2025diffusion}.
TimEHR generates missingness masks and then conditional values in an image representation~\citep{karami2025timehr}, establishing a precedent for pattern-to-value generation.
mTAN predicts values at supplied times, while TFM models continuous trajectories and predicts the next observation time~\citep{shukla2021multitime,zhang2024trajectory}.

\paragraph{Observation-process and event generation.}
LatentODE jointly models times and values through a Poisson-process extension~\citep{rubanova2019latent}.
Intensity-free point processes model inter-event times, and FlexTPP generates mixed-type event sequences~\citep{shchur2020intensity,draxler2025transformers}.
Our distinction is an explicit occasion--panel representation with global count and feature-frequency conditioning, joint time--panel generation, and conditional values.
The contribution lies in this formulation and evaluation, using established flow matching and attention components.

\paragraph{Evaluation of full irregular samples.}
Prior evaluations include regular-grid fidelity, conditional prediction, and event generation~\citep{yoon2019time,shukla2021multitime,draxler2025transformers}.
We organize established comparisons, including classifier discrimination and kernel discrepancies~\citep{yoon2019time,JMLR:v13:gretton12a}, around complete irregular samples.
The protocol combines diagnostics of counts, timing, co-observation, and value dependencies, with controlled corruptions probing their sensitivity.

\section{Proposed Method: \method{}}
\label{sec:method}

This section presents \method{} and its hierarchical factorization by statistical granularity.
We first represent an irregular time series as
$\mathcal{S}=(\boldsymbol{\tau},\boldsymbol{B},\mathbf{x})$,
where $\boldsymbol{\tau}\in\mathbb{R}^{M}$ contains the $M$ observation times,
$\boldsymbol{B}\in\{0,1\}^{M\times F}$ is the corresponding binary feature panel over $F$ features,
and $\mathbf{x}$ contains the observed values.
Specifically, $\boldsymbol{B}_{m,f}=1$ indicates that feature $f$ is observed at $\tau_m$ with value $\mathbf{x}_{m,f}$.
The total number of measurements is
$N=\sum_{m,f}\boldsymbol{B}_{m,f}$.
For each training sample, the feature-wise observation frequency is
\begin{equation}
r_f
=
\frac{1}{M}
\sum_{m=1}^{M}
\boldsymbol{B}_{m,f},
\qquad
\boldsymbol{r}=(r_1,\ldots,r_F)\in[0,1]^F.
\end{equation}

\begin{figure*}[!t]
\centering
\includegraphics[width=\textwidth]{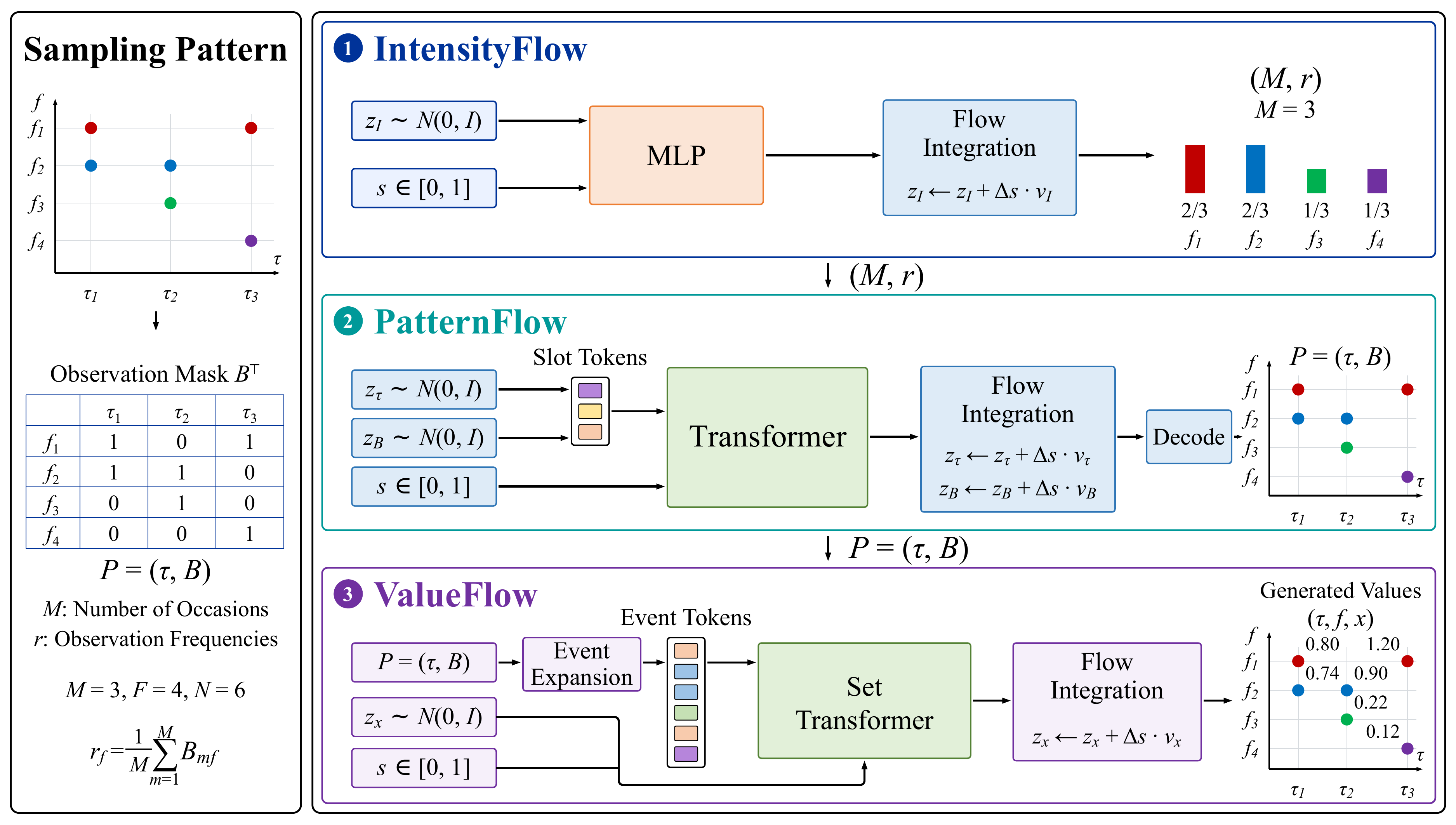}
\vspace{-.1in}
\caption{
Overview of \method{} for irregular time series generation.
IntensityFlow generates global statistics, PatternFlow joint times and panels, and ValueFlow conditional values.
The feature-by-time display is $\boldsymbol B^\top$. Figure notation $s$, $z_I$, and $P$ corresponds to flow time $t$, state $\boldsymbol u$, and pattern $(\boldsymbol\tau,\boldsymbol B)$.
}
\label{fig:main_figure}
\end{figure*}

Direct generation combines variable cardinality, continuous times and values, and discrete feature panels.
\method{} organizes this heterogeneous problem by \emph{statistical granularity}, from global statistics through occasion-level patterns to measurement-level values.
Writing $p$ for the model distribution, we factorize the augmented variables $(M,\boldsymbol{r},\mathcal{S})$ as follows:
\begin{equation}
\label{eq:chronoflow-factorization}
\underbrace{p(M,\boldsymbol{r},\mathcal{S})}_{\textrm{\method{}}}
=
\underbrace{p(M,\boldsymbol{r})}_{\text{IntensityFlow}}
\underbrace{p(\boldsymbol{\tau},\boldsymbol{B}\mid M,\boldsymbol{r})}_{\text{PatternFlow}}
\underbrace{p(\mathbf{x}\mid\boldsymbol{\tau},\boldsymbol{B})}_{\text{ValueFlow}}.
\end{equation}
During generation, $(M,\boldsymbol{r})$ serve as sampled conditioning variables and are discarded after generating $\mathcal{S}$, whose decoded panels need not reproduce $\boldsymbol{r}$ exactly.
As illustrated in \Cref{fig:main_figure}, we first generate global observation statistics with IntensityFlow, then the sampling pattern with PatternFlow, and finally the corresponding values with ValueFlow.

\subsection{Background: Flow Matching}
\label{subsec:flow-matching}

All three stages are instantiated with flow matching~\citep{lipman2022flow}, providing a common generative principle across continuous variables and continuous representations of discrete quantities.
Given $\mathbf{x}_0\sim p_0$, $\mathbf{x}_1\sim p_{\mathrm{data}}$, and $t\sim\mathcal{U}[0,1]$, we use the linear path
$\mathbf{x}_t=(1-t)\mathbf{x}_0+t\mathbf{x}_1$ with target velocity
$\mathbf{x}_1-\mathbf{x}_0$.
The velocity field $v_\theta$ is trained by
\begin{equation}
\mathcal{L}_{\mathrm{FM}}
=
\mathbb{E}
\left[
\left\|
v_\theta(\mathbf{x}_t,t)
-
(\mathbf{x}_1-\mathbf{x}_0)
\right\|_2^2
\right].
\end{equation}
After training, samples are generated from $\mathbf{x}_0\sim p_0$ by solving
$\frac{d\mathbf{x}_t}{dt}=v_\theta(\mathbf{x}_t,t)$ for $t\in[0,1]$.

\subsection{Hierarchical Architecture of \method{}}
\label{subsec:chronoflow-arch}

\paragraph{IntensityFlow.}
We first model the global observation statistics $(M,\boldsymbol{r})$ with IntensityFlow.
We use a dataset-specific upper bound $M_{\max}$ on the number of retained observation occasions.
We map $M/M_{\max}$ and $\boldsymbol{r}$ to $[-1,1]$ and form
$\boldsymbol{u}_1=[\,2M/M_{\max}-1,\;2\boldsymbol{r}-1\,]\in[-1,1]^{1+F}$.

Given $\boldsymbol{u}_0\sim\mathcal{N}(0,I_{1+F})$ and
$\boldsymbol{u}_t=(1-t)\boldsymbol{u}_0+t\boldsymbol{u}_1$,
IntensityFlow is trained with
\begin{equation}
\mathcal{L}_{I}
=
\mathbb{E}
\left[
\left\|
v_{\theta_I}(\boldsymbol{u}_t,t)
-
(\boldsymbol{u}_1-\boldsymbol{u}_0)
\right\|_2^2
\right].
\end{equation}
Since $(M,\boldsymbol{r})$ is a fixed-dimensional global summary with no sequential structure, a lightweight MLP suffices to parameterize $v_{\theta_I}$.
Generating $(M,\boldsymbol{r})$ first supplies PatternFlow with global cardinality and feature-frequency information, allowing it to focus on their local arrangement.
At inference, $M$ is rounded to an integer and each $r_f$ is projected onto
$\{0,1/M,\ldots,1\}$.

\paragraph{PatternFlow.}
We then generate the occasion-level sampling structure
$(\boldsymbol{\tau},\boldsymbol{B})$ with PatternFlow, conditioned on $(M,\boldsymbol{r})$.
We model observation times and feature panels jointly so that their current states can inform each other during generation.
After normalizing $\boldsymbol{\tau}\in[0,1]^M$, we set
$\boldsymbol{\tau}_1=2\boldsymbol{\tau}-1$ and
$\boldsymbol{B}_1=2\boldsymbol{B}-1$.
Given $\boldsymbol{\tau}_0\sim\mathcal{N}(0,I_M)$ and
$\boldsymbol{B}_0$ with i.i.d.\ standard Gaussian entries,
$\boldsymbol{\tau}_t=(1-t)\boldsymbol{\tau}_0+t\boldsymbol{\tau}_1$ and
$\boldsymbol{B}_t=(1-t)\boldsymbol{B}_0+t\boldsymbol{B}_1$.

We represent the $M$ occasion slots as \emph{slot tokens}, each jointly encoding the current time and panel states of one occasion.
A Transformer~\citep{vaswani2017attention} processes these tokens with learned slot embeddings and a shared embedding of $(M,\boldsymbol r)$ (\Cref{app:conditioning}).
The joint velocity field $v_{\theta_P}$ uses separate heads
$v_{\theta_{P,\tau}}$ and $v_{\theta_{P,B}}$ for the time and panel components:
\begin{equation}
\mathcal{L}_{P}
=
\mathbb{E}
\left[
\left\|
v_{\theta_{P,\tau}}
-
(\boldsymbol{\tau}_1-\boldsymbol{\tau}_0)
\right\|_2^2
+
\frac{1}{F}
\left\|
v_{\theta_{P,B}}
-
(\boldsymbol{B}_1-\boldsymbol{B}_0)
\right\|_2^2
\right].
\end{equation}
At inference, the initial time noise is sorted before integration. Generated times are rescaled and clipped to $[0,1]$, while panels are thresholded and reordered with their times. Each empty panel is assigned its highest-scoring feature.

Occasion slots have no intrinsic correspondence to independently sampled Gaussian noise, so arbitrary endpoint pairings can induce unnecessarily long transport paths.
We therefore use the standard one-dimensional optimal-transport assignment by sorting $\boldsymbol{\tau}_0$ and $\boldsymbol{\tau}_1$ independently and applying the corresponding permutations to $\boldsymbol{B}_0$ and $\boldsymbol{B}_1$.

\begin{proposition}[One-dimensional optimal-transport coupling]
\label{thm:ot-patternflow}
Let $\tilde{\boldsymbol{\tau}}_0$ and
$\tilde{\boldsymbol{\tau}}_1$ denote the sorted versions of
$\boldsymbol{\tau}_0$ and $\boldsymbol{\tau}_1$, and let
$\Pi_M$ denote the set of permutations of $\{1,\ldots,M\}$.
Then
\begin{equation}
\min_{\pi\in\Pi_M}
\sum_{m=1}^{M}
\left|
(\boldsymbol{\tau}_0)_m
-
(\boldsymbol{\tau}_1)_{\pi(m)}
\right|^2
=
\sum_{m=1}^{M}
\left|
(\tilde{\boldsymbol{\tau}}_0)_m
-
(\tilde{\boldsymbol{\tau}}_1)_m
\right|^2.
\end{equation}
\end{proposition}

This minimizes squared temporal assignment cost, without asserting optimality for the joint time--panel distribution. The proof is provided in \Cref{sec:proof_thm1}.

\paragraph{ValueFlow.}
Finally, once the sampling pattern is realized, we generate its values with ValueFlow.
Let
$\mathcal{I}=\{(m,f):\boldsymbol{B}_{m,f}=1\}$, with $|\mathcal{I}|=N$.
For an enumeration
$\mathcal{I}=\{(m_i,f_i)\}_{i=1}^{N}$,
we define
$\boldsymbol{f}=(f_i)_{i=1}^{N}$,
$\boldsymbol{\tau}^{\mathrm{obs}}=(\tau_{m_i})_{i=1}^{N}$, and
$\mathbf{x}^{\mathrm{obs}}=(\mathbf{x}_{m_i,f_i})_{i=1}^{N}$.
Conditioning on the realized pattern allows ValueFlow to focus on dependencies among observed measurements rather than where they occur.

We represent the $N$ active measurements as \emph{event tokens}, each containing its feature identity, observation time, and current noisy value.
A Set Transformer~\citep{lee2019set,horn2020set} captures dependencies across features and times without imposing an arbitrary ordering.
Learnable register tokens provide shared global context.
Given $\mathbf{z}_1=\mathbf{x}^{\mathrm{obs}}$,
$\mathbf{z}_0\sim\mathcal{N}(0,I_N)$, and
$\mathbf{z}_t=(1-t)\mathbf{z}_0+t\mathbf{z}_1$,
\begin{equation}
\mathcal{L}_{V}
=
\mathbb{E}
\left[
\left\|
v_{\theta_V}
(
\boldsymbol{f},
\boldsymbol{\tau}^{\mathrm{obs}},
\mathbf{z}_t,
t
)
-
(\mathbf{z}_1-\mathbf{z}_0)
\right\|_2^2
\right].
\end{equation}

Shared embeddings without index-dependent positional encodings make ValueFlow equivariant to event-token permutations.
This standard attention property~\citep{lee2019set}, also used in equivariant flow models~\citep{klein2023equivariant}, is stated below.

\begin{proposition}[Permutation Equivariance of ValueFlow]
\label{thm:equiv-valueflow}
For any permutation matrix
$P\in\{0,1\}^{N\times N}$ acting on the $N$ event tokens
while leaving the register tokens fixed,
\begin{equation}
v_{\theta_V}
\left(
P\boldsymbol{f},
P\boldsymbol{\tau}^{\mathrm{obs}},
P\mathbf{z}_t,
t
\right)
=
P\,
v_{\theta_V}
\left(
\boldsymbol{f},
\boldsymbol{\tau}^{\mathrm{obs}},
\mathbf{z}_t,
t
\right).
\end{equation}
\end{proposition}

The proof is provided in \Cref{sec:proof_thm2}.

For batched training, variable-length occasion and measurement sets are padded, with attention and losses masked to valid entries.
The three stages are trained simultaneously, with each downstream stage conditioned on the ground-truth outputs of preceding stages.
Detailed procedures are provided in \Cref{alg:chronoflow-training,alg:chronoflow-inference}.
\section{Evaluation Metrics for Irregular Time Series Generation}
\label{sec:evaluation}

Complete generation requires matching both observation patterns and values, which conditional interpolation or extrapolation alone does not assess~\citep{rubanova2019latent,zhang2024trajectory}.
We combine six complementary diagnostics using established statistical comparisons, including classifier discrimination~\citep{yoon2019time} and kernel MMD~\citep{JMLR:v13:gretton12a}.
The contribution is their construction and joint use for irregular samples.
Superscripts $R$ and $G$ denote real and generated collections, and $d_\sigma$ denotes the reported Gaussian-kernel MMD$^2$ estimator.

\subsection{Full-Sample Distinguishability}
\label{subsec:metric-discriminative}

\paragraph{Set-Discr.}
A permutation-invariant classifier distinguishes complete real and generated samples using times, feature identities, and values.
We report its held-out accuracy deviation from chance:
\begin{equation}
\mathrm{Set\mbox{-}Discr}=|\mathrm{Acc}_{\mathrm{held}}-0.5|.
\end{equation}
A low score indicates difficulty for this trained classifier, rather than equality of the underlying distributions.

\subsection{Sampling-Pattern Fidelity}
\label{subsec:metric-pattern}

Realistic sampling requires preserving when measurements occur and which features are observed together, since these choices determine the structure of the observed record.

\paragraph{CoOb.}
We divide time into common bins and compare pairwise co-observation frequencies:
\begin{equation}
\mathrm{CoOb}=\|\mathbf c^R-\mathbf c^G\|_2,
\qquad C_{fg}=\frac{\sum_i\sum_{b\in\mathcal B_i}b_{i,b,f}b_{i,b,g}}{\sum_i|\mathcal B_i|}.
\end{equation}
Here $\mathcal B_i$ contains sample $i$'s occupied bins, $b_{i,b,f}$ indicates feature presence, and $\mathbf c=(C_{fg})_{f<g}$.
This measures whether frequently co-observed feature pairs remain so in generated data.

\paragraph{Pattern-MMD$^2$.}
To assess feature-wise counts and timing beyond pairwise co-observation, we sum Fourier time features $\phi$ without normalizing by each feature's count:
\begin{equation}
e_f(\mathcal S)=c^{-1}\sum_m B_{m,f}\phi(\tau_m),
\qquad \mathrm{Pattern\mbox{-}MMD}^2=d_{\sigma_P}(\mathcal E_R,\mathcal E_G).
\end{equation}
Here $c$ is a shared training-derived scale and $\mathcal E_Q$ contains concatenated summaries $[e_1,\ldots,e_F]$ from collection $Q$.
Their magnitude reflects counts and their components summarize timing at the chosen Fourier resolution.

\subsection{Value Fidelity and Dependencies}
\label{subsec:metric-value}

Realistic values must reproduce marginal distributions as well as temporal and cross-feature dependencies.

\paragraph{Value-W1.}
We compare each feature's empirical distribution of observed values and average across training-eligible features $\mathcal F_V$:
\begin{equation}
\mathrm{Value\mbox{-}W1}=|\mathcal F_V|^{-1}\sum_{f\in\mathcal F_V}W_1(\widehat P_f^R,\widehat P_f^G).
\end{equation}
This tests value marginals without assessing their ordering or dependence on observation patterns.

\paragraph{Trans-MMD$^2$.}
For each feature, consecutive within-sample observations form tuples of log time gap and endpoint values, $[\log(1+\Delta q),x_{\mathrm{prev}},x_{\mathrm{next}}]$.
Using benchmark time indices $q$, training-scaled tuple collections $\widetilde{\mathcal Z}_f^Q$, and eligible features $\mathcal F_T$, we compute
\begin{equation}
\mathrm{Trans\mbox{-}MMD}^2=|\mathcal F_T|^{-1}\sum_{f\in\mathcal F_T}d_{\sigma_f}(\widetilde{\mathcal Z}_f^R,\widetilde{\mathcal Z}_f^G).
\end{equation}
This assesses local value transitions together with their elapsed times.

\paragraph{Set-Corr.}
For asynchronous feature pairs, we compute Pearson correlations $\rho_{fg}$ from within-sample measurements weighted by temporal proximity:
\begin{equation}
\mathrm{Set\mbox{-}Corr}=|\mathcal J|^{-1}\sum_{(f,g)\in\mathcal J}|\rho_{fg}^R-\rho_{fg}^G|.
\end{equation}
Here $\mathcal J$ contains pairs valid in both collections, so the comparison depends on pair coverage.
All scores are lower-is-better and capture complementary, non-exhaustive aspects of fidelity.
Definitions, calibration, and interpretation are given in \Cref{app:metric_details}.

\section{Experiments}

We evaluate \method{} through four questions:
\textbf{(RQ1)} Does \method{} faithfully generate irregular time series across diverse benchmarks?
\textbf{(RQ2)} Do the hierarchical factorization and optimal-transport coupling improve generation fidelity?
\textbf{(RQ3)} Do the protocol's metrics detect targeted distributional distortions?
\textbf{(RQ4)} How robust and computationally practical is \method{} under upstream errors and limited sampling budgets?

\subsection{Experimental Setup}
\label{sec:exp_setup}

\paragraph{Datasets.}
We evaluate on five irregular time series benchmarks from two domains: electronic health records and wearable sensing.
The four clinical benchmarks are eICU~\citep{pollard2018eicu}, PhysioNet 2012 (P12)~\citep{citi2012physionet}, MIMIC-III~\citep{johnson2016mimic}, and MIMIC-IV~\citep{johnson2023mimic}, while Activity~\citep{vidulin2010localization} provides a denser non-clinical benchmark.
Dataset statistics and preprocessing details are provided in \Cref{app:dataset_details}.

\paragraph{Baselines.}
We compare \method{} with Poisson + LatentODE~\citep{rubanova2019latent},
LogNormMix + mTAN~\citep{shchur2020intensity,shukla2021multitime},
LogNormMix + TFM~\citep{shchur2020intensity,zhang2024trajectory},
and FlexTPP~\citep{draxler2025transformers}.
Poisson + LatentODE combines a Poisson observation process with latent dynamics, while the LogNormMix baselines pair sampling-pattern generation with mTAN or TFM for values.
FlexTPP directly generates variable-length event sequences.
We evaluate these implementations on generated patterns and values using the metrics in \Cref{sec:evaluation}.
Additional details are provided in \Cref{app:baseline_details}.

\paragraph{Evaluation protocol.}
For each model, we generate as many samples as in the test set and compare them against the full test set.
Any data-dependent metric calibration is performed using training data only, avoiding test-set information in metric construction.
We additionally report an empirical real-data reference, denoted ``Oracle'' in Tables, by comparing the test set against an equally sized subset of real training samples not used for metric calibration.

\paragraph{Implementation details.}
The main comparison reports mean $\pm$ sample standard deviation from three independently trained models (seeds $\{12345,12346,12347\}$) and their generated samples.
Bold indicates the lowest mean among generators, not a significance test.
Additional experimental and baseline details are provided in \Cref{app:exp_details,app:baseline_details}, with accompanying code available at \url{https://anonymous.4open.science/r/ChronoFlow}.

\begin{table*}[t]
\centering
\caption{
Generation fidelity across five irregular time series benchmarks.
We compare \method{} with four baselines using six complementary metrics covering overall sample realism, sampling structure, temporal dynamics, value fidelity, and cross-feature dependencies.
Oracle denotes a real--real reference. Bold indicates the lowest mean among generators for each dataset and metric.
}
\label{tab:main-fidelity}
\vspace{-.1in}
\small
\setlength{\tabcolsep}{4pt}
\newcommand{\gcell}[1]{\cellcolor{gray!20}#1}
\resizebox{\textwidth}{!}{%
\begin{tabular}{l|l|cccccc}
\toprule
Dataset & Model
& Set-Discr. $\downarrow$
& CoOb $\downarrow$
& Pattern-MMD$^2$ $\downarrow$
& Trans-MMD$^2$ $\downarrow$
& Value-W1 $\downarrow$
& Set-Corr. $\downarrow$ \\
\midrule

\multirow{6}{*}{eICU}
& Oracle
& 0.0019{\scriptsize$\pm$0.0012}
& 0.0306{\scriptsize$\pm$0.0155}
& 0.0000{\scriptsize$\pm$0.0000}
& 0.0004{\scriptsize$\pm$0.0001}
& 0.0312{\scriptsize$\pm$0.0007}
& 0.0364{\scriptsize$\pm$0.0015} \\
& Poisson + LatentODE
& 0.4242{\scriptsize$\pm$0.0317}
& 1.9710{\scriptsize$\pm$0.0017}
& 0.0047{\scriptsize$\pm$0.0024}
& 0.0249{\scriptsize$\pm$0.0038}
& 0.2216{\scriptsize$\pm$0.0321}
& 0.0643{\scriptsize$\pm$0.0049} \\
& LogNormMix + mTAN
& 0.4555{\scriptsize$\pm$0.0021}
& 0.0383{\scriptsize$\pm$0.0141}
& 0.0011{\scriptsize$\pm$0.0015}
& 0.0169{\scriptsize$\pm$0.0025}
& 0.2910{\scriptsize$\pm$0.0319}
& 0.0813{\scriptsize$\pm$0.0060} \\
& LogNormMix + TFM
& 0.4518{\scriptsize$\pm$0.0122}
& \textbf{0.0374{\scriptsize$\pm$0.0042}}
& 0.0014{\scriptsize$\pm$0.0019}
& 0.0247{\scriptsize$\pm$0.0071}
& 0.3585{\scriptsize$\pm$0.0321}
& 0.1435{\scriptsize$\pm$0.0071} \\
& FlexTPP
& 0.3629{\scriptsize$\pm$0.0288}
& 0.1336{\scriptsize$\pm$0.1033}
& 0.0032{\scriptsize$\pm$0.0014}
& 0.0070{\scriptsize$\pm$0.0030}
& 0.1081{\scriptsize$\pm$0.0396}
& 0.0532{\scriptsize$\pm$0.0112} \\
& \gcell{ChronoFlow}
& \gcell{\textbf{0.0995{\scriptsize$\pm$0.0098}}}
& \gcell{0.0380{\scriptsize$\pm$0.0047}}
& \gcell{\textbf{0.0002{\scriptsize$\pm$0.0001}}}
& \gcell{\textbf{0.0045{\scriptsize$\pm$0.0002}}}
& \gcell{\textbf{0.0504{\scriptsize$\pm$0.0047}}}
& \gcell{\textbf{0.0385{\scriptsize$\pm$0.0003}}} \\

\midrule

\multirow{6}{*}{P12}
& Oracle
& 0.0070{\scriptsize$\pm$0.0042}
& 0.0740{\scriptsize$\pm$0.0078}
& 0.0005{\scriptsize$\pm$0.0002}
& 0.0012{\scriptsize$\pm$0.0008}
& 0.0387{\scriptsize$\pm$0.0014}
& 0.0447{\scriptsize$\pm$0.0013} \\
& Poisson + LatentODE
& 0.4772{\scriptsize$\pm$0.0042} 
& 1.8353{\scriptsize$\pm$0.0001}
& 0.0066{\scriptsize$\pm$0.0011} 
& 0.0450{\scriptsize$\pm$0.0017} 
& 0.2326{\scriptsize$\pm$0.0059} 
& 0.0889{\scriptsize$\pm$0.0045} \\
& LogNormMix + mTAN
& 0.4819{\scriptsize$\pm$0.0095}
& 0.1225{\scriptsize$\pm$0.0517}
& 0.0035{\scriptsize$\pm$0.0026}
& 0.0280{\scriptsize$\pm$0.0096}
& 0.3297{\scriptsize$\pm$0.0422}
& 0.1047{\scriptsize$\pm$0.0336} \\
& LogNormMix + TFM
& 0.4816{\scriptsize$\pm$0.0042}
& 0.1089{\scriptsize$\pm$0.0575}
& 0.0028{\scriptsize$\pm$0.0025}
& 0.0367{\scriptsize$\pm$0.0013}
& 0.4316{\scriptsize$\pm$0.0074}
& 0.1140{\scriptsize$\pm$0.0076} \\
& FlexTPP
& \textbf{0.2018{\scriptsize$\pm$0.0592}}
& 0.0919{\scriptsize$\pm$0.0311}
& 0.0093{\scriptsize$\pm$0.0014}
& \textbf{0.0048{\scriptsize$\pm$0.0014}}
& \textbf{0.0904{\scriptsize$\pm$0.0038}}
& 0.0559{\scriptsize$\pm$0.0016} \\
& \gcell{ChronoFlow}
& \gcell{0.2154{\scriptsize$\pm$0.0094}}
& \gcell{\textbf{0.0889{\scriptsize$\pm$0.0114}}}
& \gcell{\textbf{0.0018{\scriptsize$\pm$0.0003}}}
& \gcell{0.0191{\scriptsize$\pm$0.0033}}
& \gcell{0.1012{\scriptsize$\pm$0.0038}}
& \gcell{\textbf{0.0521{\scriptsize$\pm$0.0015}}} \\
\midrule

\multirow{6}{*}{MIMIC-III}
& Oracle
& 0.0130{\scriptsize$\pm$0.0051}
& 0.0151{\scriptsize$\pm$0.0063}
& 0.0001{\scriptsize$\pm$0.0001}
& 0.0002{\scriptsize$\pm$0.0001}
& 0.0274{\scriptsize$\pm$0.0055}
& 0.0233{\scriptsize$\pm$0.0010} \\
& Poisson + LatentODE
& 0.4918{\scriptsize$\pm$0.0020} 
& 2.8069{\scriptsize$\pm$0.0001} 
& 0.0047{\scriptsize$\pm$0.0020} 
& 0.0390{\scriptsize$\pm$0.0012} 
& 0.1990{\scriptsize$\pm$0.0055} 
& 0.0556{\scriptsize$\pm$0.0018} \\
& LogNormMix + mTAN
& 0.4946{\scriptsize$\pm$0.0033}
& 0.0662{\scriptsize$\pm$0.0277}
& 0.0042{\scriptsize$\pm$0.0028}
& 0.0219{\scriptsize$\pm$0.0030}
& 0.3109{\scriptsize$\pm$0.0182}
& 0.0774{\scriptsize$\pm$0.0116} \\
& LogNormMix + TFM
& 0.4924{\scriptsize$\pm$0.0008}
& 0.0685{\scriptsize$\pm$0.0339}
& 0.0041{\scriptsize$\pm$0.0022}
& 0.0557{\scriptsize$\pm$0.0203}
& 0.5011{\scriptsize$\pm$0.0662}
& 0.1280{\scriptsize$\pm$0.0209} \\
& FlexTPP
& 0.4610{\scriptsize$\pm$0.0040}
& 0.0722{\scriptsize$\pm$0.0189}
& 0.0212{\scriptsize$\pm$0.0163}
& \textbf{0.0058{\scriptsize$\pm$0.0024}}
& 0.0918{\scriptsize$\pm$0.0180}
& \textbf{0.0401{\scriptsize$\pm$0.0028}} \\
& \gcell{ChronoFlow}
& \gcell{\textbf{0.0587{\scriptsize$\pm$0.0205}}}
& \gcell{\textbf{0.0440{\scriptsize$\pm$0.0098}}}
& \gcell{\textbf{0.0028{\scriptsize$\pm$0.0004}}}
& \gcell{0.0083{\scriptsize$\pm$0.0007}}
& \gcell{\textbf{0.0816{\scriptsize$\pm$0.0033}}}
& \gcell{0.0403{\scriptsize$\pm$0.0057}} \\
\midrule

\multirow{6}{*}{MIMIC-IV}
& Oracle
& 0.0057{\scriptsize$\pm$0.0039}
& 0.0144{\scriptsize$\pm$0.0023}
& 0.0000{\scriptsize$\pm$0.0000}
& 0.0001{\scriptsize$\pm$0.0000}
& 0.0215{\scriptsize$\pm$0.0006}
& 0.0217{\scriptsize$\pm$0.0009} \\
& Poisson + LatentODE
& 0.4960{\scriptsize$\pm$0.0020} 
& 1.9999{\scriptsize$\pm$0.0002} 
& 0.0070{\scriptsize$\pm$0.0025} 
& 0.0407{\scriptsize$\pm$0.0005} 
& 0.2081{\scriptsize$\pm$0.0125} 
& 0.0518{\scriptsize$\pm$0.0008} \\
& LogNormMix + mTAN
& 0.4956{\scriptsize$\pm$0.0034} 
& 0.0757{\scriptsize$\pm$0.0394} 
& \textbf{0.0017{\scriptsize$\pm$0.0007}} 
& 0.0653{\scriptsize$\pm$0.0660}  
& 0.4814{\scriptsize$\pm$0.2661} 
& 0.1326{\scriptsize$\pm$0.0664} \\
& LogNormMix + TFM
& 0.4963{\scriptsize$\pm$0.0031}  
& 0.0836{\scriptsize$\pm$0.0360}   
& 0.0019{\scriptsize$\pm$0.0006} 
& 0.0558{\scriptsize$\pm$0.0116}  
& 0.5133{\scriptsize$\pm$0.0744} 
& 0.1318{\scriptsize$\pm$0.0293} \\
& FlexTPP
& 0.4728{\scriptsize$\pm$0.0179} 
& \textbf{0.0379{\scriptsize$\pm$0.0117}} 
& 0.0150{\scriptsize$\pm$0.0009} 
& 0.0341{\scriptsize$\pm$0.0007} 
& 0.0719{\scriptsize$\pm$0.0057} 
& 0.0329{\scriptsize$\pm$0.0014} \\
& \gcell{ChronoFlow}
& \gcell{\textbf{0.2644{\scriptsize$\pm$0.0453}}}
& \gcell{0.1094{\scriptsize$\pm$0.0163}}
& \gcell{0.0038{\scriptsize$\pm$0.0008}}
& \gcell{\textbf{0.0084{\scriptsize$\pm$0.0003}}}
& \gcell{\textbf{0.0660{\scriptsize$\pm$0.0041}}}
& \gcell{\textbf{0.0311{\scriptsize$\pm$0.0008}}} \\
\midrule

\multirow{6}{*}{Activity}
& Oracle
& 0.1997{\scriptsize$\pm$0.0384}
& 0.0730{\scriptsize$\pm$0.0026}
& 0.0307{\scriptsize$\pm$0.0028}
& 0.0092{\scriptsize$\pm$0.0007}
& 0.0916{\scriptsize$\pm$0.0064}
& 0.0673{\scriptsize$\pm$0.0078}
\\
& Poisson + LatentODE
& 0.3767{\scriptsize$\pm$0.0210}
& 0.8366{\scriptsize$\pm$0.0004}
& 0.1194{\scriptsize$\pm$0.0038}
& 0.1929{\scriptsize$\pm$0.0020}
& 0.2409{\scriptsize$\pm$0.0150}
& 0.0949{\scriptsize$\pm$0.0031} \\
& LogNormMix + mTAN
& 0.3698{\scriptsize$\pm$0.0290}
& 0.0758{\scriptsize$\pm$0.0038}
& 0.0339{\scriptsize$\pm$0.0029}
& 0.0243{\scriptsize$\pm$0.0014}
& 0.2586{\scriptsize$\pm$0.0137}
& 0.1088{\scriptsize$\pm$0.0345} \\
& LogNormMix + TFM
& 0.4601{\scriptsize$\pm$0.0210}
& 0.0710{\scriptsize$\pm$0.0111}
& 0.0294{\scriptsize$\pm$0.0082}
& 0.0328{\scriptsize$\pm$0.0041}
& 0.3187{\scriptsize$\pm$0.0253}
& 0.1596{\scriptsize$\pm$0.0323} \\
& FlexTPP
& 0.1997{\scriptsize$\pm$0.0484}
& \textbf{0.0598{\scriptsize$\pm$0.0144}}
& 0.0319{\scriptsize$\pm$0.0049}
& 0.0127{\scriptsize$\pm$0.0073}
& 0.1125{\scriptsize$\pm$0.0206}
& 0.0739{\scriptsize$\pm$0.0093} \\
& \gcell{ChronoFlow}
& \gcell{\textbf{0.1753{\scriptsize$\pm$0.0217}}}
& \gcell{0.0739{\scriptsize$\pm$0.0100}}
& \gcell{\textbf{0.0278{\scriptsize$\pm$0.0080}}}
& \gcell{\textbf{0.0100{\scriptsize$\pm$0.0015}}}
& \gcell{\textbf{0.1018{\scriptsize$\pm$0.0068}}}
& \gcell{\textbf{0.0474{\scriptsize$\pm$0.0067}}} \\
\bottomrule
\end{tabular}%
}
\end{table*}

\begin{table*}[t]
\centering
\caption{
Ablation study of \method{} on eICU.
We compare alternative generation factorizations and evaluate the effect of optimal-transport coupling in PatternFlow.
$M$ is the number of observation occasions and $\boldsymbol r$ contains feature-wise observation frequencies.
$\boldsymbol\tau$ denotes observation times, $\boldsymbol B$ feature panels, and $\mathbf x$ feature values.
}
\label{tab:fidelity-ablation-eicu}
\vspace{-.1in}
\small
\setlength{\tabcolsep}{4pt}
\resizebox{\textwidth}{!}{%
\begin{tabular}{l|l|c|cccccc}
\toprule
Model & Process & Size & Set-Discr. $\downarrow$ & CoOb $\downarrow$ & Pattern-MMD$^2$ $\downarrow$ & Trans-MMD$^2$ $\downarrow$ & Value-W1 $\downarrow$ & Set-Corr. $\downarrow$ \\
\midrule
Oracle
& --
& --
& 0.0019{\scriptsize$\pm$0.0012}
& 0.0306{\scriptsize$\pm$0.0155}
& 0.0000{\scriptsize$\pm$0.0000}
& 0.0004{\scriptsize$\pm$0.0001}
& 0.0312{\scriptsize$\pm$0.0007}
& 0.0364{\scriptsize$\pm$0.0015} \\
\midrule

Value$\to$Pattern
& $M\to\mathbf x\to(\boldsymbol\tau,\boldsymbol B)$
& 2.70M
& 0.4511{\scriptsize$\pm$0.0217}
& 0.2354{\scriptsize$\pm$0.0098}
& 0.0051{\scriptsize$\pm$0.0001}
& 0.0099{\scriptsize$\pm$0.0003}
& 0.1142{\scriptsize$\pm$0.0022}
& 0.0461{\scriptsize$\pm$0.0046} \\
Joint TPV
& $M\to(\boldsymbol\tau,\boldsymbol B,\mathbf x)$
& 1.12M
& 0.2639{\scriptsize$\pm$0.0039}
& 0.0867{\scriptsize$\pm$0.0033}
& 0.0008{\scriptsize$\pm$0.0002}
& 0.0275{\scriptsize$\pm$0.0013}
& 0.3179{\scriptsize$\pm$0.0044}
& 0.0698{\scriptsize$\pm$0.0101} \\
Joint All
& $(M,\boldsymbol\tau,\boldsymbol B,\mathbf x)$
& 2.21M
& 0.2445{\scriptsize$\pm$0.1387}
& 0.0731{\scriptsize$\pm$0.0055}
& 0.0018{\scriptsize$\pm$0.0002}
& 0.0074{\scriptsize$\pm$0.0009}
& 0.0779{\scriptsize$\pm$0.0011}
& 0.0423{\scriptsize$\pm$0.0012} \\
\midrule
w/o OT
& $(M,\boldsymbol r)\to(\boldsymbol\tau,\boldsymbol B)\to\mathbf x$
& 2.06M
& 0.4316{\scriptsize$\pm$0.0085}
& 0.0854{\scriptsize$\pm$0.0042}
& 0.0382{\scriptsize$\pm$0.0031}
& 0.0083{\scriptsize$\pm$0.0004}
& 0.0731{\scriptsize$\pm$0.0038}
& \textbf{0.0382{\scriptsize$\pm$0.0037}} \\
\midrule
\rowcolor{gray!20}
ChronoFlow
& $(M,\boldsymbol r)\to(\boldsymbol\tau,\boldsymbol B)\to\mathbf x$
& 2.06M
& \textbf{0.0995{\scriptsize$\pm$0.0098}}
& \textbf{0.0380{\scriptsize$\pm$0.0047}}
& \textbf{0.0002{\scriptsize$\pm$0.0001}}
& \textbf{0.0045{\scriptsize$\pm$0.0002}}
& \textbf{0.0504{\scriptsize$\pm$0.0047}}
& 0.0385{\scriptsize$\pm$0.0003} \\
\bottomrule
\end{tabular}%
}
\vspace{-.1in}
\end{table*}

\subsection{Main Results}

\paragraph{Quantitative results.}
To address \textbf{RQ1}, we compare generation fidelity across five benchmarks using six metrics. In \Cref{tab:main-fidelity}, \method{} achieves the lowest mean in $21$ of $30$ dataset--metric comparisons. It ranks first on five metrics each for eICU and Activity, four for MIMIC-III and MIMIC-IV, and three for P12. It also achieves the lowest mean on four of five benchmarks for Set-Discr, Pattern-MMD$^2$, Value-W1, and Set-Corr.
The largest Set-Discr gains are on eICU and MIMIC-III, decreasing from $0.3629$ to $0.0995$ and from $0.4610$ to $0.0587$, respectively. On eICU, Value-W1 decreases by $53.4\%$.
Performance varies across metrics and datasets. On MIMIC-IV, \method{} ranks first on four metrics but has larger CoOb and Pattern-MMD$^2$ discrepancies, illustrating that the metrics capture complementary aspects of fidelity. FlexTPP also achieves lower Trans-MMD$^2$ on P12 and MIMIC-III and lower CoOb on Activity.
On Activity, the splits are constructed by subject rather than i.i.d., so the real--real reference compares windows from different subjects and reflects cross-subject shift rather than a lower bound; generator scores can therefore fall below it.

\begin{figure*}[t]
\centering
\includegraphics[width=\textwidth]{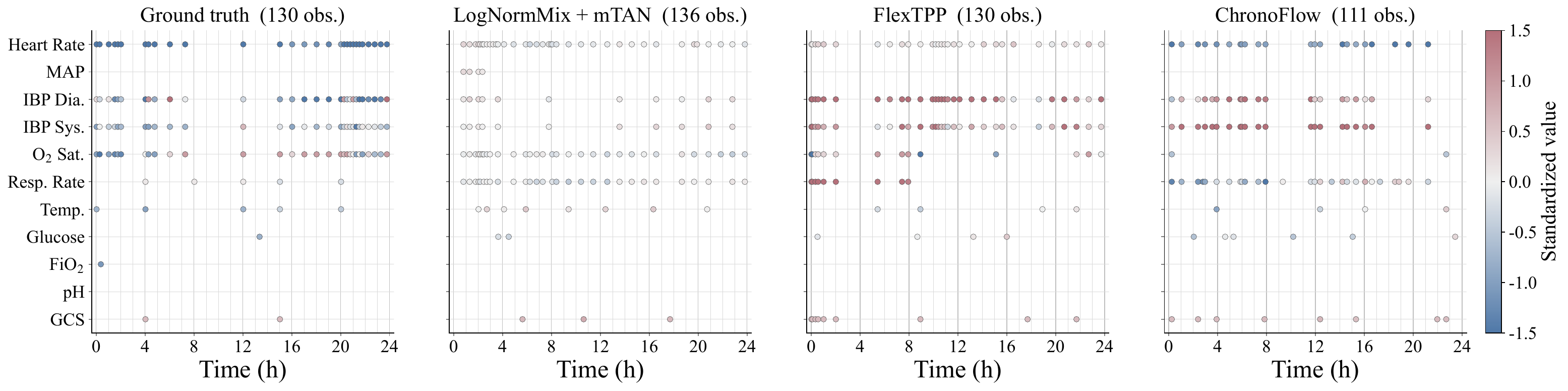}\par
\includegraphics[width=\textwidth]{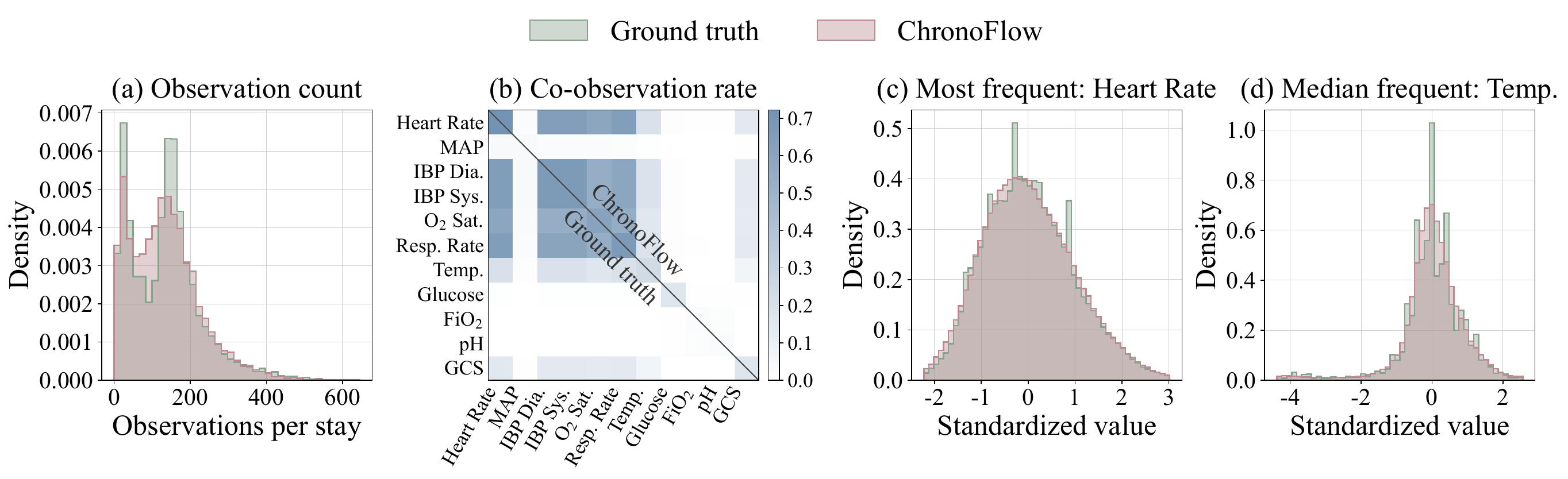}
\vspace{-.3in}
\caption{
Qualitative analysis on eICU.
Top: real and generated samples from LogNormMix + mTAN, FlexTPP, and \method{}.
Bottom: ground-truth versus \method{} distributions of observation counts, feature co-observation, and Heart Rate and Temperature values.
}
\label{fig:qualitative-analysis}
\vspace{-0.2in}
\end{figure*}

\paragraph{Qualitative results.}
\Cref{fig:qualitative-analysis} further shows that \method{} closely reproduces key properties of eICU, including the observation-count distribution, co-observation structure, and value marginals.
In particular, it preserves frequently co-observed feature pairs such as IBP Dia.\ and IBP Sys., while generated samples exhibit grouped observations, irregular temporal gaps, and feature-specific sparsity similar to real data.
Additional aggregate distributions, individual samples, and UMAP visualizations across all five datasets are provided in \Cref{app:visualization}.

\subsection{Ablation Study}

\paragraph{Generation factorization.}
To address \textbf{RQ2}, we compare generation factorizations and optimal-transport coupling on eICU (\Cref{tab:fidelity-ablation-eicu}).
Compared with joint and reverse-order factorizations, ChronoFlow achieves lower means on all six metrics.
Reversing the generation order raises Set-Discr.\ from $0.0995$ to $0.4511$ and CoOb from $0.0380$ to $0.2354$, despite using more parameters.
Joint All has higher errors across all metrics, while Joint TPV increases Value-W1 from $0.0504$ to $0.3179$.
On eICU, these configurations favor conditioning values on finalized patterns, but differences in capacity and conditioning prevent isolating generation order alone.

\paragraph{Optimal-transport coupling.}
Removing OT coupling from PatternFlow without changing the architecture increases Pattern-MMD$^2$ from $0.0002$ to $0.0382$ and Set-Discr. from $0.0995$ to $0.4316$. CoOb, Trans-MMD$^2$, and Value-W1 also worsen, whereas Set-Corr.\ changes little.
These eICU results support the evaluated temporal-coupling configuration.

\subsection{Error Propagation Analysis}

\paragraph{Oracle intervention.}
To examine the robustness aspect of \textbf{RQ4}, we replace generated upstream variables with their ground-truth counterparts at inference (\Cref{tab:error-propagation}).
On eICU, providing ground-truth global observation statistics $(M,\boldsymbol r)$ reduces Set-Discr.\ from $0.0995$ to $0.0153$ and CoOb from $0.0380$ to $0.0128$.
This suggests that IntensityFlow errors contribute to residual co-observation discrepancies, while PatternFlow better reproduces co-observation under ground-truth conditioning.
Providing the ground-truth sampling pattern $(\boldsymbol\tau,\boldsymbol B)$ further brings temporal-transition fidelity close to the empirical real--real reference.
Set-Discr.\ remains highly variable and Value-W1 changes little, indicating metric-dependent effects rather than uniform improvement (\Cref{app:intervention-scope}).

\paragraph{Perturbation sensitivity.}
We perturb ground-truth upstream variables to examine sensitivity to upstream errors.
Increasing the severity of perturbations to global observation statistics worsens CoOb and Trans-MMD$^2$, while also perturbing the sampling pattern leads to larger CoOb discrepancies.
In contrast, Value-W1 remains between $0.0460$ and $0.0600$ across all settings.
Together, these eICU results identify IntensityFlow as a promising improvement target while showing that value marginals remain relatively stable under the tested upstream perturbations.

\subsection{Further Evaluation}

\begin{figure*}[t]
    \centering
    \includegraphics[height=0.2675\textwidth]{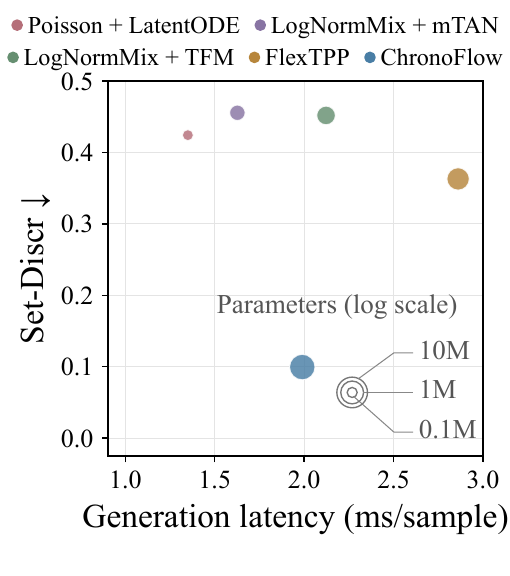}%
    \includegraphics[height=0.2675\textwidth]{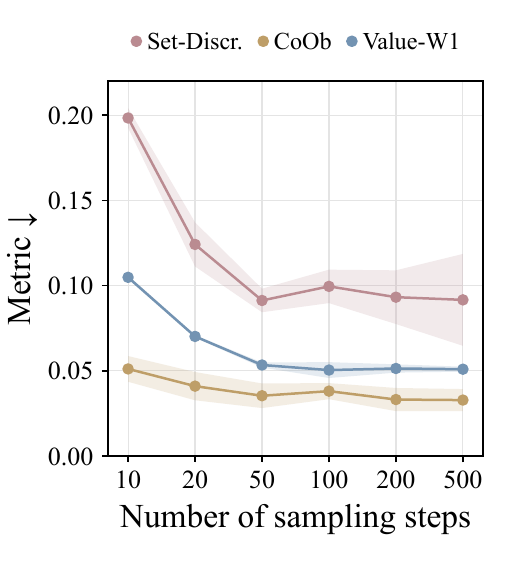}%
    \includegraphics[height=0.2675\textwidth]{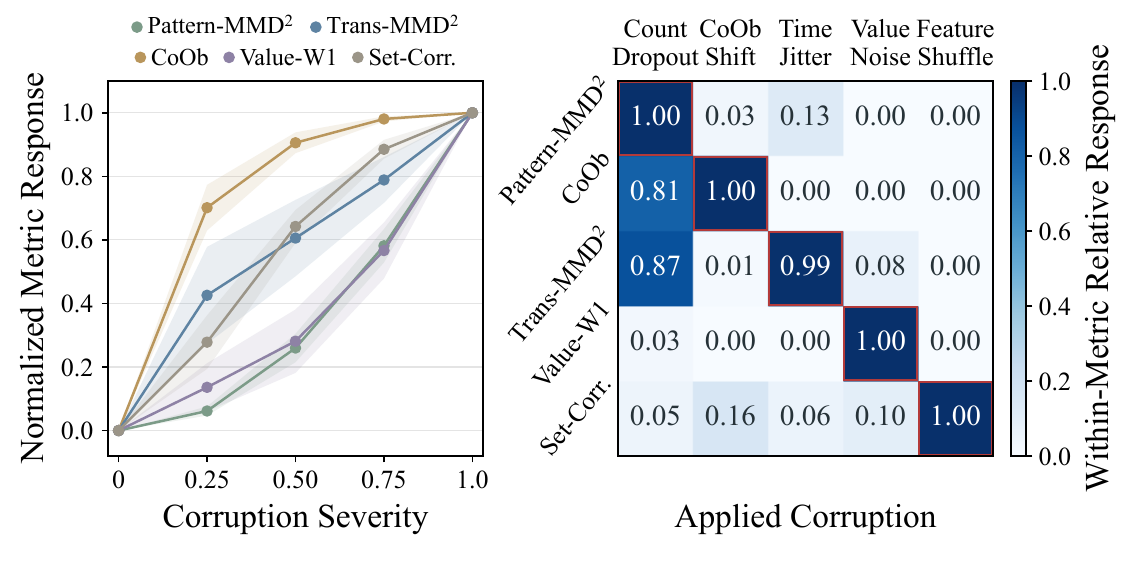}%
    \par\noindent
    \makebox[.237\textwidth]{\small(a)}%
    \makebox[.237\textwidth]{\small(b)}%
    \makebox[.237\textwidth]{\small(c)}%
    \makebox[.288\textwidth]{\small(d)}\par

    \vspace{-.05in}
    \caption{
    Further analysis. (a,b) eICU latency--fidelity and ODE-step sensitivity, with bubble size indicating parameter count in (a). (c) Designated-corruption responses normalized by endpoint change within each dataset, then averaged across five datasets. Shading is between-dataset sample standard deviation. (d) Relative positive responses across corruptions, averaged over available datasets. Definitions and missing-support exceptions are in \Cref{app:controlled-corruptions}.
    }
    \label{fig:efficiency-sensitivity-validation}
    \vspace{-.15in}
\end{figure*}

\paragraph{Computational efficiency.}
As part of \textbf{RQ4}, \Cref{fig:efficiency-sensitivity-validation} compares end-to-end sampling latency and Set-Discr.\ on eICU, with bubble size indicating parameter count.
In the plotted eICU configuration, \method{} has lower Set-Discr.\ and latency than FlexTPP and LogNormMix + TFM.
Poisson + LatentODE and LogNormMix + mTAN are faster but more distinguishable.

\paragraph{Sampling-step sensitivity.}
\Cref{fig:efficiency-sensitivity-validation} also varies the number of ODE integration steps with the trained model fixed.
Increasing the budget from $10$ to $50$ steps reduces Set-Discr.\ from $0.1983$ to $0.0912$ and Value-W1 from $0.1048$ to $0.0534$.
Across $50$--$500$ steps, both metrics change little, and CoOb shows similarly small variation.
The default of $100$ steps therefore lies in a region of diminishing returns, while $50$ steps offers a smaller integration budget with similar fidelity.

\paragraph{Metric validation.}
To address \textbf{RQ3}, \Cref{fig:efficiency-sensitivity-validation}(c,d) evaluates five direct metrics across five datasets using count dropout, co-observation shift, time jitter, value noise, and cross-feature shuffle.
Dataset-averaged normalized responses increase with designated corruption severity.
Value-W1 and Set-Corr.\ mainly respond to their designated corruptions, while count dropout also affects CoOb and Trans-MMD$^2$.
These are relative responses to selected changes, not absolute calibration or exclusive sensitivity. Operators, normalization, and valid-repetition counts are specified in \Cref{app:controlled-corruptions}.

\section{Conclusion}
We have formulated complete irregular samples through observation occasions and feature panels, and proposed \method{} for their hierarchical generation.
By organizing generation by statistical granularity, \method{} decomposes the heterogeneous joint problem into structured subproblems while preserving their dependencies.
Our evaluation protocol combines complementary diagnostics of sampling structure and value fidelity, with sensitivity assessed through controlled corruptions.
Across five benchmarks, \method{} achieves the lowest mean among generators in most dataset--metric comparisons, while factorization studies on eICU support the proposed hierarchy.
These results support structured generation within the tested settings, with remaining limitations discussed in \Cref{app:limitations}.
\section*{AI Use Statement}
Generative AI tools were used to assist with implementing research ideas, verifying theoretical arguments and proofs, designing and implementing experiments, and polishing the manuscript.
All AI-assisted outputs were independently reviewed and verified by the authors, who take full responsibility for the final content of this work.

\section*{Ethics Statement}
We use four de-identified EHR benchmarks: eICU~\citep{pollard2018eicu}, Physionet-2012~\citep{citi2012physionet}, MIMIC-III~\citep{johnson2016mimic}, and MIMIC-IV~\citep{johnson2023mimic}.
MIMIC-III, MIMIC-IV, and eICU are handled under the PhysioNet Credentialed Health Data Use Agreement.
Although de-identified, these datasets contain sensitive health information, and generated samples may inherit biases or sensitive patterns from the source data.
They should therefore not be assumed to provide formal privacy guarantees or to support clinical decision-making without further validation.

\section*{Reproducibility Statement}
To facilitate reproducibility, we release the complete implementation of \method{}, including data preprocessing, model configurations, training and generation procedures, and evaluation code.
Key experimental settings are provided in the paper and appendix, and complete architecture and training configurations are included in the accompanying code repository.
Our code is available at \url{https://anonymous.4open.science/r/ChronoFlow}.

\bibliography{appendix/reference}
\bibliographystyle{iclr2027_conference}

\clearpage
\appendix
\renewcommand{\thefigure}{A\arabic{figure}} 
\renewcommand{\thetable}{A\arabic{table}} 
\renewcommand{\thetheorem}{A\arabic{theorem}} 

\section{Limitations and Future Work}
\label{app:limitations}

\paragraph{Discrete structure and conditioning.}
IntensityFlow uses continuous relaxations followed by rounding and frequency projection, and PatternFlow thresholds panels.
This may be suboptimal for discrete distributions and does not enforce agreement between sampled frequencies and decoded panels.
Discrete flow matching~\citep{gat2024discrete} and consistency-preserving decoding are promising extensions.
Downstream stages are trained with ground-truth conditions but receive generated conditions at inference, leaving a conditioning shift that the eICU interventions diagnose without resolving.

\paragraph{Resolution and value support.}
Benchmark timestamps have finite resolutions and recurring intervals.
Adaptive temporal codebooks or hybrid discrete--continuous models could exploit this structure while allowing off-grid generation.
Likewise, continuous value generation does not explicitly enforce ordinal categories or other feature-specific support constraints.

\paragraph{Scalability and evaluation scope.}
Self-attention in PatternFlow and ValueFlow scales quadratically with occasion and measurement counts.
Efficient architectures and evaluation on longer, denser records remain necessary.
Our evidence concerns fidelity on the selected benchmarks and adapters, with factorization ablations limited to eICU.
It does not establish downstream utility, privacy, or absence of memorization, and does not isolate all sampling--value dependencies.
Direct comparisons with grid-and-mask generators, sequential time--panel alternatives, and broader ablations remain future work.

\section{Algorithms}

\definecolor{algorithmgray}{gray}{0.45}

\algrenewcommand\algorithmiccomment[1]{%
    \hfill\textcolor{algorithmgray}{\(\triangleright\) #1}%
}

\begin{algorithm}[t]
\caption{Training procedure for \method{}.}
\label{alg:chronoflow-training}
\begin{algorithmic}[1]
\Require Preprocessed training set
$\mathcal{D}=\{\mathcal{S}^{(i)}\}_{i=1}^{N_{\mathrm{train}}}$,
where $M\le M_{\max}$
\Require Parameters $\theta_I,\theta_P,\theta_V$

\For{each optimization step}
    \State Sample
    $\mathcal{S}=(\boldsymbol{\tau},\boldsymbol{B},\mathbf{x})
    \sim\mathcal{D}$
    \State
    $M\gets|\boldsymbol{\tau}|$,
    $\displaystyle
    r_f\gets\frac{1}{M}\sum_{m=1}^{M}\boldsymbol{B}_{m,f}$

    \Statex \vspace{0.15em}
    \State \textcolor{algorithmgray}{/* IntensityFlow */}

    \State
    $\boldsymbol{u}_1
    \gets
    [\,2M/M_{\max}-1,\;2\boldsymbol{r}-1\,]$
    \State Sample
    $\boldsymbol{u}_0\sim\mathcal{N}(0,I_{1+F})$,
    $t_I\sim\mathcal{U}(0,1)$
    \State
    $\boldsymbol{u}_{t_I}
    \gets
    (1-t_I)\boldsymbol{u}_0+t_I\boldsymbol{u}_1$
    \State
    $\displaystyle
    \mathcal{L}_I
    \gets
    \left\|
    v_{\theta_I}(\boldsymbol{u}_{t_I},t_I)
    -
    (\boldsymbol{u}_1-\boldsymbol{u}_0)
    \right\|_2^2$

    \Statex \vspace{0.15em}
    \State \textcolor{algorithmgray}{/* PatternFlow */}

    \State
    $\boldsymbol{\tau}_1\gets2\boldsymbol{\tau}-1$,
    $\boldsymbol{B}_1\gets2\boldsymbol{B}-1$
    \State Sample
    $\boldsymbol{\tau}_0\sim\mathcal{N}(0,I_M)$ and
    $\boldsymbol{B}_0$ with i.i.d.\ $\mathcal{N}(0,1)$ entries
    \State
    $(\boldsymbol{\tau}_0,\boldsymbol{B}_0)
    \gets
    \mathrm{SortByTime}(\boldsymbol{\tau}_0,\boldsymbol{B}_0)$
    \State
    $(\boldsymbol{\tau}_1,\boldsymbol{B}_1)
    \gets
    \mathrm{SortByTime}(\boldsymbol{\tau}_1,\boldsymbol{B}_1)$
    \State Sample $t_P\sim\mathcal{U}(0,1)$ and set
    \[
    \boldsymbol{\tau}_{t_P}
    \gets
    (1-t_P)\boldsymbol{\tau}_0+t_P\boldsymbol{\tau}_1,
    \qquad
    \boldsymbol{B}_{t_P}
    \gets
    (1-t_P)\boldsymbol{B}_0+t_P\boldsymbol{B}_1
    \]
    \State
    $(v_{\theta_{P,\tau}},v_{\theta_{P,B}})
    \gets
    v_{\theta_P}
    (\boldsymbol{\tau}_{t_P},\boldsymbol{B}_{t_P},t_P
    \mid M,\boldsymbol{r})$
    \State
    \[
    \mathcal{L}_P
    \gets
    \left\|
    v_{\theta_{P,\tau}}
    -
    (\boldsymbol{\tau}_1-\boldsymbol{\tau}_0)
    \right\|_2^2
    +
    \frac{1}{F}
    \left\|
    v_{\theta_{P,B}}
    -
    (\boldsymbol{B}_1-\boldsymbol{B}_0)
    \right\|_2^2
    \]

    \Statex \vspace{0.15em}
    \State \textcolor{algorithmgray}{/* ValueFlow */}

    \State
    $\mathcal{I}
    \gets
    \{(m,f):\boldsymbol{B}_{m,f}=1\}
    =
    \{(m_i,f_i)\}_{i=1}^{N}$
    \State
    $\boldsymbol{f}\gets(f_i)_{i=1}^{N}$,
    $\boldsymbol{\tau}^{\mathrm{obs}}
    \gets(\tau_{m_i})_{i=1}^{N}$,
    $\mathbf{z}_1\gets(\mathbf{x}_{m_i,f_i})_{i=1}^{N}$
    \State Sample
    $\mathbf{z}_0\sim\mathcal{N}(0,I_N)$,
    $t_V\sim\mathcal{U}(0,1)$
    \State
    $\mathbf{z}_{t_V}
    \gets
    (1-t_V)\mathbf{z}_0+t_V\mathbf{z}_1$
    \State
    $\displaystyle
    \mathcal{L}_V
    \gets
    \left\|
    v_{\theta_V}
    (\boldsymbol{f},\boldsymbol{\tau}^{\mathrm{obs}},
    \mathbf{z}_{t_V},t_V)
    -
    (\mathbf{z}_1-\mathbf{z}_0)
    \right\|_2^2$

    \State Update
    $\theta_I,\theta_P,\theta_V$
    using $\mathcal{L}_I+\mathcal{L}_P+\mathcal{L}_V$
\EndFor
\end{algorithmic}
\end{algorithm}

\begin{algorithm}[t]
\caption{Sampling procedure for \method{}.}
\label{alg:chronoflow-inference}
\begin{algorithmic}[1]
\Require Trained flows $v_{\theta_I},v_{\theta_P},v_{\theta_V}$,
ODE solver $\Phi_{\mathrm{ODE}}$, number of samples $K$

\For{$k=1,\ldots,K$}

    \State \textcolor{algorithmgray}{/* IntensityFlow */}

    \State Sample
    $\boldsymbol{u}_0\sim\mathcal{N}(0,I_{1+F})$
    \State
    $\hat{\boldsymbol{u}}
    \gets
    \Phi_{\mathrm{ODE}}
    (v_{\theta_I},\boldsymbol{u}_0;0\!\rightarrow\!1)$
    \State Split
    $\hat{\boldsymbol{u}}
    =
    (\hat u_M,\hat{\boldsymbol{u}}_r)$
    \State
    $\displaystyle
    \hat M
    \gets
    \mathrm{clip}
    \left(
    \mathrm{round}
    \left[
    \frac{M_{\max}}{2}(\hat u_M+1)
    \right],
    1,M_{\max}
    \right)$
    \State
    $\displaystyle
    \hat{\boldsymbol{r}}
    \gets
    \mathrm{Proj}_{1/\hat M}
    \left(
    \mathrm{clip}
    \left[
    \frac{\hat{\boldsymbol{u}}_r+1}{2},
    0,1
    \right]
    \right)$

    \Statex \vspace{0.15em}
    \State \textcolor{algorithmgray}{/* PatternFlow */}

    \State Sample
    $\boldsymbol{\tau}_0\sim\mathcal{N}(0,I_{\hat M})$
    and $\boldsymbol{B}_0\in\mathbb{R}^{\hat M\times F}$
    with i.i.d.\ $\mathcal{N}(0,1)$ entries
    \State $\boldsymbol{\tau}_0\gets\operatorname{sort}(\boldsymbol{\tau}_0)$
    \State
    \[
    (\bar{\boldsymbol{\tau}},\bar{\boldsymbol{B}})
    \gets
    \Phi_{\mathrm{ODE}}
    \left(
    v_{\theta_P},
    (\boldsymbol{\tau}_0,\boldsymbol{B}_0);
    0\!\rightarrow\!1
    \mid
    \hat M,\hat{\boldsymbol r}
    \right)
    \]
    \State
    $\displaystyle
    \hat{\boldsymbol{\tau}}
    \gets
    \mathrm{clip}
    \left(
    \frac{\bar{\boldsymbol{\tau}}+1}{2},
    0,1
    \right)$,
    $\quad
    \hat{\boldsymbol{B}}
    \gets
    \mathbbm{1}[\bar{\boldsymbol{B}}>0]$
    \State For each empty row of $\hat{\boldsymbol{B}}$,
    activate the feature with the largest corresponding entry in
    $\bar{\boldsymbol{B}}$
    \State
    $(\hat{\boldsymbol{\tau}},\hat{\boldsymbol{B}})
    \gets
    \mathrm{SortByTime}
    (\hat{\boldsymbol{\tau}},\hat{\boldsymbol{B}})$

    \Statex \vspace{0.15em}
    \State \textcolor{algorithmgray}{/* ValueFlow */}

    \State
    $\hat{\mathcal{I}}
    \gets
    \{(m,f):\hat{\boldsymbol{B}}_{m,f}=1\}
    =
    \{(m_i,f_i)\}_{i=1}^{\hat N}$
    \State
    $\hat{\boldsymbol{f}}\gets(f_i)_{i=1}^{\hat N}$,
    $\hat{\boldsymbol{\tau}}^{\mathrm{obs}}
    \gets(\hat{\tau}_{m_i})_{i=1}^{\hat N}$
    \State Sample
    $\mathbf{z}_0\sim\mathcal{N}(0,I_{\hat N})$
    \State
    \[
    \hat{\mathbf{z}}
    \gets
    \Phi_{\mathrm{ODE}}
    \left(
    v_{\theta_V},
    \mathbf{z}_0;
    0\!\rightarrow\!1
    \mid
    \hat{\boldsymbol{f}},
    \hat{\boldsymbol{\tau}}^{\mathrm{obs}}
    \right)
    \]
    \State
    $\hat{\mathbf{x}}_{m_i,f_i}\gets\hat z_i$,
    \quad $i=1,\ldots,\hat N$

    \State
    $\hat{\mathcal{S}}^{(k)}
    \gets
    (\hat{\boldsymbol{\tau}},
    \hat{\boldsymbol{B}},
    \hat{\mathbf{x}})$

\EndFor
\end{algorithmic}
\end{algorithm}

This section summarizes the complete training and inference procedures of \method{}.
During training, the three flow matching models are optimized simultaneously, while each downstream stage is conditioned on the ground-truth outputs of the preceding stages.
At inference, generation proceeds sequentially through the hierarchy: IntensityFlow first generates the global observation statistics, PatternFlow then generates the corresponding sampling structure, and ValueFlow finally generates values on the realized observations.
\Cref{alg:chronoflow-training} and \Cref{alg:chronoflow-inference} provide the detailed training and generation procedures, respectively.

\section{Structural Properties and Proofs}
These standard properties justify the temporal coupling and event-token architecture. They are not new optimal-transport or equivariance results.

\subsection{Temporal Optimal-Transport Assignment}
\label{sec:proof_thm1}

\paragraph{Proof.}

Let $\tilde{\boldsymbol{\tau}}_0$ and
$\tilde{\boldsymbol{\tau}}_1$ denote the sorted versions of
$\boldsymbol{\tau}_0$ and $\boldsymbol{\tau}_1$.
It suffices to show that any assignment containing an inversion cannot improve the sorted-to-sorted matching.

Suppose $\pi\in\Pi_M$ contains an inversion: there exist $i<j$ such that
$\pi(i)>\pi(j)$.
Let
\[
u=(\tilde{\boldsymbol{\tau}}_0)_i,\qquad
v=(\tilde{\boldsymbol{\tau}}_0)_j,\qquad
s=(\tilde{\boldsymbol{\tau}}_1)_{\pi(j)},\qquad
w=(\tilde{\boldsymbol{\tau}}_1)_{\pi(i)}.
\]
Since both sequences are sorted, $u\le v$ and $s\le w$.
Swapping the two inverted assignments changes the cost by
\begin{align}
&
\big[(u-w)^2+(v-s)^2\big]
-
\big[(u-s)^2+(v-w)^2\big]
\nonumber\\
&=
2(v-u)(w-s)
\ge 0.
\end{align}
Thus, removing an inversion never increases the transport cost.
Repeatedly removing inversions yields the sorted-to-sorted assignment, and therefore
\begin{equation}
\min_{\pi\in\Pi_M}
\sum_{m=1}^{M}
\left|
(\boldsymbol{\tau}_0)_m
-
(\boldsymbol{\tau}_1)_{\pi(m)}
\right|^2
=
\sum_{m=1}^{M}
\left|
(\tilde{\boldsymbol{\tau}}_0)_m
-
(\tilde{\boldsymbol{\tau}}_1)_m
\right|^2.
\end{equation}
When ties occur, the minimizer need not be unique, but the minimum cost remains unchanged.
This proves \Cref{thm:ot-patternflow}.

\subsection{Permutation Equivariance of ValueFlow}
\label{sec:proof_thm2}

\paragraph{Proof.}

Let $P\in\{0,1\}^{N\times N}$ permute the $N$ event tokens.
Because ValueFlow uses shared token embeddings without index-dependent positional encodings, its tokenization map $\Psi$ satisfies
\begin{equation}
\Psi
\left(
P\boldsymbol{f},
P\boldsymbol{\tau}^{\mathrm{obs}},
P\mathbf{z}_t,
t
\right)
=
P\,
\Psi
\left(
\boldsymbol{f},
\boldsymbol{\tau}^{\mathrm{obs}},
\mathbf{z}_t,
t
\right).
\end{equation}

Let $H\in\mathbb{R}^{N\times d}$ denote the event tokens and
$G\in\mathbb{R}^{R\times d}$ the $R$ register tokens.
For
\[
\bar H=
\begin{bmatrix}
H\\
G
\end{bmatrix},
\qquad
\bar P=
\begin{bmatrix}
P&0\\
0&I_R
\end{bmatrix},
\]
permuting the event tokens corresponds to $\bar H'=\bar P\bar H$.

For a self-attention head,
\[
Q=\bar H W_Q,\qquad
K=\bar H W_K,\qquad
V=\bar H W_V.
\]
Under $\bar P$, we have
$Q'=\bar P Q$, $K'=\bar P K$, and $V'=\bar P V$.
Using row-wise softmax,
\begin{align}
\operatorname{Attn}(Q',K',V')
&=
\operatorname{softmax}
\left(
\frac{\bar P QK^\top\bar P^\top}{\sqrt{d_h}}
\right)
\bar P V
\nonumber\\
&=
\bar P\,
\operatorname{Attn}(Q,K,V).
\end{align}
Thus, self-attention is permutation equivariant with respect to the event tokens while leaving the register tokens fixed.

Multi-head attention, shared token-wise feed-forward layers, normalization, output projections, and residual connections preserve this equivariance.
By composition, the ValueFlow backbone is equivariant under $\bar P$.
Restricting the output to the $N$ event tokens gives
\begin{equation}
v_{\theta_V}
\left(
P\boldsymbol{f},
P\boldsymbol{\tau}^{\mathrm{obs}},
P\mathbf{z}_t,
t
\right)
=
P\,
v_{\theta_V}
\left(
\boldsymbol{f},
\boldsymbol{\tau}^{\mathrm{obs}},
\mathbf{z}_t,
t
\right),
\end{equation}
which proves \Cref{thm:equiv-valueflow}.
\section{Details of Evaluation Metrics}
\label{app:metric_details}

This section provides detailed definitions of the six metrics introduced in
\Cref{sec:evaluation}.
We first specify the common evaluation protocol and MMD estimator, and then
describe full-sample distinguishability, sampling-pattern fidelity, and value
fidelity and dependencies.

\subsection{Evaluation Representation and Protocol}
\label{app:metric-protocol}

Let $\mathcal D_R$ and $\mathcal D_G$ denote the real and generated
evaluation collections.
Following \Cref{sec:method}, each sample is represented as
$\mathcal S=(\boldsymbol{\tau},\boldsymbol B,\mathbf x)$.
For a benchmark with $T$ temporal positions, let $q_m\in\{0,\ldots,T-1\}$
denote the time index corresponding to occasion $m$, such that
$\tau_m=q_m/\max(T-1,1)$.
These are evaluation coordinates at the benchmark resolution. They do not imply that the continuous PatternFlow trajectory itself is constrained to this grid, and the reported scores do not establish fidelity below that resolution.

All metrics use the benchmark's preprocessed observation windows and masks.
Observed entries are identified by $\boldsymbol B$, so an observed zero remains
a valid measurement.
Value-based metrics use the training normalization, with additional
transition-specific scaling described below.

\paragraph{Reference and generated samples.}
The full test split serves as the real reference, and we generate the same
number of samples.
Set-Discr uses a separate classifier-training/evaluation split within these
collections.

\paragraph{Training-based calibration.}
A seeded subset of up to $1{,}024$ training samples is used to determine
MMD bandwidths, transition scales, and eligible features.
These quantities are fixed across generator comparisons, with calibration
seed $12345$.

\paragraph{Real--real reference.}
For real--real comparison, we draw an equally sized training subset excluding
samples used for metric calibration.
These scores provide an empirical reference for finite-sample variation and
train--test differences, rather than theoretical lower bounds.

\paragraph{Reporting.}
We report means and sample standard deviations across evaluated repetitions. Mean rankings are descriptive, without paired significance tests. In particular, Activity has an elevated real--real Set-Discr reference, whose cause is not isolated here. Generator scores should be interpreted relative to this reference as well as the direct diagnostics.

\subsection{Common MMD Estimator}
\label{app:metric-mmd}

Pattern-MMD$^2$ and Trans-MMD$^2$ use the same Gaussian-kernel
estimator~\citep{JMLR:v13:gretton12a}.
For $A=\{a_i\}_{i=1}^{n}$ and $B=\{b_j\}_{j=1}^{m}$, let
$k_\sigma(a,b)
=
\exp\!\left(-\lVert a-b\rVert_2^2/(2\sigma^2)\right)$.
We use the unbiased estimator
\begin{align}
\widehat{\mathrm{MMD}}_{\mathrm U,\sigma}^{\,2}(A,B)
&=
\frac{1}{n(n-1)}
\sum_{i\ne i'}k_\sigma(a_i,a_{i'})
+
\frac{1}{m(m-1)}
\sum_{j\ne j'}k_\sigma(b_j,b_{j'})
\nonumber\\
&\quad
-
\frac{2}{nm}
\sum_{i=1}^{n}\sum_{j=1}^{m}
k_\sigma(a_i,b_j),
\end{align}
requiring at least two elements in each collection.
For reporting, negative estimates are clipped as
$d_\sigma(A,B)
=
\max\{0,\widehat{\mathrm{MMD}}_{\mathrm U,\sigma}^{\,2}(A,B)\}$.
These are descriptive discrepancies, not calibrated hypothesis tests. Negative-estimate clipping and four-decimal rounding mean that a reported zero does not establish equality of distributions.

Bandwidths are determined from training calibration data using the median
positive pairwise distance,
\begin{equation}
\sigma
=
\left[
\operatorname{median}_{i<j,\,
\lVert a_i-a_j\rVert_2^2>10^{-12}}
\lVert a_i-a_j\rVert_2^2
\right]^{1/2}.
\end{equation}
At most $2{,}048$ calibration vectors are used, with $\sigma=1$ as a fallback
when no positive distance is available.

\subsection{Full-Sample Distinguishability}
\label{app:metric-discriminative}

\paragraph{Set-Discr.}
Set-Discr trains a permutation-invariant binary classifier to distinguish
complete real and generated samples from their observation times, feature
identities, and values.
Real and generated collections use approximately $80\%$ outer training and $20\%$ classifier test splits with seeds $12345$ and $12346$, respectively. Approximately $10\%$ of each outer training subset is held out for validation using seeds $12446$ and $12447$. Training uses balanced binary cross-entropy (BCE).
The score is
\begin{equation}
\mathrm{Set\mbox{-}Discr}
=
\left|
\mathrm{Acc}_{\mathrm{held}}-0.5
\right|.
\end{equation}

Each active measurement is represented by embeddings of its value, feature
identity, and normalized observation time.
A two-layer Transformer encoder with hidden dimension $128$, four attention
heads, feed-forward dimension $512$, and dropout $0.1$ processes these
measurement tokens.
Masked mean pooling is followed by a linear classification head, with no
measurement-index positional encoding.
The final reevaluation runner trains with Adam at learning rate $3\times10^{-4}$ for $100$ epochs and batch size $256$, with classifier seed $12345$ in an independent RNG context. A fresh classifier is trained for each comparison. Validation BCE is checked every five epochs and at the final epoch to select the minimum-BCE checkpoint. All $100$ epochs run, and classifier test data are used only for final scoring. A low score concerns this classifier and training procedure.

\subsection{Sampling-Pattern Fidelity}
\label{app:metric-pattern}

\paragraph{CoOb.}
CoOb measures feature co-observation within common time bins.
For sample $i$, let
\begin{equation}
\mathcal B_i
=
\left\{
\left\lfloor q_m^{(i)}/\delta\right\rfloor
:
m=1,\ldots,M_i
\right\}
\end{equation}
denote its occupied bins.
Define $b_{i,t,f}=1$ if feature $f$ is observed in bin $t$, and $0$ otherwise.
Repeated observations of the same feature within a bin count only once.

For each feature pair $(f_1,f_2)$, define
\begin{equation}
C_{f_1,f_2}(\mathcal D)
=
\frac{
\sum_i\sum_{t\in\mathcal B_i}
b_{i,t,f_1}b_{i,t,f_2}
}{
\sum_i|\mathcal B_i|
}.
\end{equation}
CoOb then compares the real and generated co-observation frequencies:
\begin{equation}
\mathrm{CoOb}
=
\sqrt{
\sum_{f_1<f_2}
\left(
C_{f_1,f_2}(\mathcal D_R)
-
C_{f_1,f_2}(\mathcal D_G)
\right)^2
}.
\end{equation}
We use $\delta=1$, corresponding to one minute for clinical benchmarks and $10$\,ms for Activity. Occupied bins are weighted equally. Without normalization over feature pairs, CoOb can exceed one and its scale depends on $F$. The bin width also determines which observations count as co-observed.

\paragraph{Pattern-MMD$^2$.}
Pattern-MMD$^2$ compares feature-wise observation frequency and timing across
complete samples.
We draw $L=32$ frequencies once as
$\omega_\ell\sim\mathcal N(0,\ell_t^{-2})$ with $\ell_t=0.15$, and define
\begin{equation}
\phi(\tau)
=
\frac{1}{\sqrt L}
\left[
\cos(\omega_1\tau),\ldots,\cos(\omega_L\tau),
\sin(\omega_1\tau),\ldots,\sin(\omega_L\tau)
\right]^\top.
\end{equation}
The frequencies remain fixed across comparisons.

Let $\overline N_{\mathrm{cal}}$ be the mean measurement count in the training
calibration collection and set $c=\max(1,\overline N_{\mathrm{cal}})$.
For each feature,
\begin{equation}
e_f(\mathcal S^{(i)})
=
\frac{1}{c}
\sum_{m=1}^{M_i}
\boldsymbol B_{m,f}^{(i)}
\phi\!\left(\tau_m^{(i)}\right).
\end{equation}
Because the sum is not normalized by the feature-specific observation count,
it retains both observation frequency and timing information.
Concatenating the feature summaries gives
$e(\mathcal S)
=
[e_1(\mathcal S),\ldots,e_F(\mathcal S)]$.
With
$\mathcal E_Q
=
\{e(\mathcal S):\mathcal S\in\mathcal D_Q\}$
for $Q\in\{R,G\}$,
\begin{equation}
\mathrm{Pattern\mbox{-}MMD}^{2}
=
d_{\sigma_P}
\left(
\mathcal E_R,\mathcal E_G
\right),
\end{equation}
where $\sigma_P$ is calibrated from training pattern embeddings.

\subsection{Value Fidelity and Dependencies}
\label{app:metric-value}

\paragraph{Value-W1.}
For $Q\in\{R,G\}$, define the observed values of feature $f$ as
\begin{equation}
\mathcal V_f^Q
=
\left\{
\mathbf x_{m,f}^{(i)}
:
\mathcal S^{(i)}\in\mathcal D_Q,\,
\boldsymbol B_{m,f}^{(i)}=1
\right\},
\end{equation}
and let $\widehat P_f^Q$ denote their empirical distribution.
Let $\mathcal F_V$ contain features observed in the training calibration
collection.
We compute
\begin{equation}
\mathrm{Value\mbox{-}W1}
=
\frac{1}{|\mathcal F_V|}
\sum_{f\in\mathcal F_V}
W_1
\left(
\widehat P_f^R,
\widehat P_f^G
\right).
\end{equation}
Thus, all observations are weighted equally within each feature and features
equally in the final average.
If a required feature is absent from either evaluation collection, the score
is reported as unavailable.

\paragraph{Trans-MMD$^2$.}
Trans-MMD$^2$ evaluates how feature values evolve between consecutive
observations.
For feature $f$ in sample $i$, let
$m_{i,1},\ldots,m_{i,n_{if}}$ denote its observed occasion indices ordered by
time.
For consecutive observations, define
\begin{equation}
\mathbf z_{i,k}^{(f)}
=
\begin{bmatrix}
\log\!\left(1+\Delta q_{i,k}^{(f)}\right)\\
\mathbf x_{m_{i,k},f}^{(i)}\\
\mathbf x_{m_{i,k+1},f}^{(i)}
\end{bmatrix},
\qquad
\Delta q_{i,k}^{(f)}
=
q_{m_{i,k+1}}^{(i)}
-
q_{m_{i,k}}^{(i)}.
\end{equation}
Transitions are formed only within the same sample and feature and then pooled
by feature.

Each coordinate is scaled by
$s_{f,d}
=
\max\{\operatorname{Std}_{\mathrm{cal}}(z_d^{(f)}),0.05\}$
using training transitions.
Let $\widetilde{\mathcal Z}_f^Q$ denote the scaled transition collection for
$Q\in\{R,G\}$, and let $\mathcal F_T$ contain features with at least two
training transitions.
Then
\begin{equation}
\mathrm{Trans\mbox{-}MMD}^{2}
=
\frac{1}{|\mathcal F_T|}
\sum_{f\in\mathcal F_T}
d_{\sigma_f}
\left(
\widetilde{\mathcal Z}_f^R,
\widetilde{\mathcal Z}_f^G
\right),
\end{equation}
where $\sigma_f$ is calibrated separately for each feature.
At most $8{,}192$ transitions per feature and collection are selected by
seeded uniform sampling without replacement.

\paragraph{Set-Corr.}
Set-Corr measures dependencies between asynchronously observed features.
For feature pair $(f_1,f_2)$, measurements are paired only within the same
sample and weighted by temporal proximity:
\begin{equation}
w_{i,m,n}^{f_1,f_2}
=
\exp\left(-(\tau_m^{(i)}-\tau_n^{(i)})^2 / {2h^2} \right),
\end{equation}
for
$\boldsymbol B_{m,f_1}^{(i)}
=
\boldsymbol B_{n,f_2}^{(i)}
=1$.

For any pairwise quantity $g$, define its weighted average over collection
$\mathcal D$ as
\begin{equation}
\left\langle g\right\rangle_{f_1,f_2,\mathcal D}
=
\frac{
\sum_i\sum_{m,n}
w_{i,m,n}^{f_1,f_2}g(i,m,n)
}{
\sum_i\sum_{m,n}
w_{i,m,n}^{f_1,f_2}
},
\end{equation}
where the sums include only valid observations of $f_1$ and $f_2$.
The weighted Pearson correlation
$\rho_{f_1,f_2}(\mathcal D)$ is computed from these weighted averages.

We use $h=0.05$ in normalized-time units.
A feature pair is valid when its total weight is at least $32$ and both
weighted variances are at least $10^{-12}$.
Let $\mathcal J$ contain feature pairs valid in both real and generated
collections.
Then
\begin{equation}
\mathrm{Set\mbox{-}Corr}
=
\frac{1}{|\mathcal J|}
\sum_{(f_1,f_2)\in\mathcal J}
\left|
\rho_{f_1,f_2}(\mathcal D_R)
-
\rho_{f_1,f_2}(\mathcal D_G)
\right|.
\end{equation}
The score is reported as unavailable if $\mathcal J$ is empty. Because eligibility is checked in both collections, $\mathcal J$ may differ across generators. Missing or degenerate pairs are excluded rather than penalized, so Set-Corr alone cannot establish preservation of all feature dependencies.

\subsection{Interpretation and Validation}
\label{app:metric-interpretation}

The six metrics capture complementary aspects of complete irregular time series
generation.
CoOb and Pattern-MMD$^2$ evaluate sampling structure without using values.
Value-W1, Trans-MMD$^2$, and Set-Corr evaluate marginal, temporal, and
cross-feature value fidelity. Set-Discr assesses overall real--generated
distinguishability.

These metrics are complementary rather than exhaustive.
Pattern-MMD$^2$ uses a finite-dimensional summary, while the value metrics
capture selected marginals and dependencies.
Because the value metrics are evaluated on generated sampling patterns, they
can reflect both sampling- and value-generation errors.
Real--real comparisons contextualize score magnitudes. The normalized corruption plots demonstrate relative responses within each metric, not comparable absolute sensitivity across metrics or separation from real--real variability. Finite pattern embeddings may emphasize count changes over some timing changes. These tests do not exhaustively assess sampling--value dependence, downstream utility, or memorization.

\subsection{Controlled-Corruption Protocol}
\label{app:controlled-corruptions}

\paragraph{Data and randomness.}
We corrupt a fixed real training control, disjoint from the $1{,}024$ calibration records, against an unchanged real test reference.
Each collection has $4{,}739/2{,}397/3{,}844/4{,}955/478$ samples for eICU/P12/MIMIC-III/MIMIC-IV/Activity.
Normalization and calibration remain fixed.
For severity $s\in\{0,0.25,0.5,0.75,1\}$, each positive level uses corruption seeds $12345$--$12347$, with the RNG reset for each operator--seed--severity combination.
Severity zero shares one uncorrupted evaluation, not three independent repetitions.
Direct-metric transition subsampling uses seed $12345$.
Masks determine observation status, including observed zeros. Deleted entries have zero values and false masks.

\paragraph{Operators.}
Let $L$ be the number of benchmark time indices and $\epsilon\sim\mathcal N(0,1)$.
The designated metric is given in parentheses.
\begin{itemize}[leftmargin=*,nosep]
\item \textbf{Count dropout (Pattern-MMD$^2$).} Independently delete observed entries with probability $0.6s$. Shared random draws make deletions nested across severities. Retained values are unchanged, but counts, panels, and transition gaps can change.
\item \textbf{Co-observation shift (CoOb).} Select each sample--feature trajectory with probability $s$ and cyclically shift its mask and values by an independently uniform integer offset in $\{1,\ldots,L-1\}$. Counts and value multisets are preserved, but cross-feature alignment and boundary transitions can change.
\item \textbf{Time jitter (Trans-MMD$^2$).} Process occupied times $q$ in ascending order, moving each entire panel and value row to $\operatorname{clip}(\operatorname{round}(q+0.25Ls\epsilon),0,L-1)$. Resolve collisions at the nearest unused index, choosing the smaller index on ties. No observations are merged. Counts, panel multisets, and value multisets are preserved, but temporal order can change.
\item \textbf{Value noise (Value-W1).} Add independent $0.5s\epsilon$ to observed values in fixed normalized units, without clipping or renormalization. Masks and times remain unchanged.
\item \textbf{Cross-feature shuffle (Set-Corr.).} For each feature, group samples by equal observation count. Randomly select $\lfloor sn\rfloor$ of a group's $n$ samples and cyclically exchange their complete observed value sequences, retaining recipient masks and times and donor order. Fewer than two selected samples cause no change. Features are processed independently. Feature-wise value and adjacent-value-pair multisets are preserved, while cross-feature coupling and gap--value associations can change.
\end{itemize}
These operators differ from the upstream perturbations in \Cref{tab:error-propagation}, whose severity is not interchangeable with $s$ here.

\paragraph{Aggregation.}
Let $q_{dkc}(s)$ average defined scores over corruption seeds for dataset $d$, metric $k$, and corruption $c$, using the shared baseline at zero.
For designated corruption $c_k$, the sensitivity curve averages
\begin{equation}
A_{dk}(s)=\frac{q_{dkc_k}(s)-q_{dkc_k}(0)}{q_{dkc_k}(1)-q_{dkc_k}(0)}
\end{equation}
equally across datasets. Shading is the between-dataset sample standard deviation, not seed variability or a confidence interval.
Undefined endpoints or absolute denominators below $10^{-15}$ exclude a curve. Endpoint one is imposed by normalization and is not evidence of sensitivity by itself.
For the heatmap, define
\begin{equation}
\Delta_{dkc}=\max\{q_{dkc}(1)-q_{dkc}(0),0\},
\qquad H_{kc}=\operatorname{mean}_{d}\frac{\Delta_{dkc}}{\max_{c'}\Delta_{dkc'}}.
\end{equation}
Normalize over the five corruptions within each dataset--metric row before averaging datasets. Rows with maximum at most $10^{-15}$ contribute zeros. Undefined entries remain excluded.
Designated heatmap cells are not forced to one. These positive-increase profiles describe relative responses within metrics.

\paragraph{Valid repetitions.}
All designated sensitivity combinations have three valid seeds at positive severities.
For P12, Set-Corr is undefined for every positive count-dropout and co-observation-shift level. Its time-jitter scores use two seeds at $s=0.25,0.5,1$, and Trans-MMD$^2$ uses two seeds for count dropout at $s=1$.
Consequently, the Set-Corr/count-dropout and Set-Corr/co-observation-shift heatmap cells average four datasets. Other cells average five, with these endpoint exceptions.
Set-Discr classifier training is not part of this validation.

\section{Details of Experiments}
\label{app:exp_details}

\subsection{Details of Datasets}
\label{app:dataset_details}

We use four clinical datasets---eICU, PhysioNet 2012,
MIMIC-III, and MIMIC-IV---and one nonclinical sensor
dataset, Human Activity, collected during human activities.
Each dataset has training, validation, and test partitions.
Our experiments evaluate generated observation patterns
and feature values against the test partition.
\Cref{tab:dataset-statistics} summarizes dataset dimensions,
observation horizons, temporal resolutions, event budgets,
and split sizes.

\paragraph{Temporal representation.}
The clinical datasets use one-minute timestamp resolution.
We use 24-hour observation windows for eICU and
48-hour windows for P12, MIMIC-III, and MIMIC-IV.
Activity uses 2.28-second windows with timestamps
quantized to $10$\,ms intervals.
These representations retain variable observation counts,
irregular temporal gaps, and feature-specific missingness
within each window.

\paragraph{Observed values and missingness.}
We retain an explicit observation mask and do not
interpolate or forward-fill missing entries, so we rely
only on observed data.
The mask distinguishes observed zeros from missing entries,
and we compute feature-wise normalization statistics
only from observed values.
When constructing measurement-level representations, we represent each observed
entry by its timestamp, feature identity, and value.

\paragraph{Record filtering.}
We exclude records with no observed entries.
For the minute-resolution clinical datasets, we additionally
exclude records with observations at more than half of
the time indices in the window.
This rule removes unusually dense monitoring streams.
The split sizes in \Cref{tab:dataset-statistics} are
reported after filtering. Consequently, the results characterize these retained windows and do not establish performance on the excluded dense records.

\begin{table*}[t]
\centering
\caption{Dataset statistics for the five irregular time series
benchmarks. Clinical split sizes are reported as negative/positive
outcome counts, while Activity split sizes indicate
the total number of windows across its seven activity classes.
$N_{\max}$ is the measurement budget, not the number of observation occasions.
}
\label{tab:dataset-statistics}
\vspace{-.05in}
\setlength{\tabcolsep}{5pt}
\resizebox{.8\textwidth}{!}{%
\begin{tabular}{l|ccccc}
    \toprule
    & eICU & P12 & MIMIC-III & MIMIC-IV & Activity \\
    \midrule
    \# Features & 11 & 35 & 28 & 28 & 12 \\
    Time Window & 24h & 48h & 48h & 48h & 2.28s \\
    Min Resolution & 1m & 1m & 1m & 1m & 10ms \\
    Max Measurements ($N_{\max}$) & 672 & 1,024 & 1,024 & 1,024 & 600 \\
    Train (Neg / Pos) & 13,817 / 1,346 & 6,579 / 1,093 & 10,655 / 1,611 & 13,806 / 2,046 & 2,860 \\
    Val (Neg / Pos) & 3,454 / 336 & 1,645 / 273 & 2,662 / 404 & 3,452 / 512 & 471 \\
    Test (Neg / Pos) & 4,318 / 421 & 2,056 / 341 & 3,339 / 505 & 4,315 / 640 & 478 \\
    Total & 23,692 & 11,987 & 19,176 & 24,771 & 3,809 \\
    \bottomrule
\end{tabular}%
}
\end{table*}

\subsection{Details of Generation}
\label{app:generation_details}
We compare ChronoFlow with Poisson + LatentODE, LogNormMix + mTAN, LogNormMix + TFM, and FlexTPP on Activity, eICU, PhysioNet 2012, MIMIC-III, and MIMIC-IV.
For each model--dataset pair, we generate as many synthetic samples as in the corresponding test set and evaluate them against the full test set.
We evaluate all models in the unconditional full-generation setting, where both the sampling pattern and associated feature values are generated without a ground-truth observation pattern.

ChronoFlow first generates global observation statistics,
then jointly generates observation times and feature panels,
and finally generates feature values conditioned on the resulting sampling pattern.
The default sampling configuration uses $100$ ODE integration steps per flow module.
We vary the sampling-step budget while keeping the trained model fixed to examine its effect on generation fidelity.

\paragraph{Seeds and repetitions.}
\label{app:seed-protocol}
The main comparison uses independently trained checkpoints with seeds $12345$, $12346$, and $12347$ on fixed data partitions.
Training seeds govern initialization, minibatch order, dropout, and training noise. Generated samples add sampling variation, so the reported sample standard deviation (divisor $3-1$) reflects both training and generation, not independent classifier initializations or a confidence interval.
Set-Discr fixes its classifier seed to $12345$ (\Cref{app:metric-discriminative}). This main-table repetition scheme does not specify the repetition schemes of the separate ablation and intervention studies.

\paragraph{Conditioning and decoding.}
\label{app:conditioning}
PatternFlow adds projected noisy times and panels, slot embeddings, flow-time embeddings, and a broadcast embedding of the encoded statistics $\boldsymbol c_I=[2M/M_{\max}-1,2\boldsymbol r-1]$:
\begin{equation}
h_m=E_\tau((\boldsymbol\tau_t)_m)+E_B((\boldsymbol B_t)_{m,:})+e_m+E_t(t)+E_I(\boldsymbol c_I).
\end{equation}
The Transformer output feeds separate time and panel heads.
The measurement budget $N_{\max}$ in \Cref{tab:dataset-statistics} is distinct from the occasion bound $M_{\max}$.
Frequency projection only places each $r_f$ on its $1/M$ lattice. It does not enforce $\sum_m B_{m,f}=Mr_f$ after panel thresholding and empty-panel repair.

\paragraph{Scope of the factorization study.}
\label{app:factorization-scope}
\Cref{tab:fidelity-ablation-eicu} compares the listed eICU configurations, whose capacities and conditioning variables differ. It does not separately isolate the benefit of $\boldsymbol r$, or compare time-then-panel and panel-then-time alternatives. Thus, it supports the complete evaluated hierarchy without establishing that each factorization choice is independently optimal.

\paragraph{Interpreting oracle interventions.}
\label{app:intervention-scope}
Replacing upstream variables changes downstream conditioning while leaving the trained model fixed. Improved CoOb with ground-truth statistics implicates upstream error on eICU, but does not establish IntensityFlow as the sole bottleneck. Under ground-truth pattern conditioning, CoOb and Pattern-MMD$^2$ reflect the supplied patterns, so their reported zeros do not evaluate learned pattern generation. Stable Value-W1 concerns marginals only, and the large Set-Discr variation prevents claiming uniform improvement.

\begin{table*}[t]
\centering
\caption{
Stage-wise oracle intervention and perturbation analysis of \method{} on eICU.
At inference, generated upstream variables are replaced by their ground-truth counterparts to assess sensitivity to upstream conditioning.
We further perturb ground-truth upstream variables with relative noise of severity $s$ to evaluate downstream robustness; ground-truth variables are taken from the evaluated test samples, so these rows can fall below the real--real reference.
$(M,\boldsymbol r)$ denotes global observation statistics, $(\boldsymbol\tau,\boldsymbol B)$ the sampling pattern, and $\mathbf x$ the feature values.
}
\label{tab:error-propagation}
\setlength{\tabcolsep}{4pt}
\resizebox{\textwidth}{!}{%
\begin{tabular}{l|ccc|cccccc}
\toprule
Model
& $(M,\boldsymbol r)$
& $(\boldsymbol\tau,\boldsymbol B)$
& $\mathbf x$
& Set-Discr. $\downarrow$
& CoOb $\downarrow$
& Pattern-MMD$^2$ $\downarrow$
& Trans-MMD$^2$ $\downarrow$
& Value-W1 $\downarrow$
& Set-Corr. $\downarrow$ \\
\midrule

Oracle
& GT & GT & GT
& 0.0019{\scriptsize$\pm$0.0012}
& 0.0306{\scriptsize$\pm$0.0155}
& 0.0000{\scriptsize$\pm$0.0000}
& 0.0004{\scriptsize$\pm$0.0001}
& 0.0312{\scriptsize$\pm$0.0007}
& 0.0364{\scriptsize$\pm$0.0015} \\

\midrule
\multicolumn{10}{c}{\textit{Stage-wise Oracle Intervention}} \\
\midrule

\rowcolor{gray!20}
ChronoFlow
& sampled & sampled & sampled
& 0.0995{\scriptsize$\pm$0.0098}
& 0.0380{\scriptsize$\pm$0.0047}
& 0.0002{\scriptsize$\pm$0.0001}
& 0.0045{\scriptsize$\pm$0.0002}
& 0.0504{\scriptsize$\pm$0.0047}
& 0.0385{\scriptsize$\pm$0.0003} \\

+ GT $(M,\boldsymbol r)$
& GT & sampled & sampled
& 0.0153{\scriptsize$\pm$0.0140}
& 0.0128{\scriptsize$\pm$0.0011}
& 0.0000{\scriptsize$\pm$0.0000}
& 0.0040{\scriptsize$\pm$0.0002}
& 0.0481{\scriptsize$\pm$0.0037}
& 0.0365{\scriptsize$\pm$0.0022} \\

+ GT $(M,\boldsymbol r)$, $(\boldsymbol\tau,\boldsymbol B)$
& GT & GT & sampled
& 0.0928{\scriptsize$\pm$0.1521}
& 0.0000{\scriptsize$\pm$0.0000}
& 0.0000{\scriptsize$\pm$0.0000}
& 0.0003{\scriptsize$\pm$0.0002}
& 0.0508{\scriptsize$\pm$0.0100}
& 0.0358{\scriptsize$\pm$0.0019} \\

\midrule
\multicolumn{10}{c}{\textit{Robustness to Upstream Perturbations}} \\
\midrule

GT $(M,\boldsymbol r)$, $s=0.1$
& GT + $s$ & sampled & sampled
& 0.0113{\scriptsize$\pm$0.0066}
& 0.0616{\scriptsize$\pm$0.0021}
& 0.0000{\scriptsize$\pm$0.0000}
& 0.0044{\scriptsize$\pm$0.0003}
& 0.0470{\scriptsize$\pm$0.0006}
& 0.0376{\scriptsize$\pm$0.0038} \\

GT $(M,\boldsymbol r)$, $s=0.2$
& GT + $s$ & sampled & sampled
& 0.0413{\scriptsize$\pm$0.0524}
& 0.1620{\scriptsize$\pm$0.0042}
& 0.0007{\scriptsize$\pm$0.0001}
& 0.0057{\scriptsize$\pm$0.0006}
& 0.0494{\scriptsize$\pm$0.0023}
& 0.0341{\scriptsize$\pm$0.0032} \\

GT $(M,\boldsymbol r)$, $s=0.4$
& GT + $s$ & sampled & sampled
& 0.2704{\scriptsize$\pm$0.0516}
& 0.3898{\scriptsize$\pm$0.0047}
& 0.0088{\scriptsize$\pm$0.0009}
& 0.0115{\scriptsize$\pm$0.0004}
& 0.0600{\scriptsize$\pm$0.0016}
& 0.0374{\scriptsize$\pm$0.0058} \\

GT $(M,\boldsymbol r)$, $(\boldsymbol\tau,\boldsymbol B)$, $s=0.1$
& GT + $s$ & GT + $s$ & sampled
& 0.0109{\scriptsize$\pm$0.0068}
& 0.2604{\scriptsize$\pm$0.0005}
& 0.0000{\scriptsize$\pm$0.0000}
& 0.0013{\scriptsize$\pm$0.0001}
& 0.0460{\scriptsize$\pm$0.0015}
& 0.0378{\scriptsize$\pm$0.0049} \\

GT $(M,\boldsymbol r)$, $(\boldsymbol\tau,\boldsymbol B)$, $s=0.2$
& GT + $s$ & GT + $s$ & sampled
& 0.0680{\scriptsize$\pm$0.0737}
& 0.4594{\scriptsize$\pm$0.0005}
& 0.0002{\scriptsize$\pm$0.0000}
& 0.0037{\scriptsize$\pm$0.0002}
& 0.0473{\scriptsize$\pm$0.0026}
& 0.0364{\scriptsize$\pm$0.0050} \\

GT $(M,\boldsymbol r)$, $(\boldsymbol\tau,\boldsymbol B)$, $s=0.4$
& GT + $s$ & GT + $s$ & sampled
& 0.3034{\scriptsize$\pm$0.0650}
& 0.8015{\scriptsize$\pm$0.0012}
& 0.0054{\scriptsize$\pm$0.0003}
& 0.0069{\scriptsize$\pm$0.0001}
& 0.0503{\scriptsize$\pm$0.0013}
& 0.0406{\scriptsize$\pm$0.0058} \\

\bottomrule
\end{tabular}%
}
\vspace{-.1in}
\end{table*}

\section{Details of Baselines}
\label{app:baseline_details}

We compare ChronoFlow with four baselines that generate both
sampling patterns and feature values.
All baselines generate samples without access to test observations,
timestamps, or observation masks.
For methods that model values at given observation times,
we provide a learned sampling-pattern generator to obtain
a complete generative baseline.

\paragraph{Poisson + LatentODE.}
Latent ODE~\citep{rubanova2019latent} models continuous-time
latent dynamics and decodes them into observed values.
We use its Poisson-process extension, in which the latent
trajectory also determines feature-specific observation intensities.
At generation time, an initial latent state is sampled from
the prior and evolved through the ODE.
The resulting intensities determine the sampling pattern,
and the decoder provides the corresponding feature values.
Our implementation projects the observation process onto
the dataset's timestamp resolution by sampling whether each
time bin contains an observation.
This baseline therefore models observation patterns and values
through a shared latent trajectory.

\paragraph{LogNormMix + mTAN.}
We combine a sampling-pattern generator based on
LogNormMix~\citep{shchur2020intensity} with the variational
formulation of Multi-Time Attention Networks
(mTAN)~\citep{shukla2021multitime}.
Our LogNormMix adapter autoregressively predicts feature
identities and inter-event times.
It includes a point mass at zero to represent simultaneous
observations and an end-of-sequence token to model variable
event counts.
The mTAN component learns a latent representation using
observed values, timestamps, and masks during training.
At generation time, latent variables are sampled from its prior
and decoded at the timestamps generated by LogNormMix.
The generated feature mask selects the values retained
in the output.
This composition is our adaptation for full generation,
rather than a full-generation model proposed in the original
mTAN work.

\paragraph{LogNormMix + TFM.}
This baseline combines the same LogNormMix sampling-pattern
model family with Trajectory Flow Matching
(TFM)~\citep{zhang2024trajectory}.
TFM learns the drift and diffusion of a neural stochastic
differential equation through simulation-free training.
Our adapter trains the value model on real sampling patterns
using masked losses for observed coordinates.
Past observations are carried forward when constructing
the history provided to the value model.
For unconditional generation, we sample the initial value
state from a learned diagonal Gaussian and evolve the
trajectory at the generated observation times using
Euler--Maruyama integration.
Only values selected by the generated feature mask are
retained.
The learned initial-state distribution and LogNormMix
component allow sampling without a ground-truth initial
observation or sampling pattern.

\paragraph{FlexTPP.}
FlexTPP~\citep{draxler2025transformers} is an autoregressive Transformer
for event sequences with categorical and continuous attributes.
We adapt its event representation to encode feature identity,
a zero-gap indicator, a positive inter-event time when applicable,
and an observed value.
The zero-gap indicator allows multiple features to be
observed at the same timestamp.
We retain the upstream Transformer and spline-based density
components and introduce an end-of-sequence token for
variable-length generation.
During sampling, event attributes are generated sequentially
from the previously generated history.
The event representation and timestamp projection are
adaptations to our evaluation setting.

\subsection{Hyperparameter Optimization}

We select baseline learning rates through an initial screen on eICU, comparing multipliers of $\{0.5, 1, 2\}$ relative to each model's starting learning rate.
Screening runs used seed $12345$ and a nominal budget of $256{,}000$ training-sample presentations. Candidates were evaluated on the validation split using $256$ generated samples and selected by their mean rank across the preliminary screening metrics.
Ties were resolved by proximity to the starting learning rate and then by the smaller learning rate. The selected learning rates were used in the final baseline training runs.
The screening protocol preceded the final evaluation protocol and used a smaller generated sample set.
For the main results, we instead generate as many samples as there are records in the corresponding test set.

\Cref{tab-eicu-training-config} summarizes the recorded
eICU optimization settings.
For the baseline runs listed here, the nominal training
budget corresponds to 150,000 updates at batch size 256.
The number of optimizer updates varies with the effective
batch size.
For LogNormMix + mTAN, this budget applies separately
to the pattern and value components, which are combined
without an additional joint optimization stage.

\begin{table}[t]
\centering
\caption{
Optimization settings for the evaluated eICU checkpoints.
The listed learning rate is the peak rate for FlexTPP and the configured base rate for ChronoFlow.
ChronoFlow uses a warmup target of $8 \times 10^{-4}$ and a minimum learning rate of $10^{-5}$.
}
\label{tab-eicu-training-config}
\setlength{\tabcolsep}{5pt}
\begin{tabular}{lccc}
\toprule
Model
& Optimizer
& LR setting
& Schedule \\
\midrule
Poisson + LatentODE
& Adamax
& $10^{-2}$
& Exponential decay \\

LogNormMix + mTAN (Pattern)
& Adam
& $2 \times 10^{-4}$
& Constant \\

LogNormMix + mTAN (Value)
& Adam
& $2 \times 10^{-3}$
& Constant \\

LogNormMix + TFM
& Adam
& $2 \times 10^{-4}$
& Constant \\

FlexTPP
& AdamW
& $5 \times 10^{-4}$
& One-cycle \\

ChronoFlow
& Adam
& $4 \times 10^{-5}$
& Warmup + plateau \\
\bottomrule
\end{tabular}

\end{table}

\section{Visualization of Generated Samples}
\label{app:visualization}

\begin{figure*}[p]
\centering

\includegraphics[width=\textwidth]{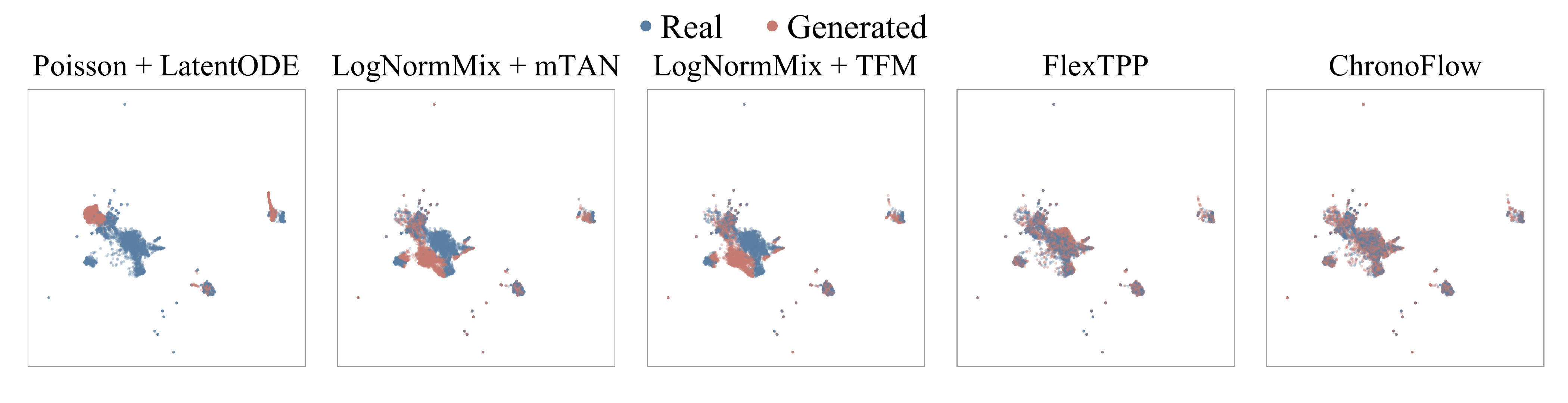}

\includegraphics[width=\textwidth]{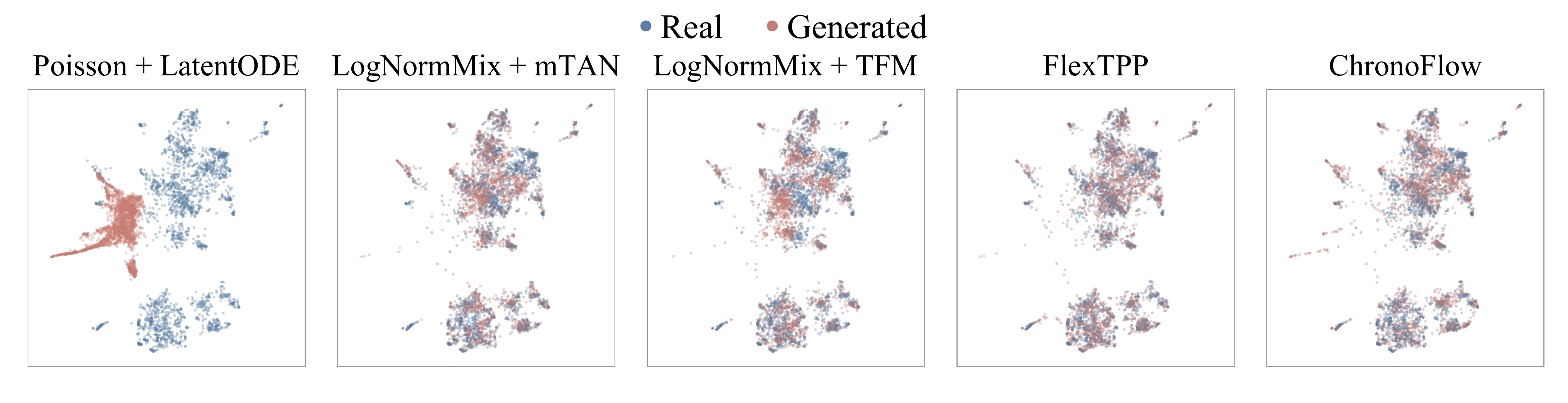}

\includegraphics[width=\textwidth]{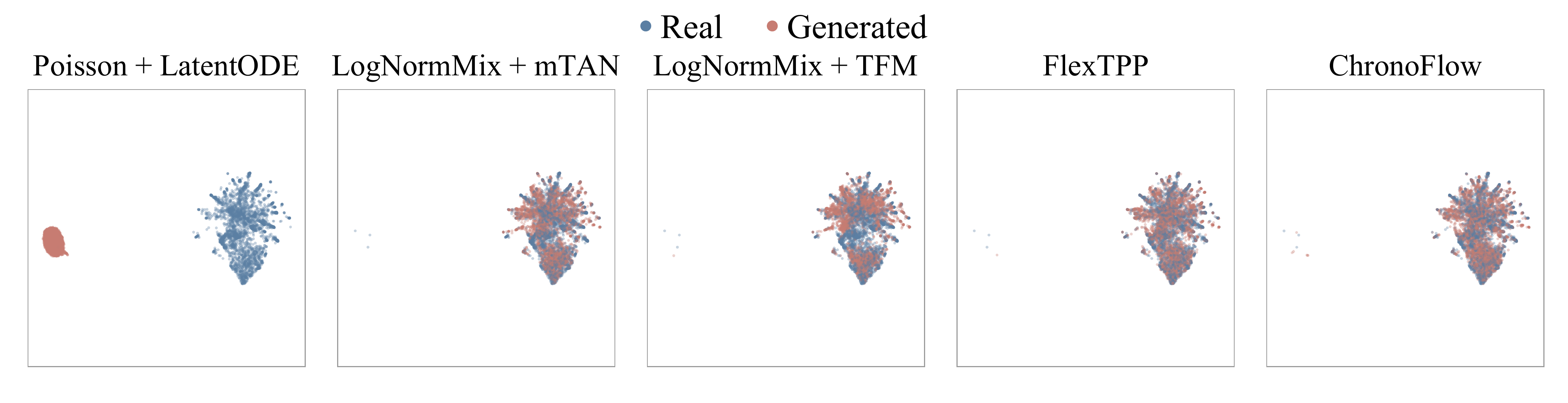}

\includegraphics[width=\textwidth]{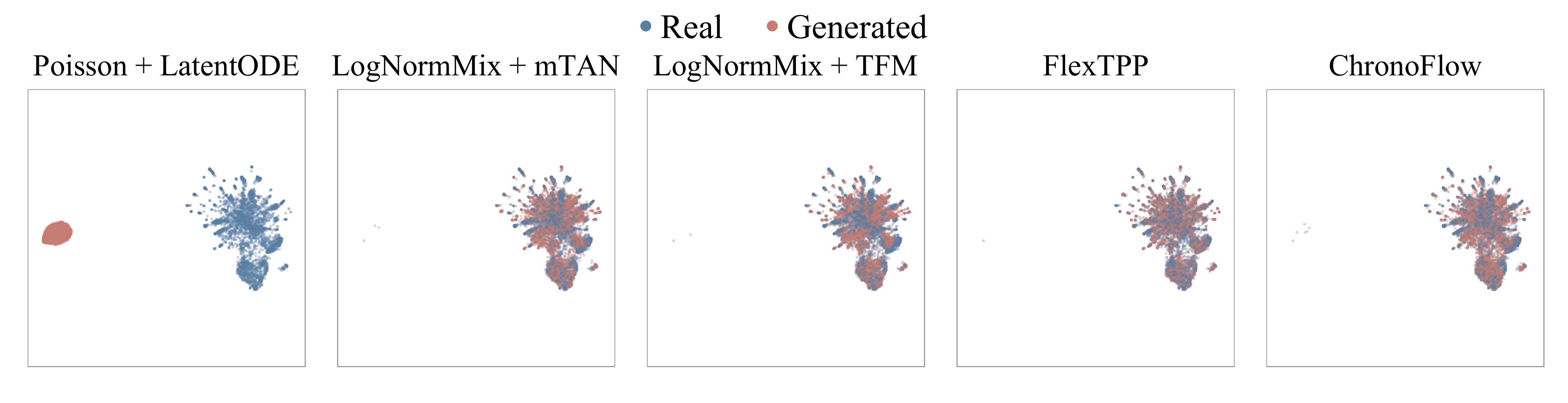}

\includegraphics[width=\textwidth]{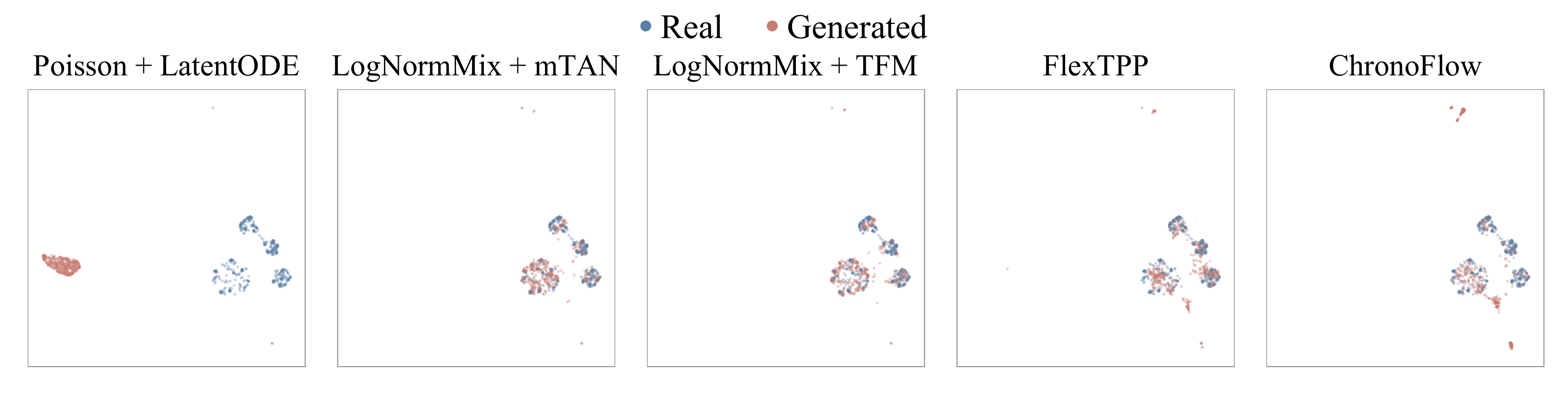}

\caption{
UMAP visualizations of real and generated samples across the five benchmarks:
eICU, P12, MIMIC-III, MIMIC-IV, and Activity, shown from top to bottom.
For each dataset, the panels compare the real-data representation with samples generated by
Poisson + LatentODE, LogNormMix + mTAN, LogNormMix + TFM, FlexTPP, and \method{}.
Overlap is interpreted qualitatively, as two-dimensional projection can hide discrepancies.
}
\label{fig:umap_all}
\end{figure*}

\begin{figure*}[t]
\centering
\includegraphics[width=\textwidth]{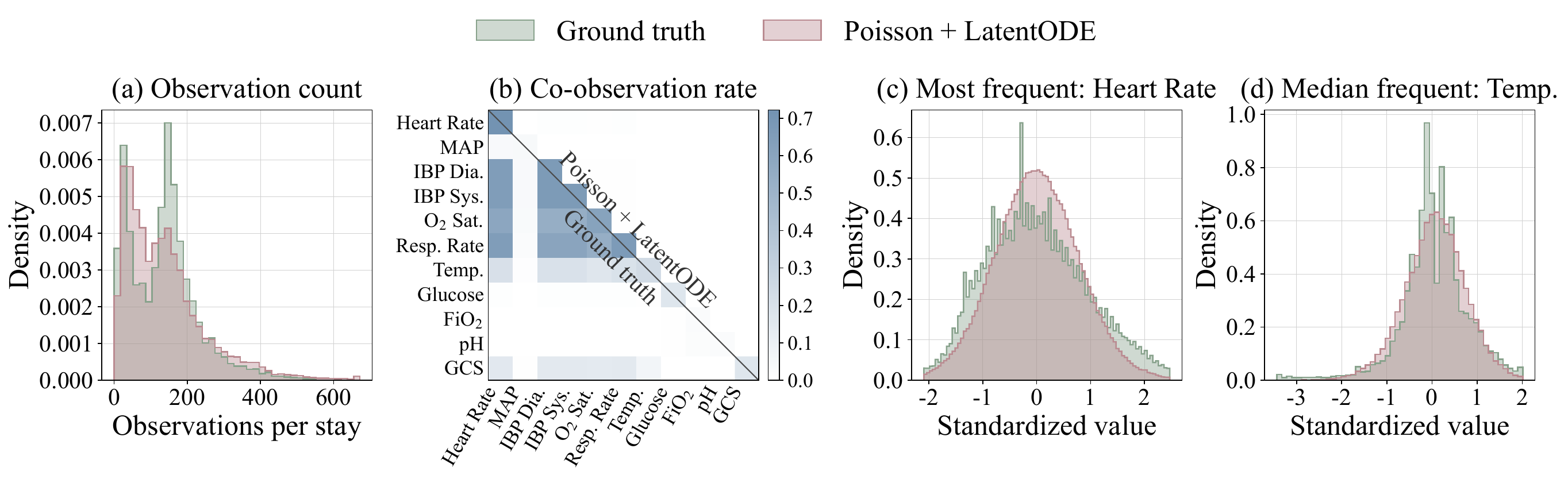} \vspace{-.12in}
\includegraphics[width=\textwidth]{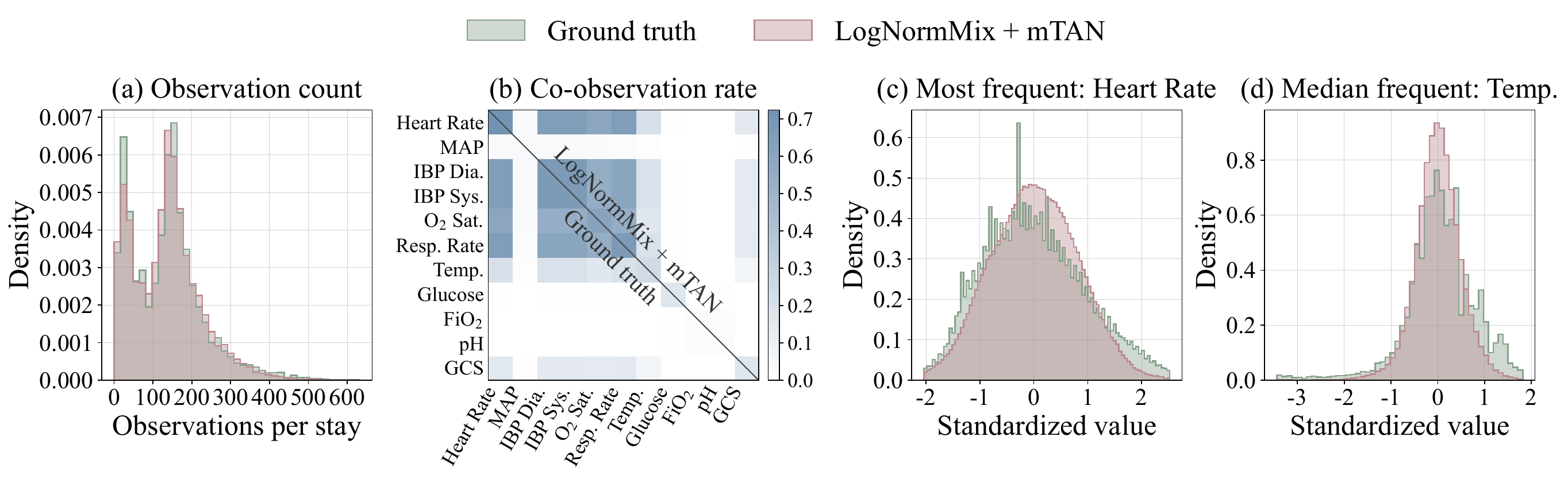} \vspace{-.12in}
\includegraphics[width=\textwidth]{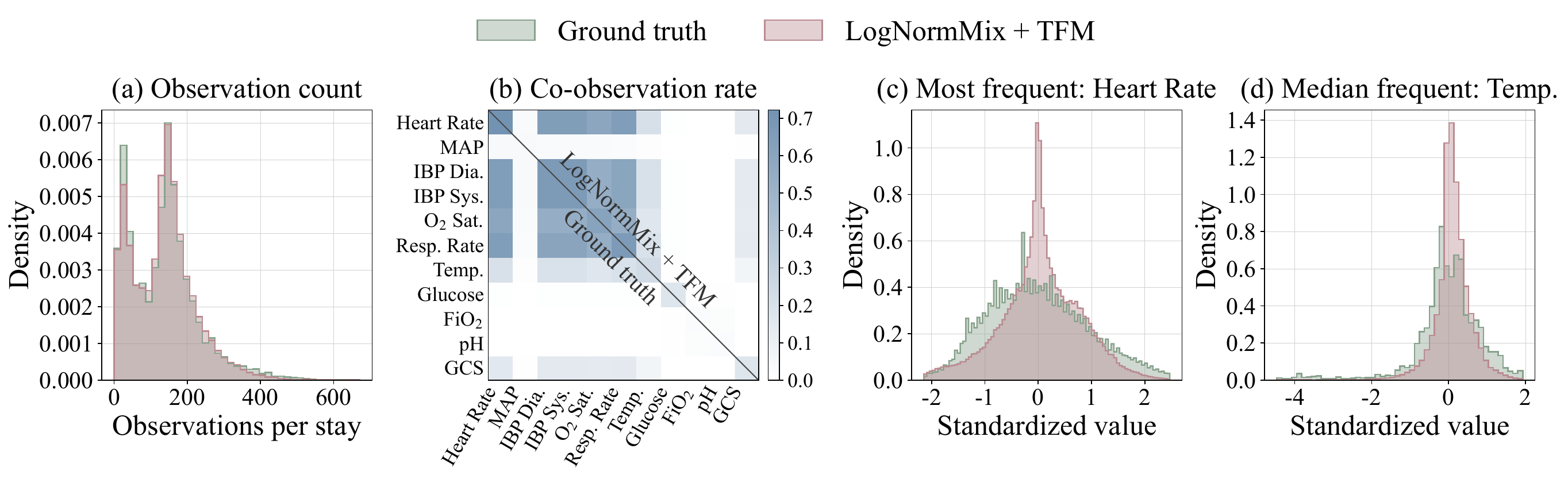} \vspace{-.12in}
\includegraphics[width=\textwidth]{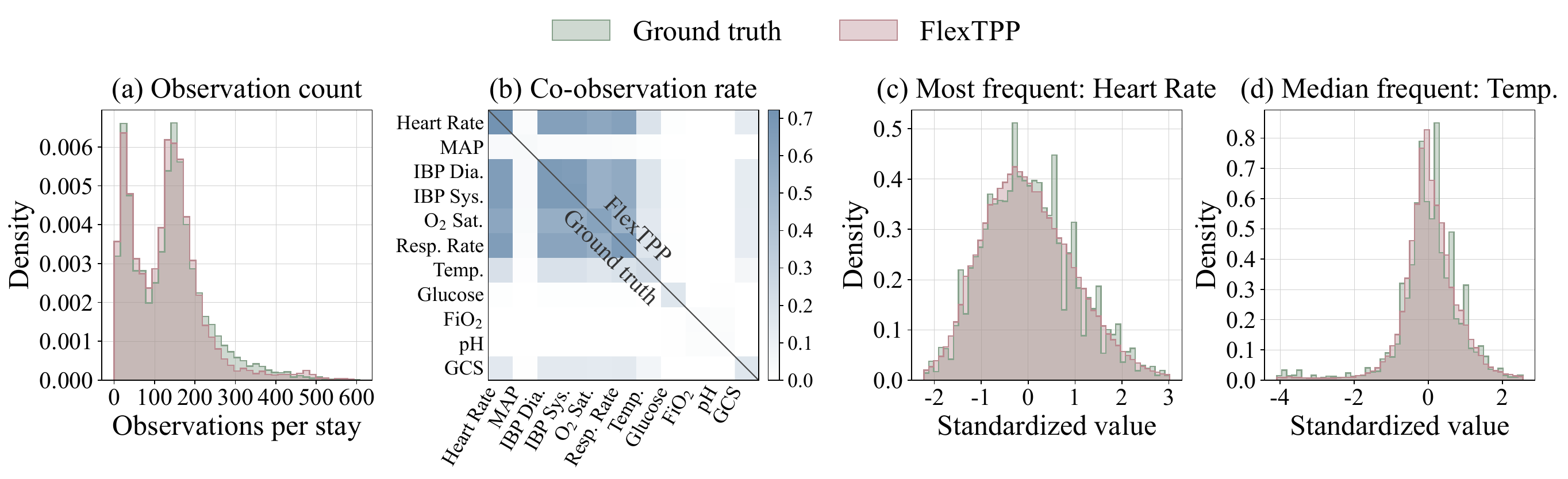} \vspace{-.12in}
\includegraphics[width=\textwidth]{raw_figures/eicu_chronoflow_fidelity.pdf} \vspace{-.15in}
\caption{
Qualitative comparison of baseline methods and ChronoFlow on the eICU benchmark.
For this dataset, rows correspond to Poisson + LatentODE, LogNormMix + mTAN, LogNormMix + TFM, FlexTPP, and ChronoFlow, respectively.
For each model, (a) shows the distribution of the total number of observations per stay.
Panel (b) shows feature co-observation rates for all 11 features.
The lower triangle shows ground truth, the upper triangle shows generated samples, and the diagonal shows ground-truth marginals.
Panels (c, d) show value distributions of the most frequent feature and the median-frequency feature.
}
\label{fig:eicu_fidelity_baselines}
\vspace{-.15in}
\end{figure*}

\begin{figure*}[t]
\centering
\includegraphics[width=\textwidth]{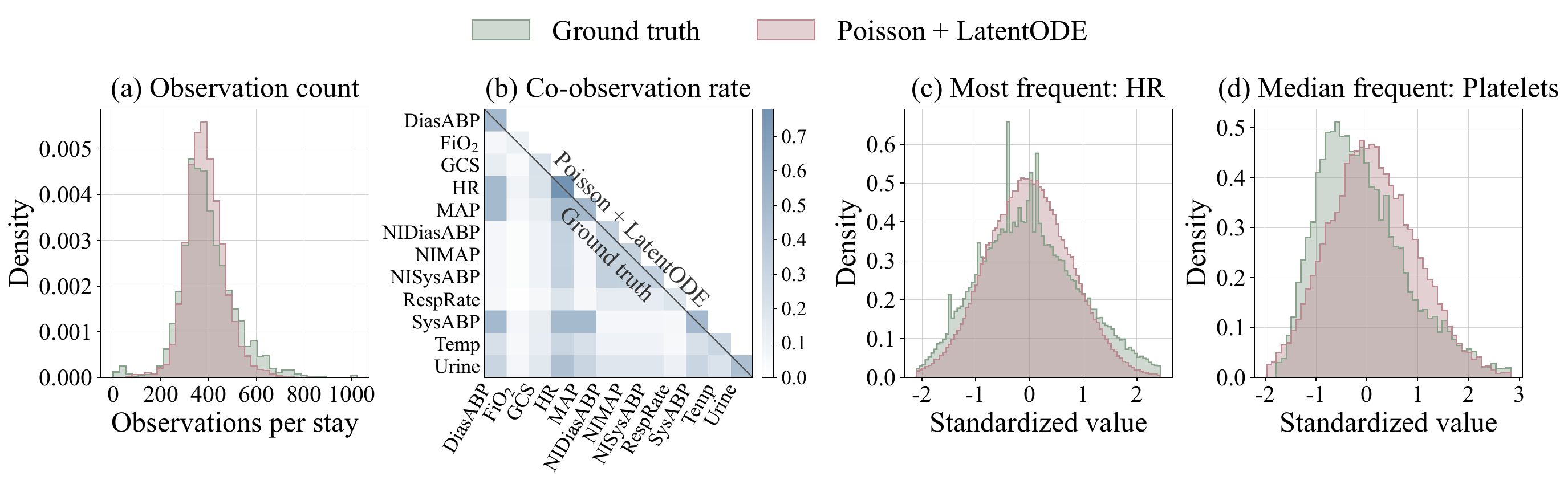} \vspace{-.12in}
\includegraphics[width=\textwidth]{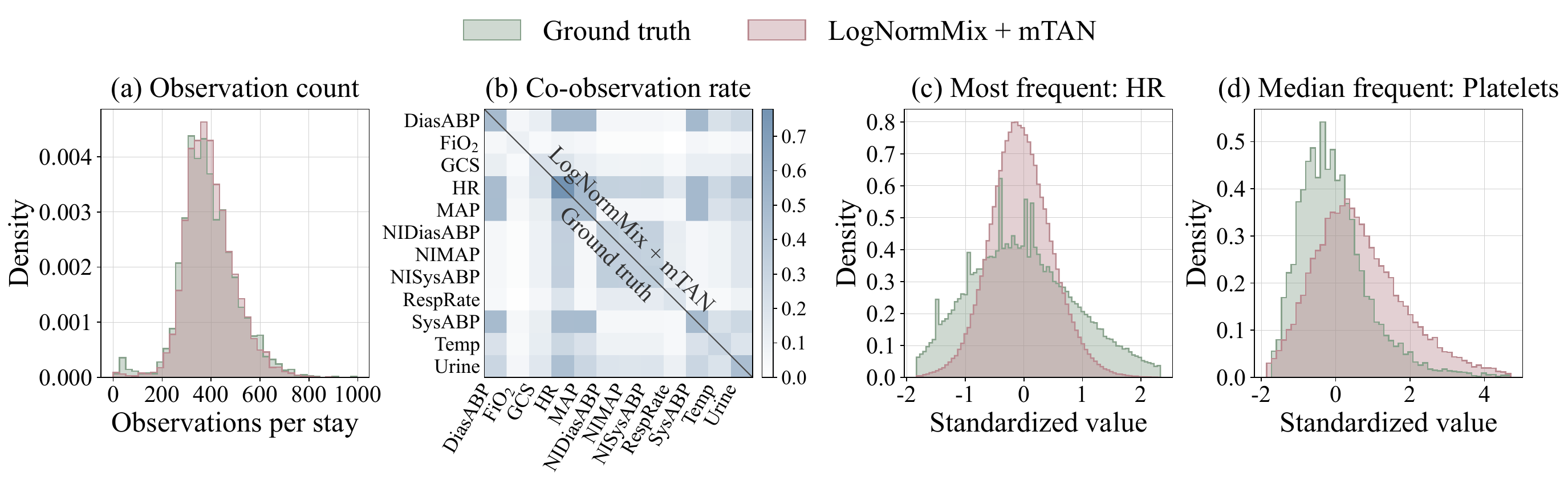} \vspace{-.12in}
\includegraphics[width=\textwidth]{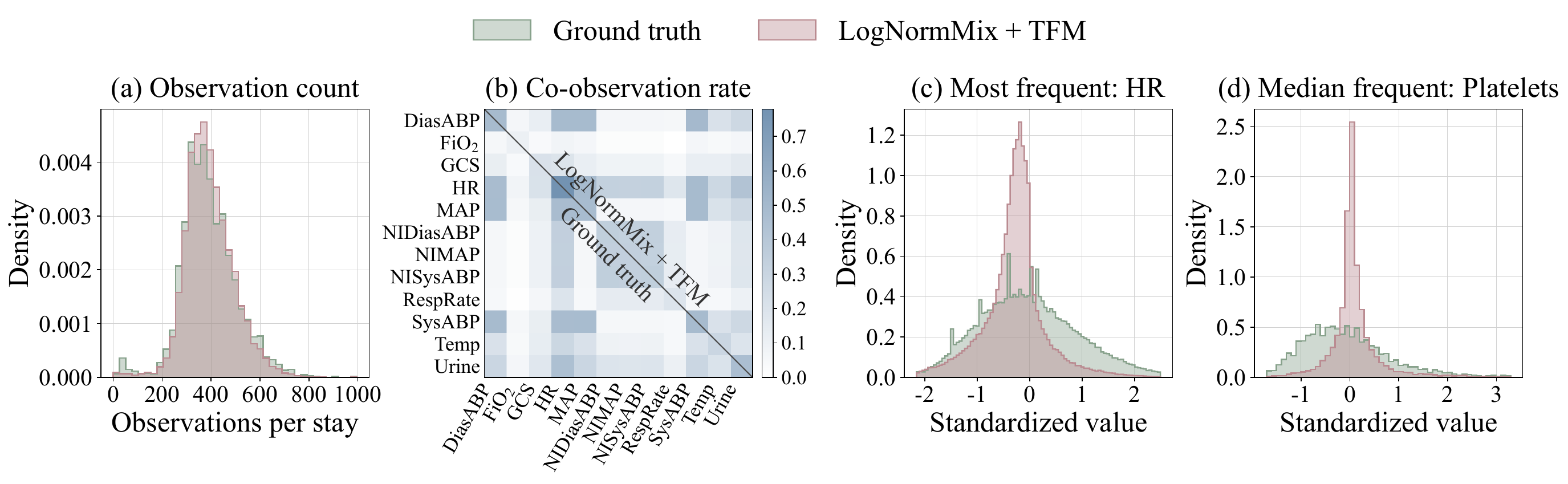} \vspace{-.12in}
\includegraphics[width=\textwidth]{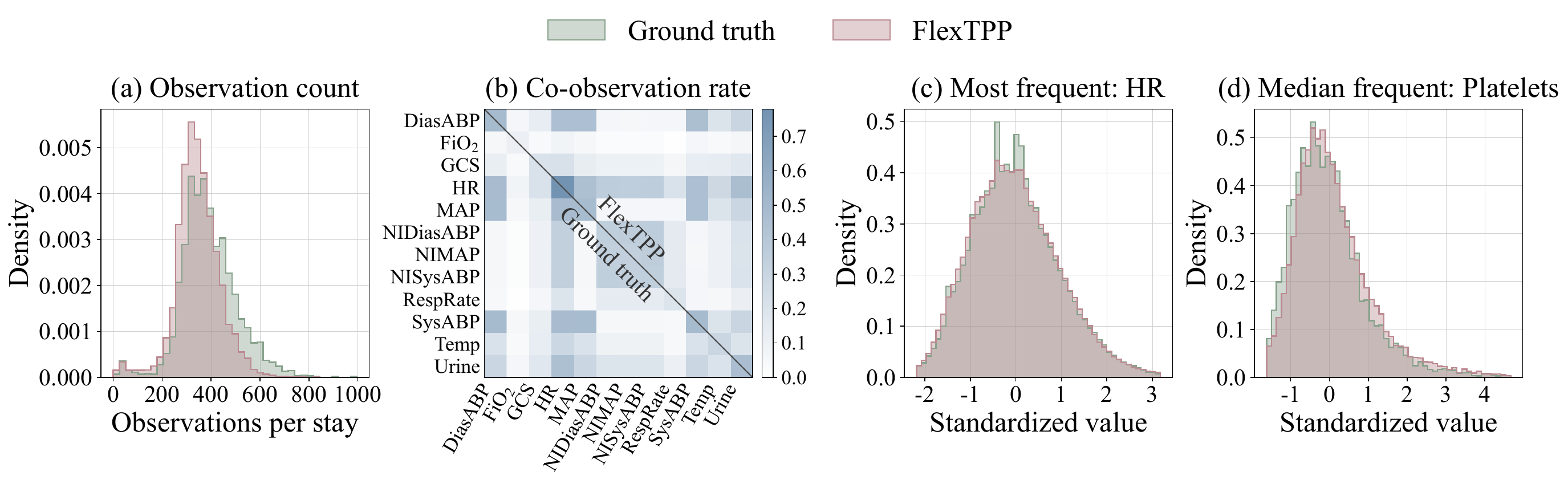} \vspace{-.12in}
\includegraphics[width=\textwidth]{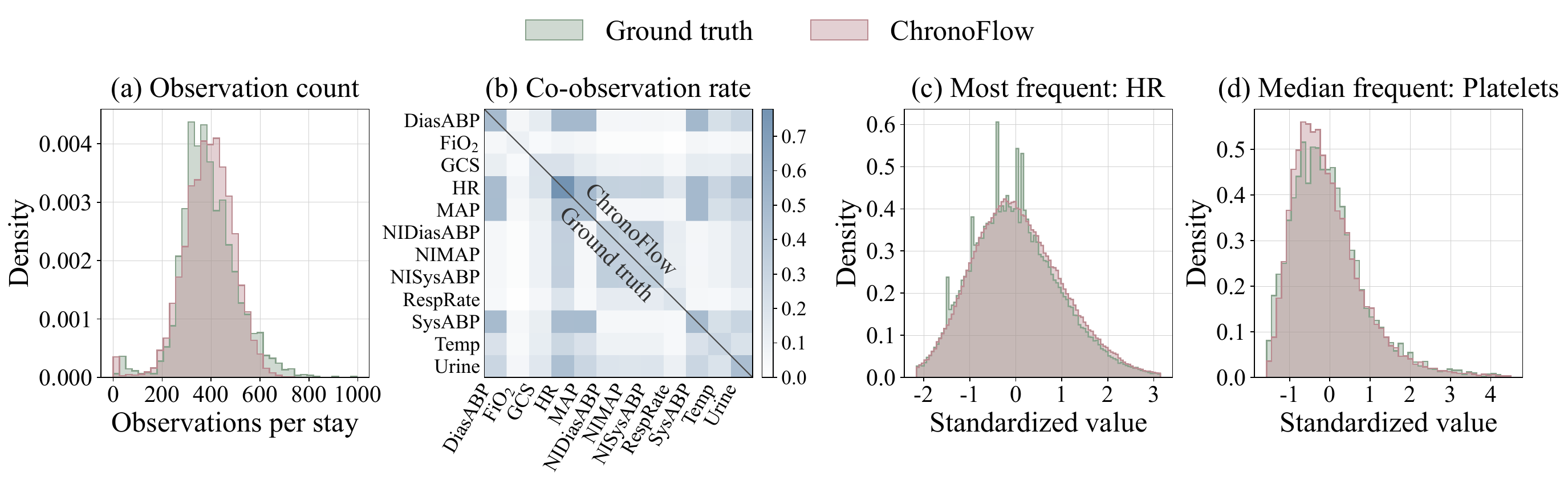} \vspace{-.15in}
\caption{
Qualitative comparison of baseline methods and ChronoFlow on the P12 benchmark.
For this dataset, rows correspond to Poisson + LatentODE, LogNormMix + mTAN, LogNormMix + TFM, FlexTPP, and ChronoFlow, respectively.
For each model, (a) shows the distribution of the total number of observations per stay.
Panel (b) shows feature co-observation rates for the 12 most frequent features.
The lower triangle shows ground truth, the upper triangle shows generated samples, and the diagonal shows ground-truth marginals.
Panels (c, d) show value distributions of the most frequent feature and the median-frequency feature.
}
\label{fig:p12_fidelity_baselines}
\vspace{-.15in}
\end{figure*}

\begin{figure*}[t]
\centering
\includegraphics[width=\textwidth]{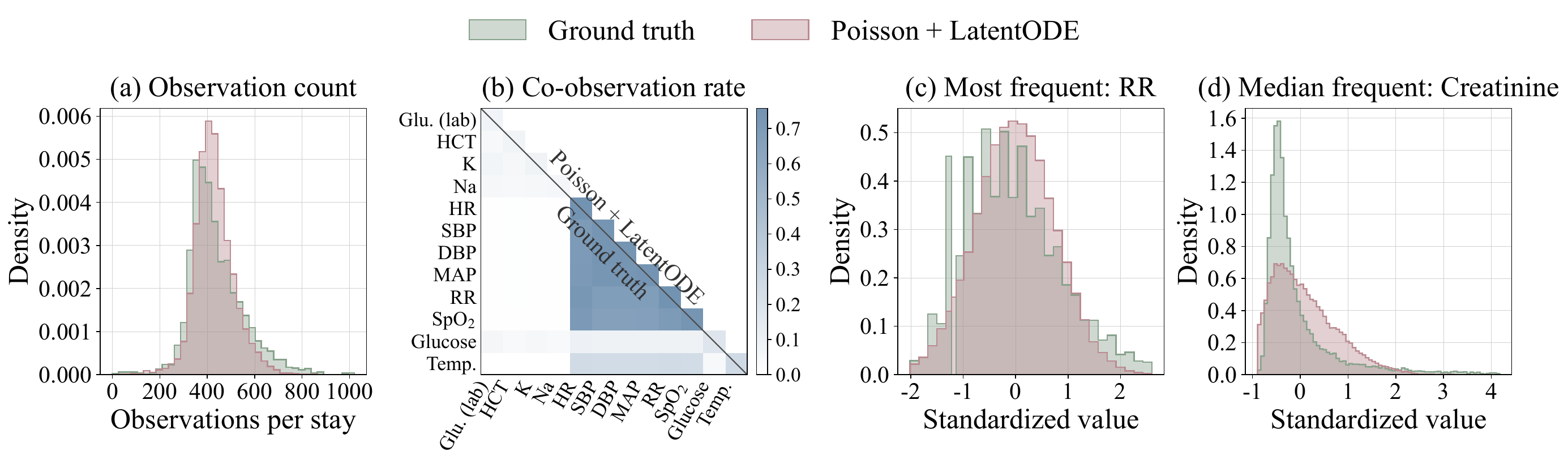} \vspace{-.12in}
\includegraphics[width=\textwidth]{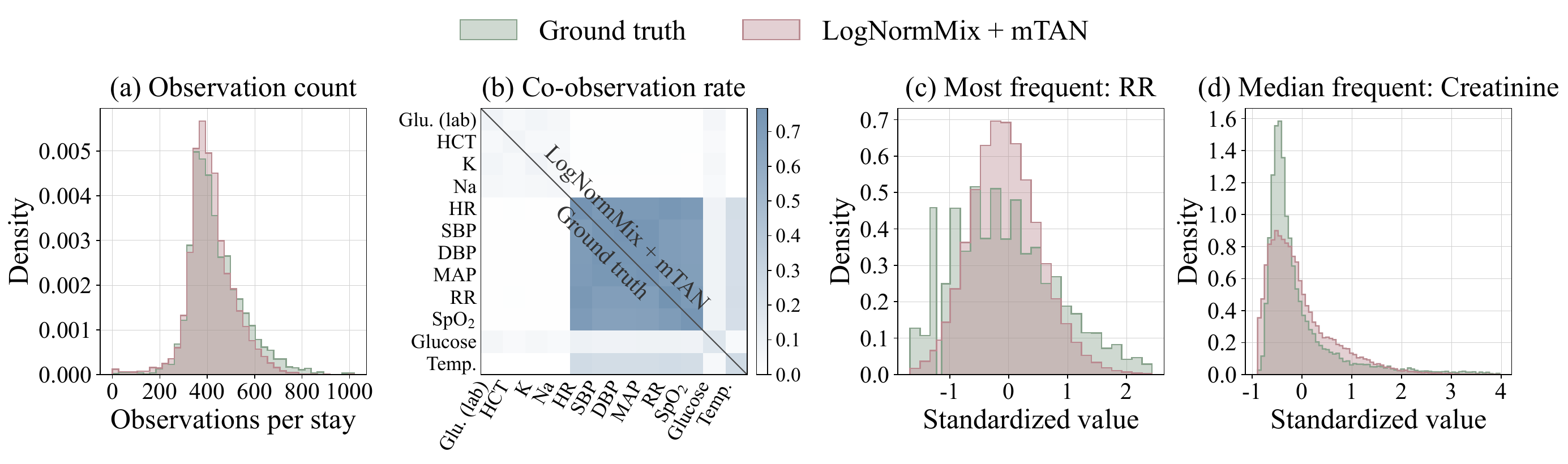} \vspace{-.12in}
\includegraphics[width=\textwidth]{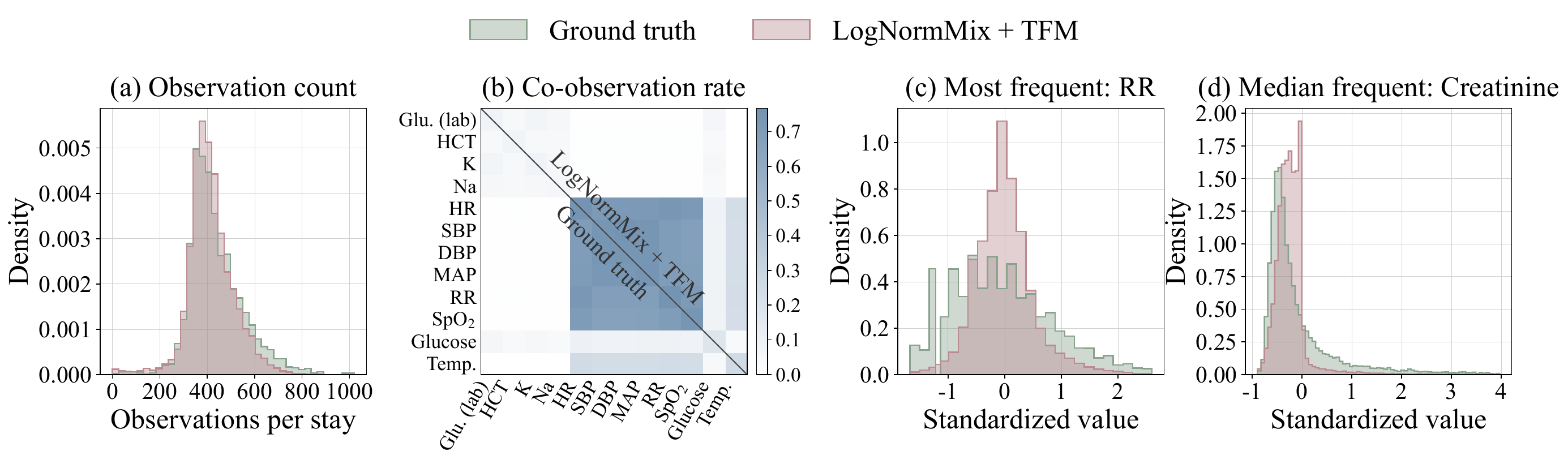} \vspace{-.12in}
\includegraphics[width=\textwidth]{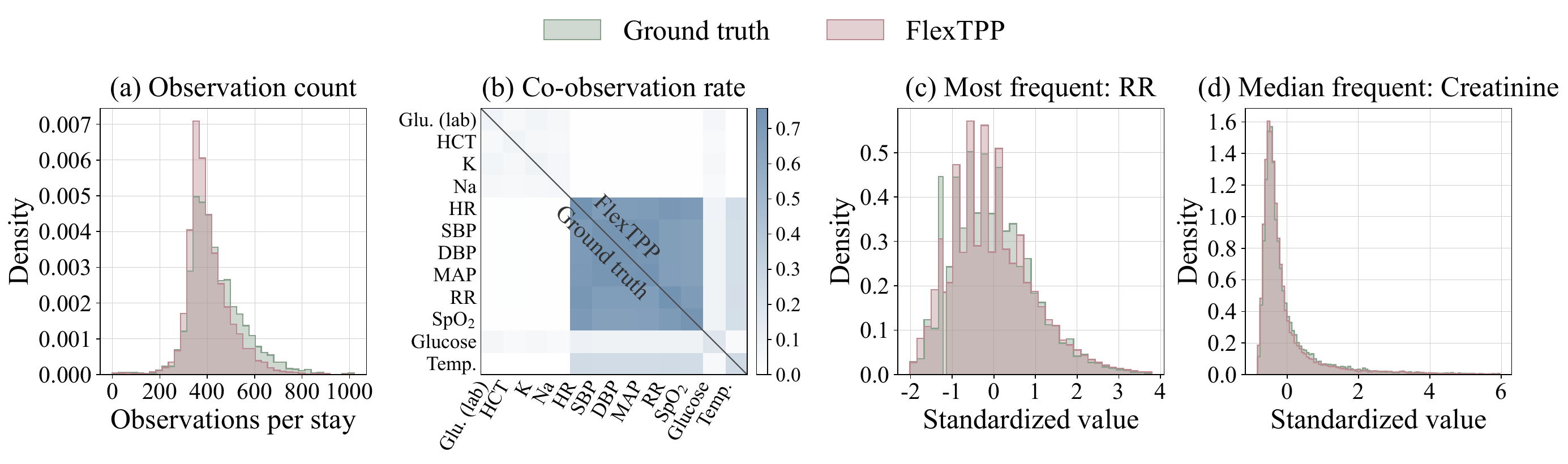} \vspace{-.12in}
\includegraphics[width=\textwidth]{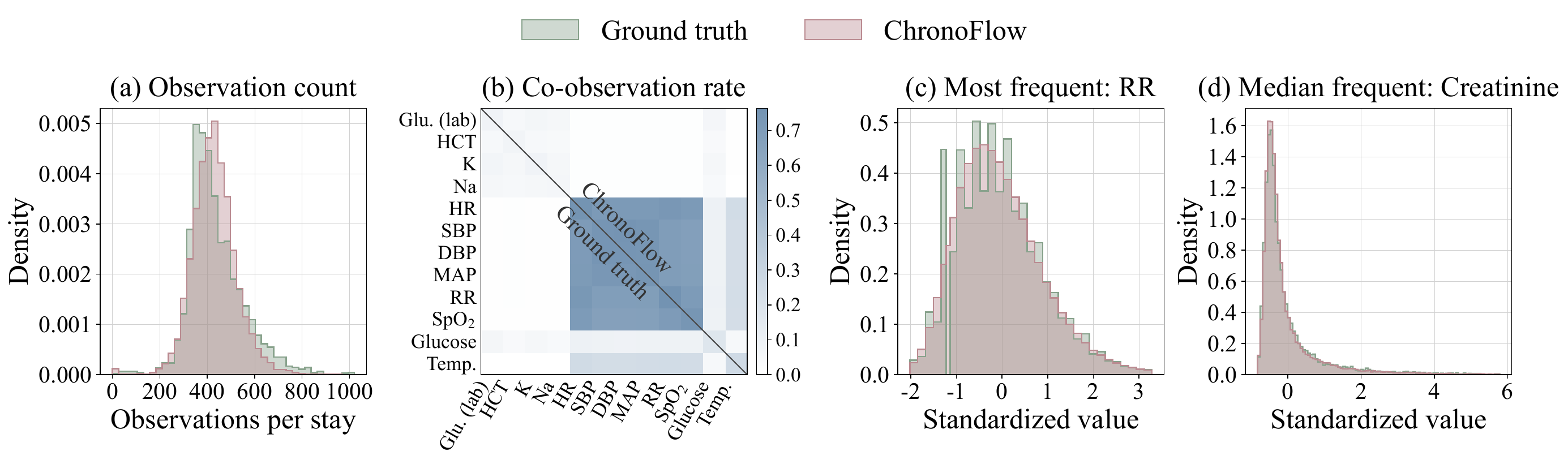} \vspace{-.15in}
\caption{
Qualitative comparison of baseline methods and ChronoFlow on the MIMIC-III benchmark.
For this dataset, rows correspond to Poisson + LatentODE, LogNormMix + mTAN, LogNormMix + TFM, FlexTPP, and ChronoFlow, respectively.
For each model, (a) shows the distribution of the total number of observations per stay.
Panel (b) shows feature co-observation rates for the 12 most frequent features.
The lower triangle shows ground truth, the upper triangle shows generated samples, and the diagonal shows ground-truth marginals.
Panels (c, d) show value distributions of the most frequent feature and the median-frequency feature.
}
\label{fig:mimic3_fidelity_baselines}
\vspace{-.15in}
\end{figure*}

\begin{figure*}[t]
\centering
\includegraphics[width=\textwidth]{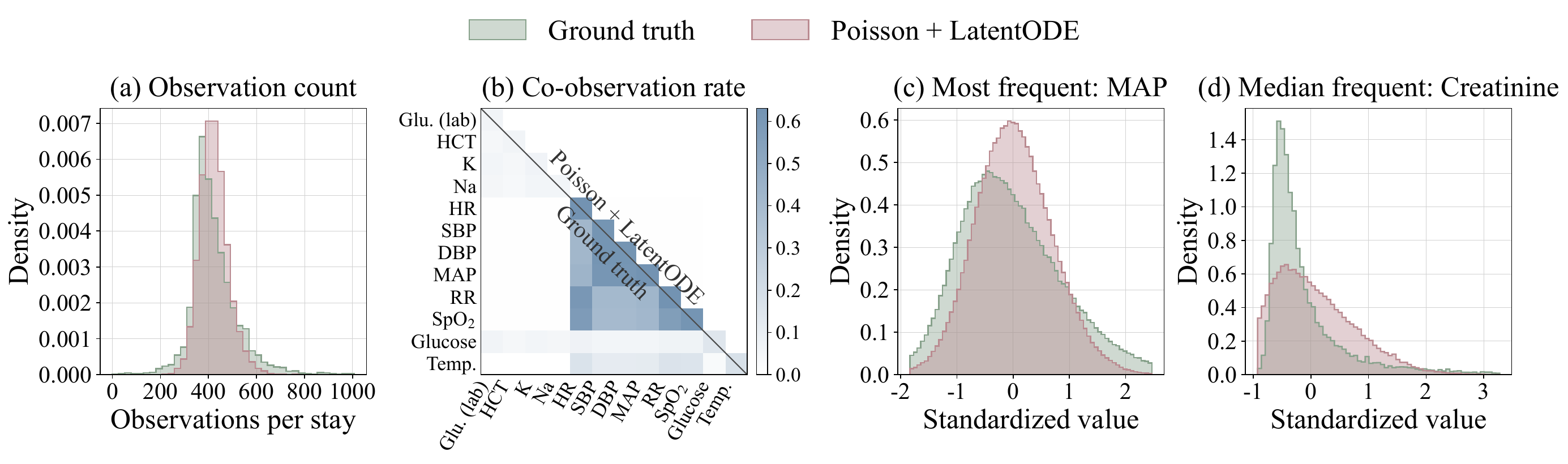} \vspace{-.12in}
\includegraphics[width=\textwidth]{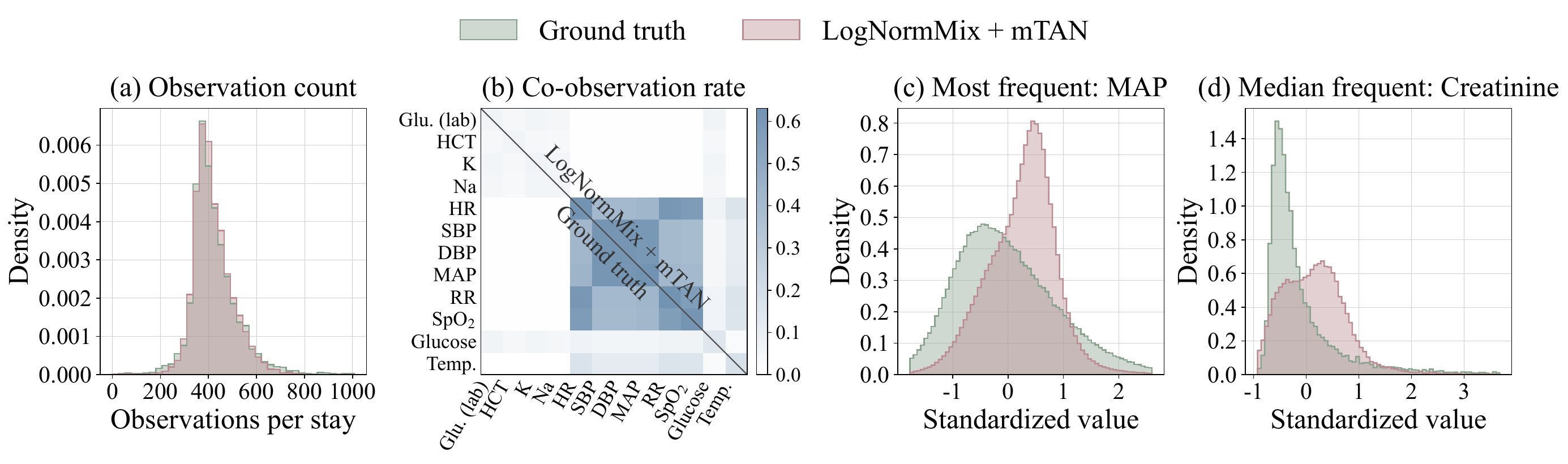} \vspace{-.12in}
\includegraphics[width=\textwidth]{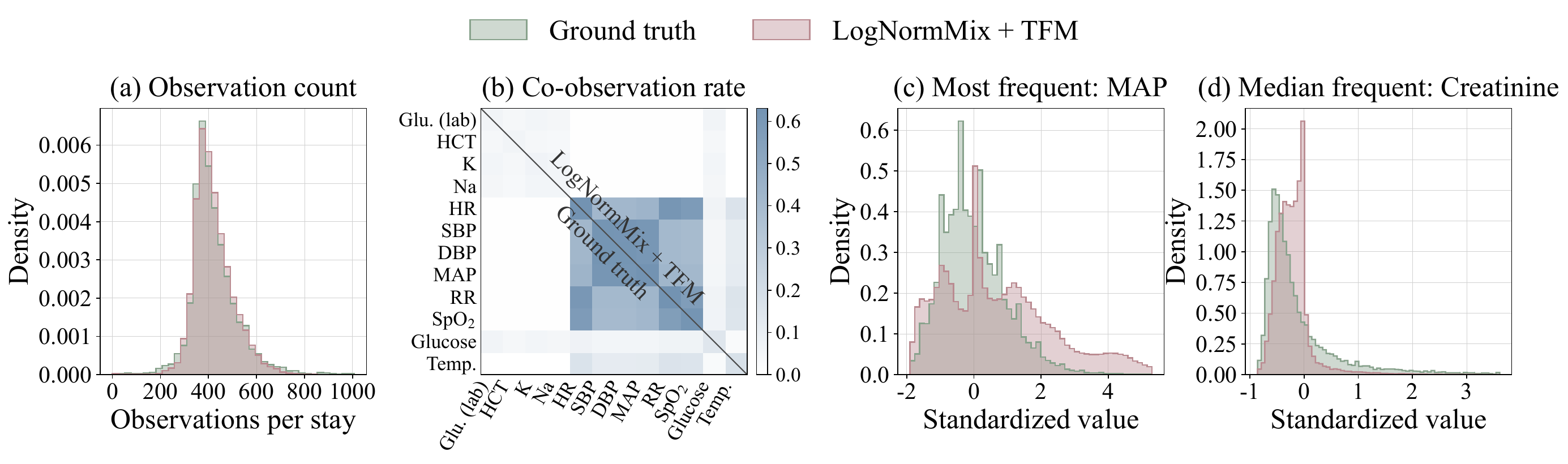} \vspace{-.12in}
\includegraphics[width=\textwidth]{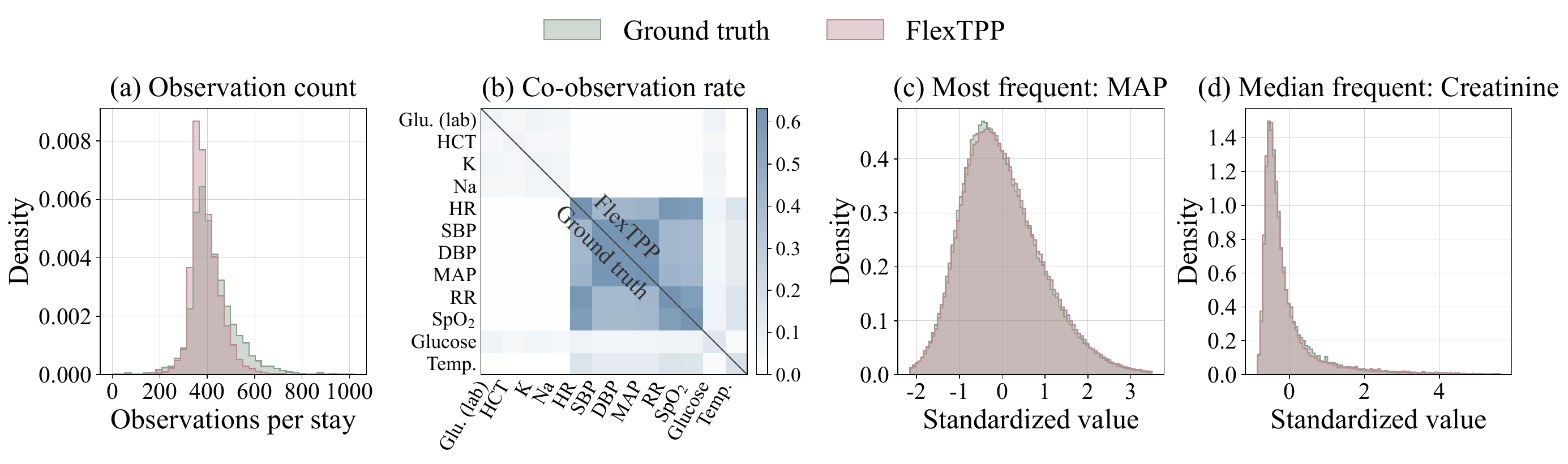} \vspace{-.12in}
\includegraphics[width=\textwidth]{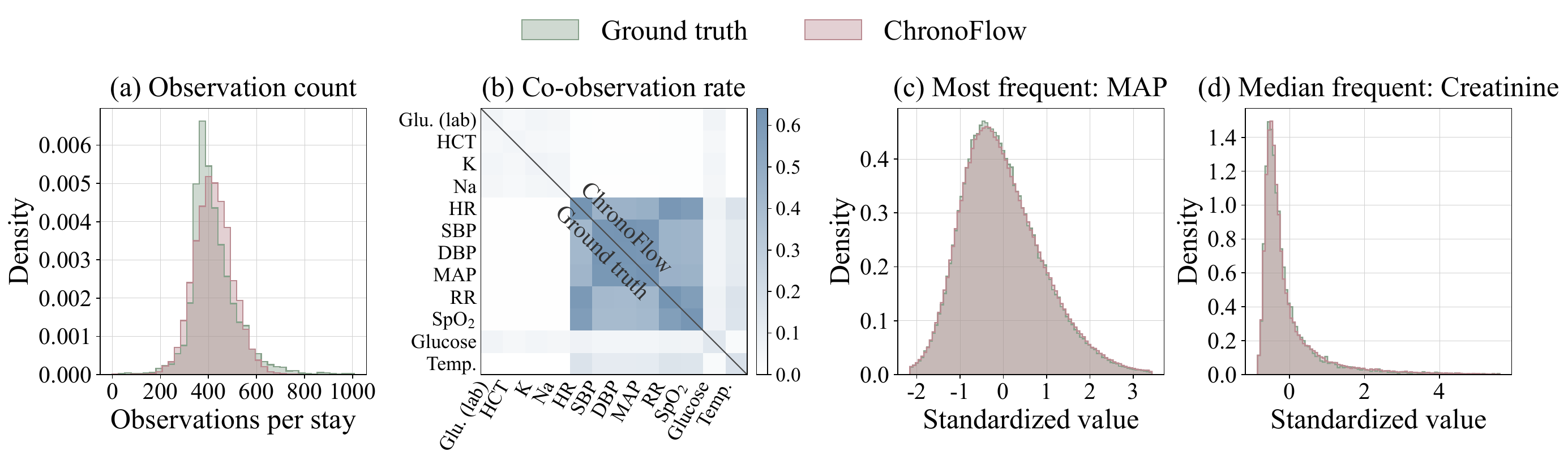} \vspace{-.15in}
\caption{
Qualitative comparison of baseline methods and ChronoFlow on the MIMIC-IV benchmark.
For this dataset, rows correspond to Poisson + LatentODE, LogNormMix + mTAN, LogNormMix + TFM, FlexTPP, and ChronoFlow, respectively.
For each model, (a) shows the distribution of the total number of observations per stay.
Panel (b) shows feature co-observation rates for the 12 most frequent features.
The lower triangle shows ground truth, the upper triangle shows generated samples, and the diagonal shows ground-truth marginals.
Panels (c, d) show value distributions of the most frequent feature and the median-frequency feature.
}
\label{fig:mimic4_fidelity_baselines}
\vspace{-.15in}
\end{figure*}

\begin{figure*}[t]
\centering
\includegraphics[width=\textwidth]{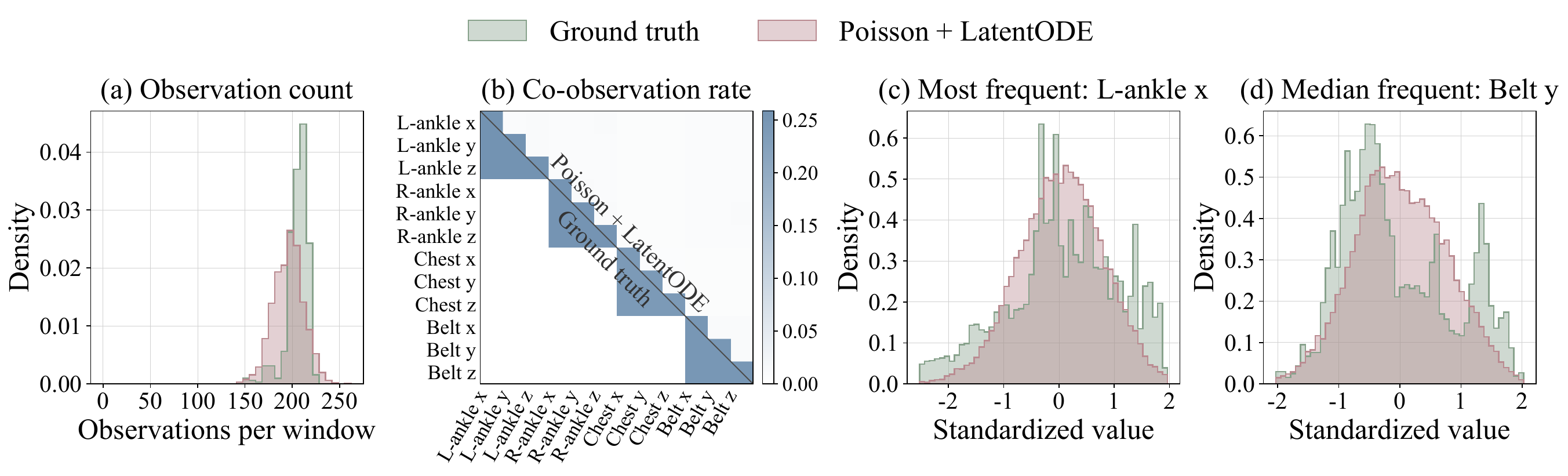} \vspace{-.12in}
\includegraphics[width=\textwidth]{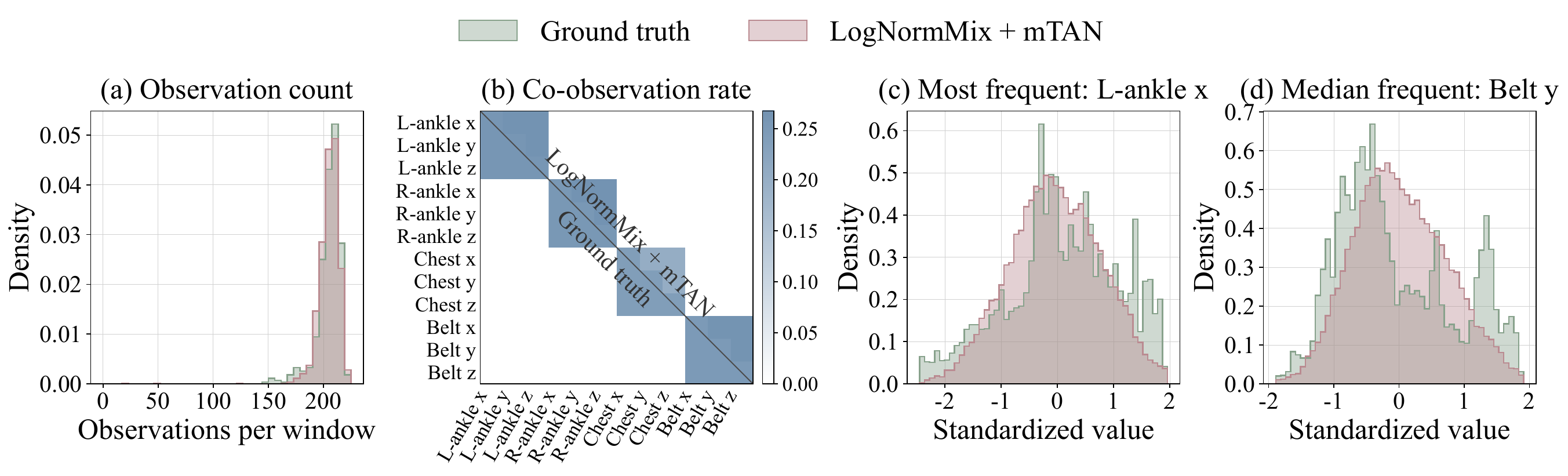} \vspace{-.12in}
\includegraphics[width=\textwidth]{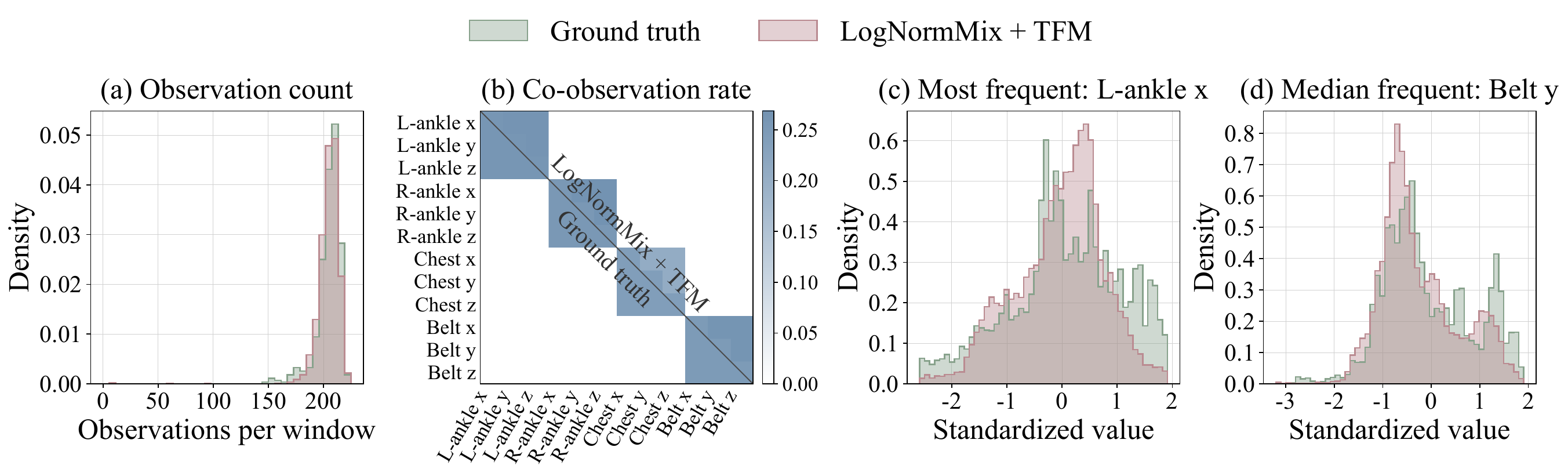} \vspace{-.12in}
\includegraphics[width=\textwidth]{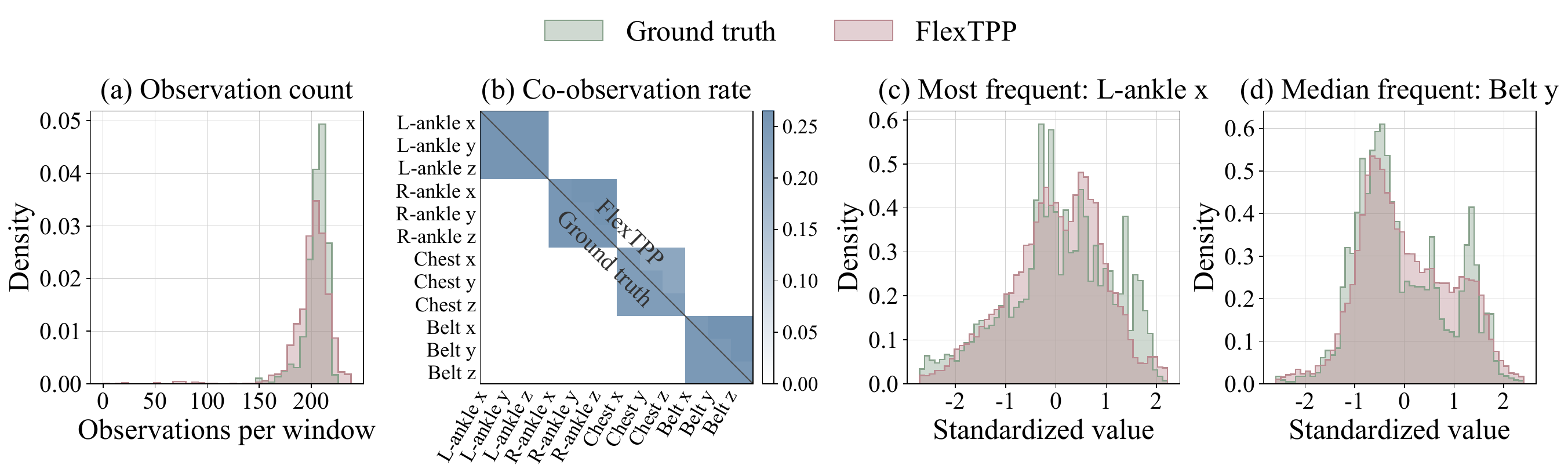} \vspace{-.12in}
\includegraphics[width=\textwidth]{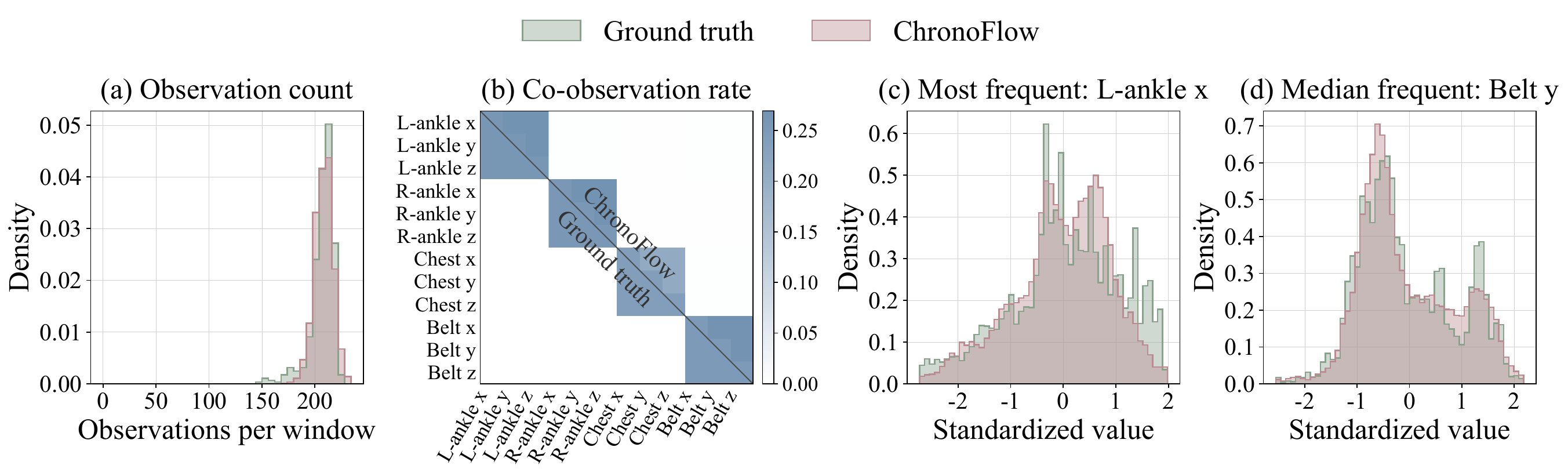} \vspace{-.15in}
\caption{
Qualitative comparison of baseline methods and ChronoFlow on the Activity benchmark.
For this dataset, rows correspond to Poisson + LatentODE, LogNormMix + mTAN, LogNormMix + TFM, FlexTPP, and ChronoFlow, respectively.
For each model, (a) shows the distribution of the total number of observations per window.
Panel (b) shows feature co-observation rates for the 12 most frequent features.
The lower triangle shows ground truth, the upper triangle shows generated samples, and the diagonal shows ground-truth marginals.
Panels (c, d) show value distributions of the most frequent feature and the median-frequency feature.
}
\label{fig:activity_fidelity_baselines}
\vspace{-.15in}
\end{figure*}

\begin{figure*}[t]
\centering
\includegraphics[width=\textwidth]{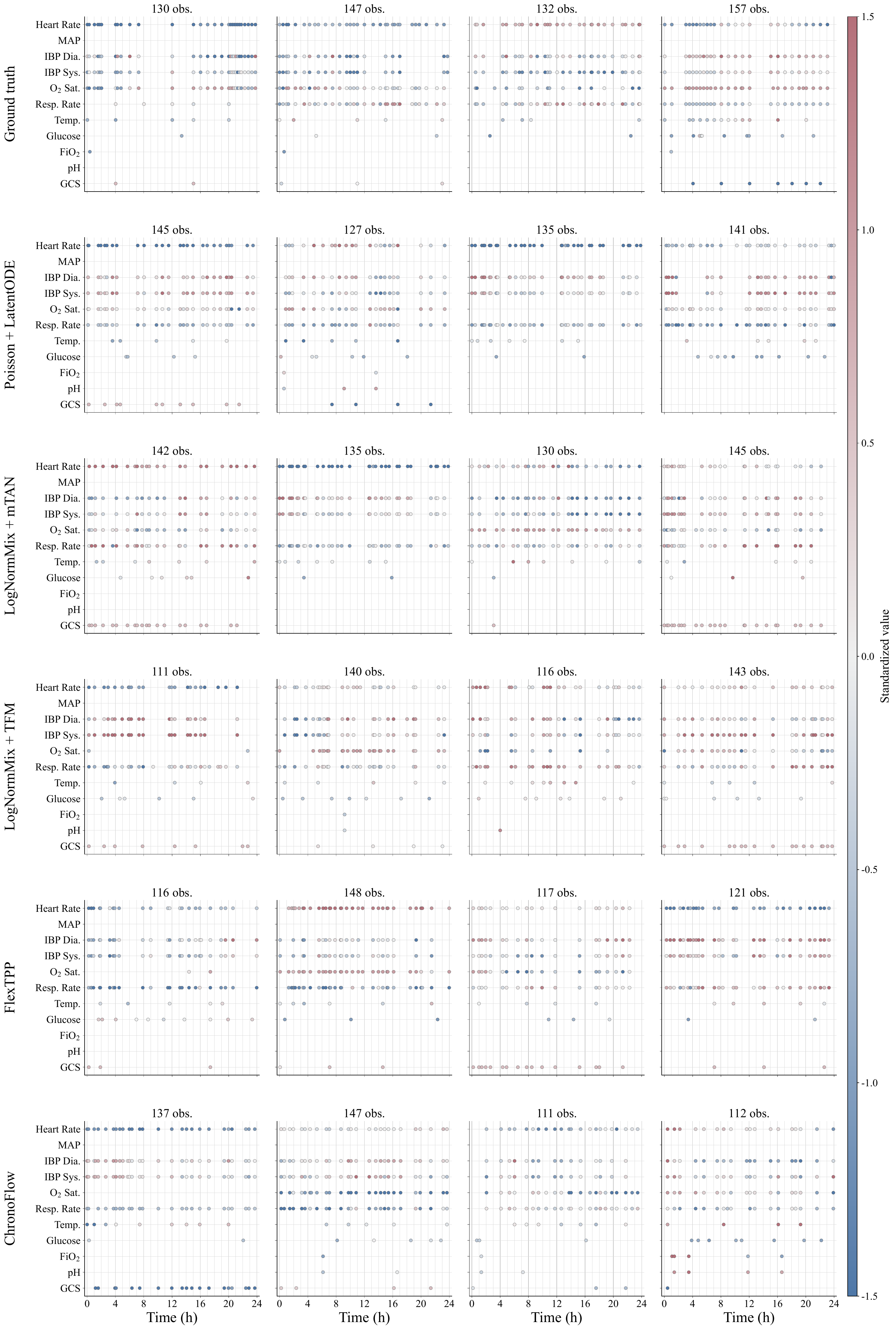}
\vspace{-.12in}
\caption{
Qualitative comparison of generated samples on the eICU benchmark.
Rows correspond to ground truth, Poisson + LatentODE, LogNormMix + mTAN,
LogNormMix + TFM, FlexTPP, and ChronoFlow, respectively.
Each row shows four samples.
Colors indicate standardized feature values, and the total number of observations
is shown above each sample.
}
\label{fig:eicu_main_samples}
\vspace{-.15in}
\end{figure*}

\begin{figure*}[t]
\centering
\includegraphics[width=\textwidth]{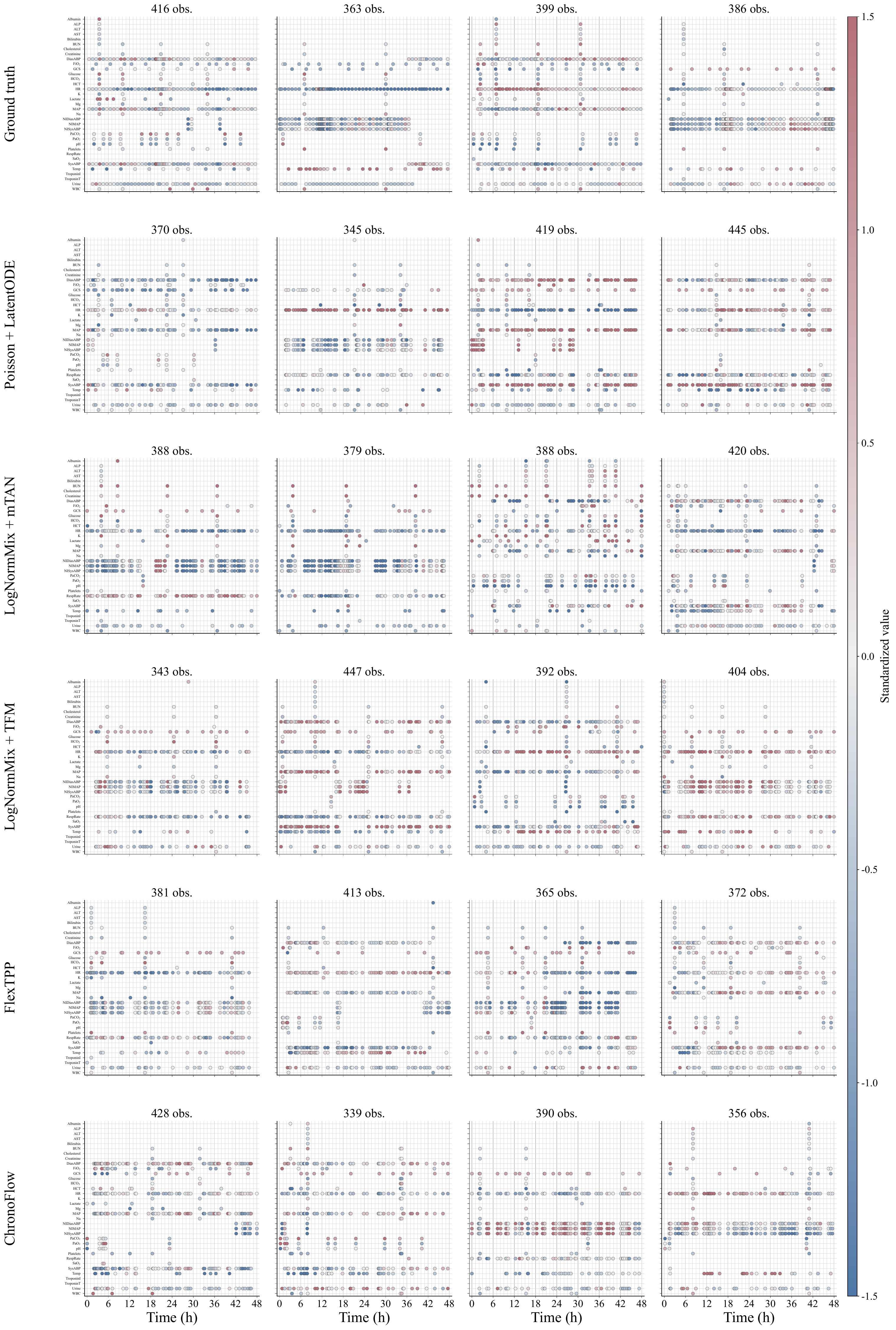}
\vspace{-.12in}
\caption{
Qualitative comparison of generated samples on the P12 benchmark.
Rows correspond to ground truth, Poisson + LatentODE, LogNormMix + mTAN,
LogNormMix + TFM, FlexTPP, and ChronoFlow, respectively.
Each row shows four samples.
Colors indicate standardized feature values, and the total number of observations
is shown above each sample.
}
\label{fig:p12_main_samples}
\vspace{-.15in}
\end{figure*}

\begin{figure*}[t]
\centering
\includegraphics[width=\textwidth]{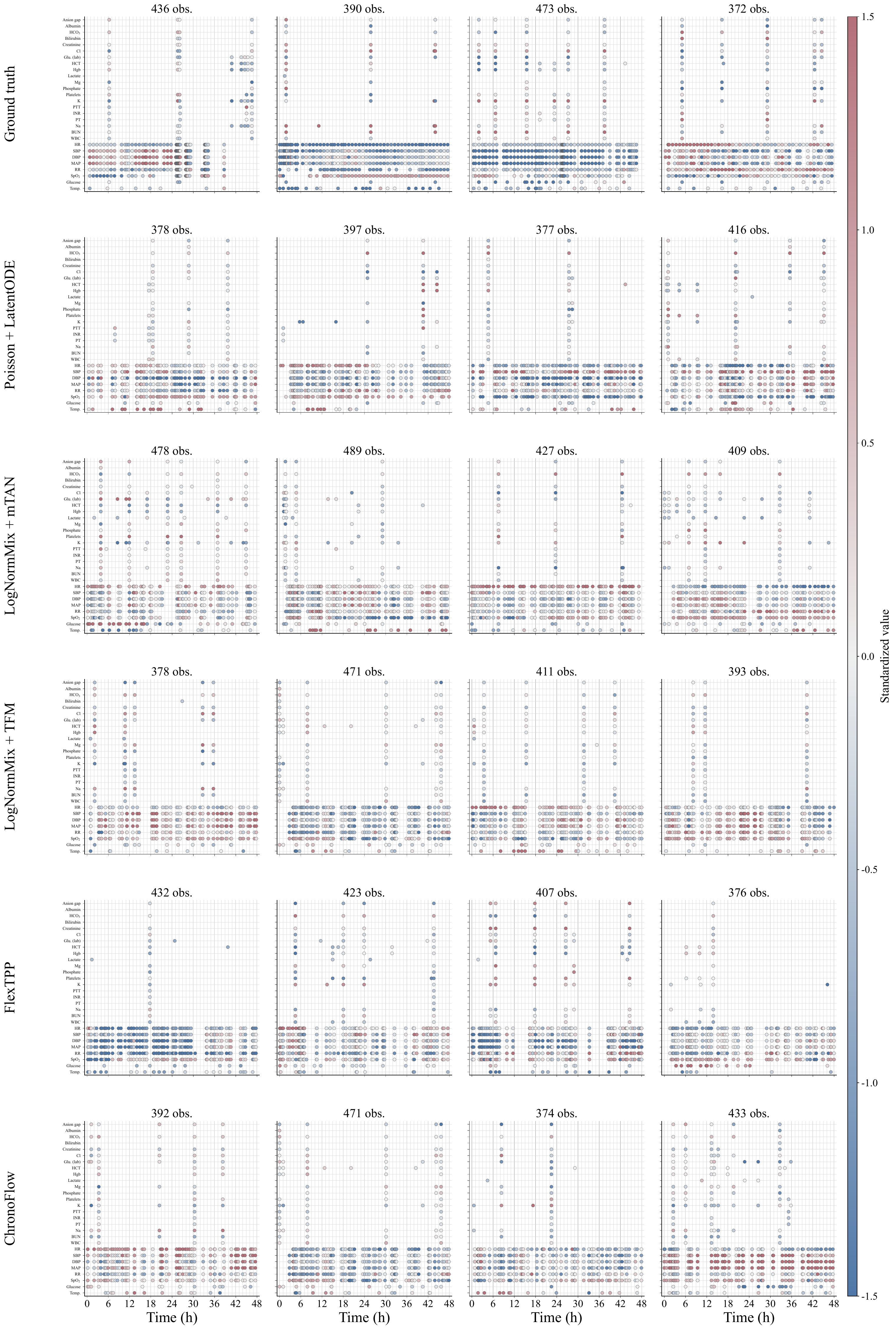}
\vspace{-.12in}
\caption{
Qualitative comparison of generated samples on the MIMIC-III benchmark.
Rows correspond to ground truth, Poisson + LatentODE, LogNormMix + mTAN,
LogNormMix + TFM, FlexTPP, and ChronoFlow, respectively.
Each row shows four samples.
Colors indicate standardized feature values, and the total number of observations
is shown above each sample.
}
\label{fig:mimic3_main_samples}
\vspace{-.15in}
\end{figure*}

\begin{figure*}[t]
\centering
\includegraphics[width=\textwidth]{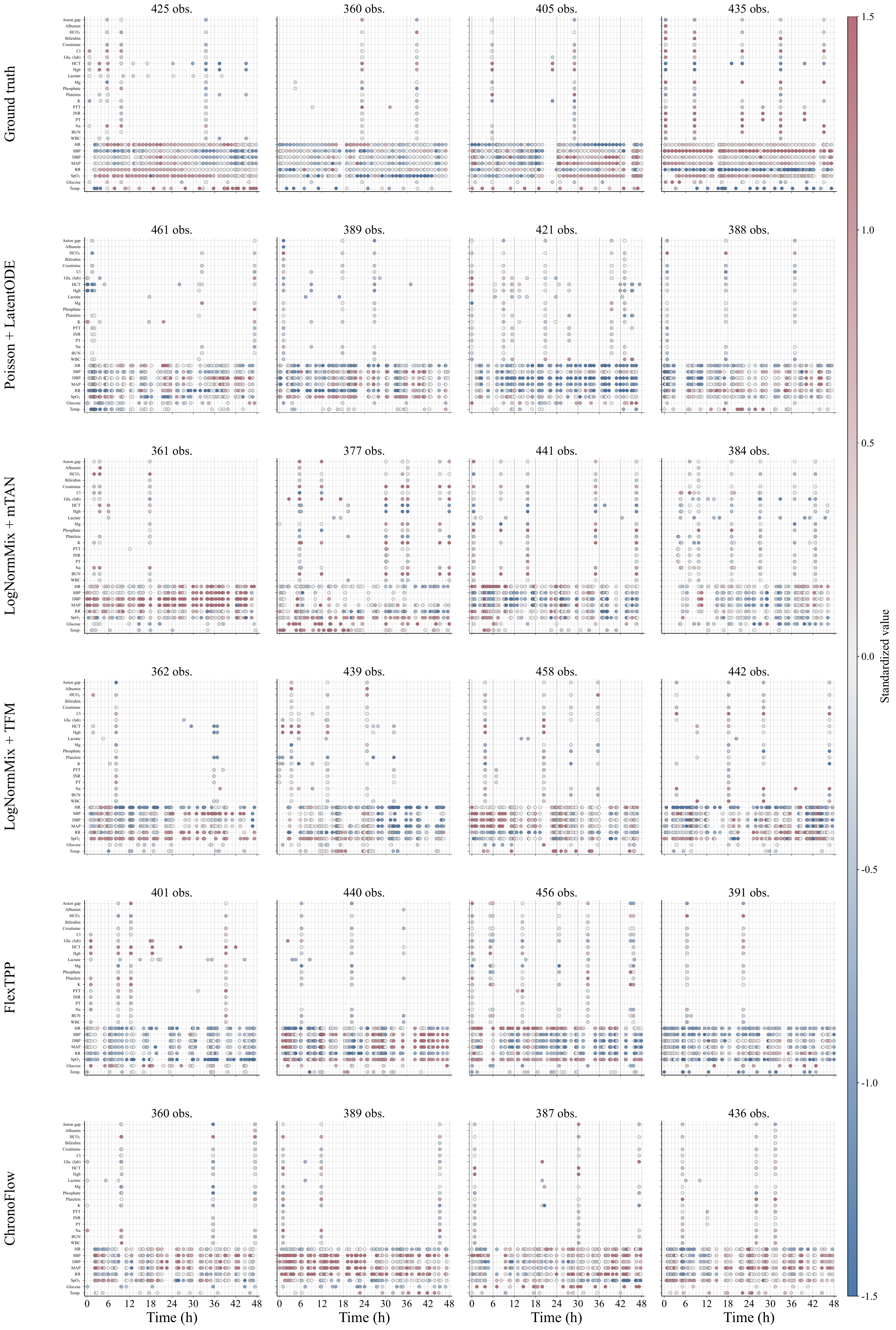}
\vspace{-.12in}
\caption{
Qualitative comparison of generated samples on the MIMIC-IV benchmark.
Rows correspond to ground truth, Poisson + LatentODE, LogNormMix + mTAN,
LogNormMix + TFM, FlexTPP, and ChronoFlow, respectively.
Each row shows four samples.
Colors indicate standardized feature values, and the total number of observations
is shown above each sample.
}
\label{fig:mimic4_main_samples}
\vspace{-.15in}
\end{figure*}

\begin{figure*}[t]
\centering
\includegraphics[width=\textwidth]{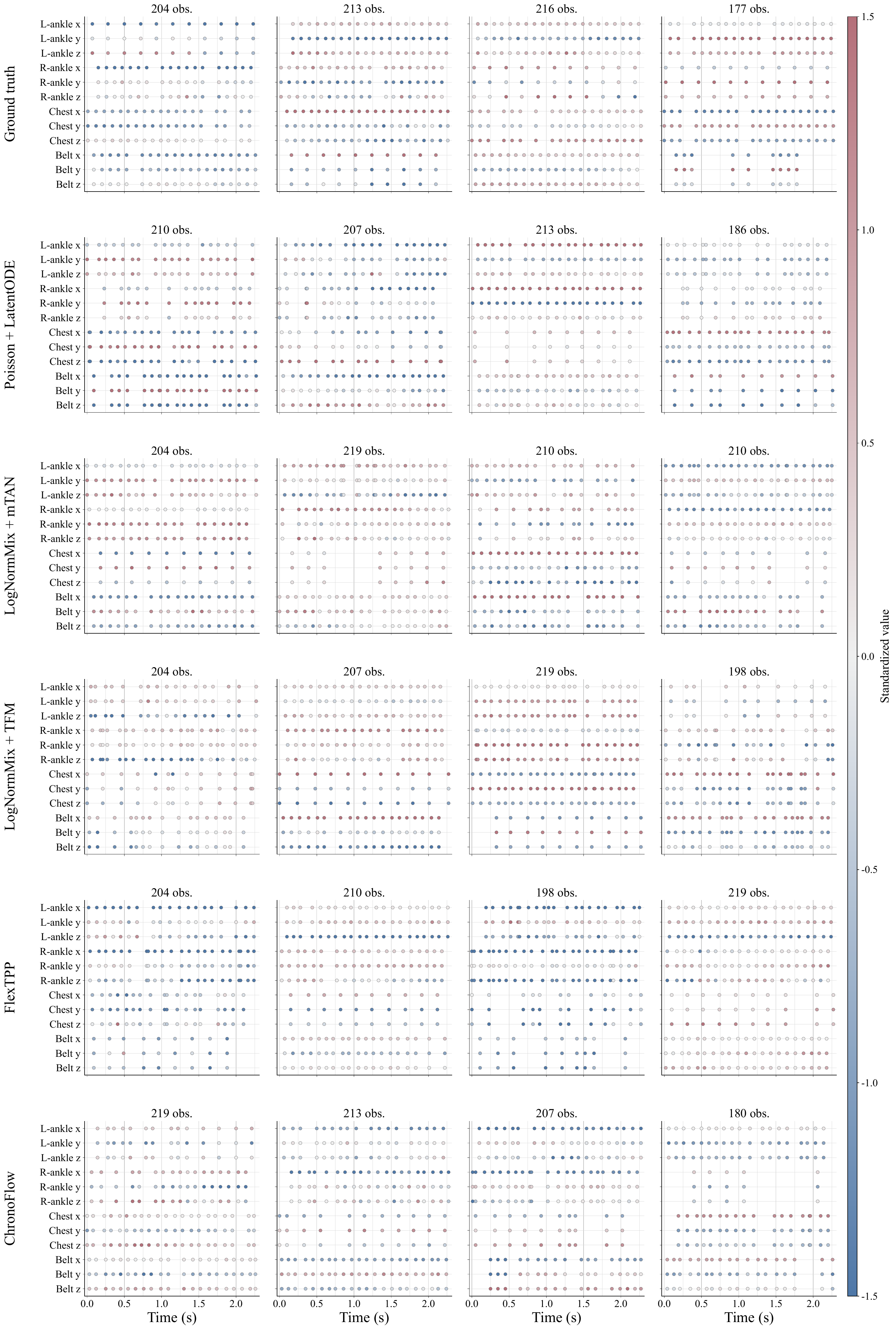}
\vspace{-.12in}
\caption{
Qualitative comparison of generated samples on the Activity benchmark.
Rows correspond to ground truth, Poisson + LatentODE, LogNormMix + mTAN,
LogNormMix + TFM, FlexTPP, and ChronoFlow, respectively.
Each row shows four samples.
Colors indicate standardized feature values, and the total number of observations
is shown above each sample.
}
\label{fig:activity_main_samples}
\vspace{-.15in}
\end{figure*}

This section provides qualitative evidence complementing the quantitative fidelity results in the main paper.
\Cref{fig:umap_all} provides qualitative projections of real and generated samples across all five benchmarks. Sample selection and UMAP randomness use seed $12345$, giving a single-seed illustration rather than a three-seed performance average. Apparent overlap in two dimensions is not a distributional test or evidence against memorization.
\Cref{fig:eicu_fidelity_baselines,fig:p12_fidelity_baselines,fig:mimic3_fidelity_baselines,fig:mimic4_fidelity_baselines,fig:activity_fidelity_baselines} further compare aggregate sampling and value distributions, including observation counts, feature co-observation, and feature-wise value distributions.
Finally, \Cref{fig:eicu_main_samples,fig:p12_main_samples,fig:mimic3_main_samples,fig:mimic4_main_samples,fig:activity_main_samples} present individual generated samples, illustrating feature-wise observation counts, sparse asynchronous temporal layouts, and local value patterns.


\clearpage

\clearpage

\clearpage

\clearpage

\clearpage

\clearpage

\clearpage

\clearpage

\clearpage

\clearpage

\end{document}